\documentclass[11pt]{article}

\usepackage[final]{acl}

\usepackage{times}
\usepackage{latexsym}
\usepackage{float}

\usepackage[T1]{fontenc}

\usepackage[utf8]{inputenc}

\usepackage{microtype}

\usepackage{inconsolata}

\usepackage{graphicx}

\usepackage{booktabs}

\usepackage[T1]{fontenc}
\usepackage[utf8]{inputenc}
\usepackage{fvextra}

\usepackage{algorithmicx}
\usepackage{amsmath,amssymb}

\usepackage[most]{tcolorbox}
\usepackage{xcolor}

\definecolor{ack-bc}{HTML}{1C5CCC}

\newtcbox{\conceptbox}{
  on line,
  colback=ack-bc!15,
  colframe=ack-bc!15,
  arc=6pt,          
  boxrule=0pt,      
  left=4pt,
  right=4pt,
  top=1.5pt,
  bottom=1.5pt
}

\newcommand{\model}[1]{\textcolor[HTML]{2567E9}{\textbf{D#1}}}
\newcommand{\mean}[1]{\textcolor{orange}{\textbf{M#1}}}

\usepackage{txfonts}

\title{Bridging Human Interpretation and Machine Representation: A Landscape of Qualitative Data Analysis in the LLM Era}

\author{
Xinyu Pi\thanks{Equal contribution} \\
UC San Diego \\
La Jolla, CA, USA \\
\texttt{xpi@ucsd.edu}
\And
Qisen Yang\footnotemark[1] \\
UC San Diego \\
La Jolla, CA, USA \\
\texttt{qsyang@ucsd.edu}
\And
Chuong Nguyen\footnotemark[1] \\
UC San Diego \\
La Jolla, CA, USA \\
\texttt{chn021@ucsd.edu}
\And
Hua Shen \\
NYU Shanghai \\
New York University \\
New York, NY, USA \\
\texttt{huashen@nyu.edu}
}

\begin{document}
\maketitle


\begin{abstract}
LLMs are increasingly used to support qualitative research, yet existing systems produce outputs that vary widely--from trace-faithful summaries to theory-mediated explanations and system models.
To make these differences explicit, we introduce a 4$\times$4 landscape crossing four levels of meaning-making (descriptive, categorical, interpretive, theoretical) with four levels of modeling (static structure, stages/timelines, causal pathways, feedback dynamics).
Applying the landscape to prior LLM-based automation highlights a strong skew toward low-level meaning and low-commitment representations, with few reliable attempts at interpretive/theoretical inference or dynamical modeling.
Based on the revealed gap, we outline an agenda for applying and building LLM-systems that make their interpretive and modeling commitments explicit, selectable, and governable.
\end{abstract}

\section{Introduction}
\label{sec:intro}

Qualitative data analyasis, or Qualitative Research (QR) is a family of methods for constructing meaning from text, discourse, and other rich records of human activity \cite{flick2013sage}.
It underwrites how we study experience, organizations, culture, and institutions when the relevant evidence is not naturally reducible to numerical measurements.
With its property of turning unstructured text corpora into defensible meaning and structure, QR is increasingly popular well beyond its traditional homes in the social science community.
Across domains, researchers and practitioners must make sense of messy, high-volume traces where the relevant structure is only partially observable: 
(i) \emph{AI and system builders} (e.g., LLM development, multi-agent systems, architect engineering) need failure narratives, typologies, and mechanism hypotheses rather than only aggregate accuracy \citep{cemri2025mast, szajnfarber2017qualitative}; 
(ii) \emph{human- and society-facing communities} (e.g., HCI, education, policy, ethics) need interpretations of norms, values, and context-dependent meaning that scalar ratings often erase \citep{berg2001qualitative}; and 
(iii) \emph{observational scientific communities} often integrate heterogeneous records to reconstruct processes when controlled experiments are infeasible \citep{Cleland2011PredictionAE}.

However, conducting QR is difficult for many practical reasons.
Three exemplar recurring scenarios are: 
(i) aligning and iteratively revise a codebook across hundreds of interviews while maintaining audit trails and resolving disagreements;
(ii) extracting and reconciling latent mechanisms from heterogeneous artifacts (interviews, policies, reports) with various units of analysis; 
(iii) diagnosing multi-agent system logs that are long, branching, where failures can be distributed across roles and time \citep{saldana2021coding, miles2014qualitative, cemri2025mast}.
In such settings, the bottleneck is not access to text, but the labor of constructing and defending coherent meaning.

\begin{figure}
\includegraphics[width=1\columnwidth]{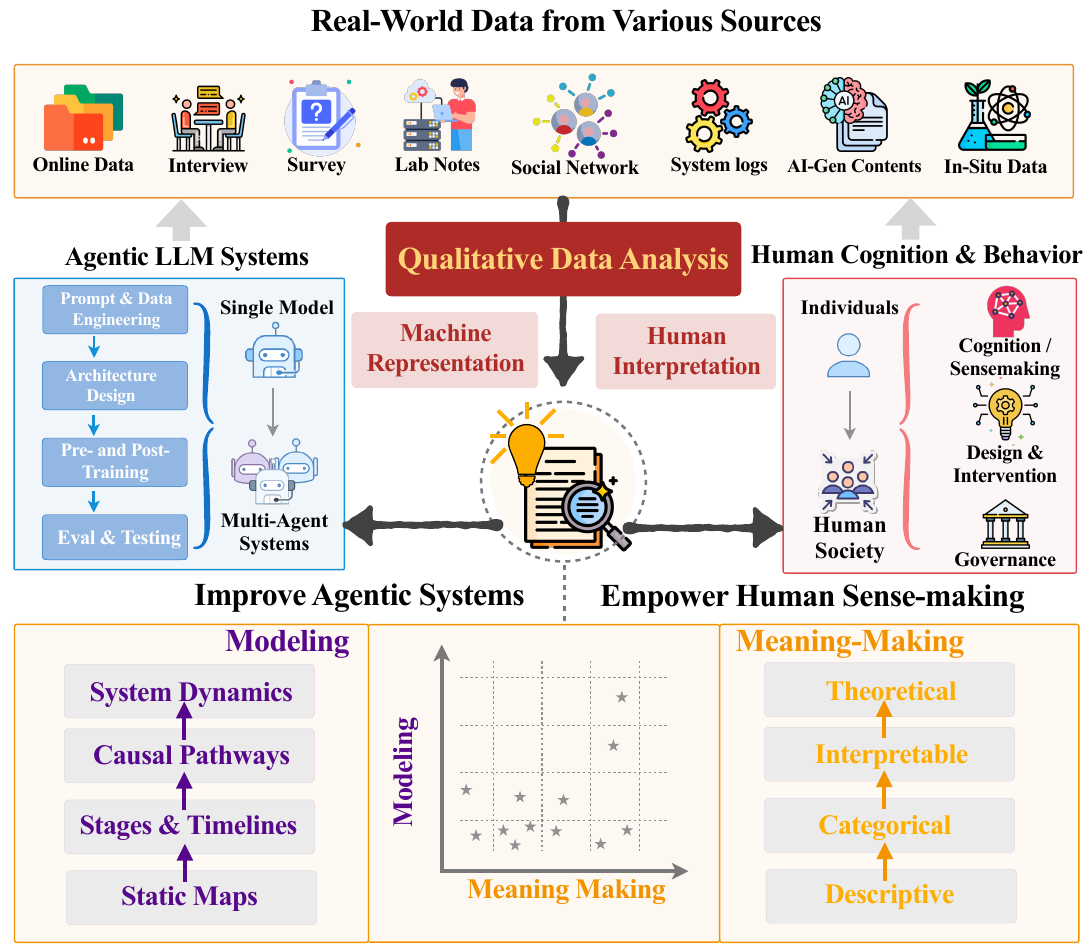}
  \caption{The Synergy of Qualitative Research and LLM developements in broader scientific context.}
  \label{fig:overview}
  \vspace{-6mm}
\end{figure}

To address these barriers, researchers have explored computational support for QR \citep{ferreira2024large, parfenova2025text}.
Recently, LLMs offer a promising lever by combining scalable extraction with increasingly capable interpretation and reasoning \citep{openai2023gpt4}, and have been applied to workflows such as qualitative coding and thematic analysis \citep{Dai2023LLMintheloopLL, Lam2024ConceptIA, Qiao2025ThematicLMAL, pi2025logos}.

Despite this momentum, existing LLM-driven QR efforts exhibit two structural limitations:
\textit{(i)} Most systems treat qualitative analysis as \emph{code induction plus aggregation}: they generate labels, themes, or clusters, but stop short of inducing the structured representations that many QR outcomes ultimately reside on (e.g., phase models of change and causal pathways).
\textit{(ii)} QR is a family of methods with different analytic commitments and output forms---including interpretative phenomenological analysis (IPA), grounded theory, thematic analysis, discourse analysis, narrative analysis, and related traditions \citep{smith2009ipa, glaser1967discovery, charmaz2014constructing, braun2006thematic, miles2014qualitative}.
Current automation tends to be fragmented across these ``species,'' encouraging bespoke pipelines per method even though many share reusable subroutines \citep{barros2025largelanguagemodelqualitative}.

We follow principles of modern system design \citep{farley2021modern}:
rather than having a bespoke pipeline for each species, we aim to identify a small set of interoperable building blocks and a \emph{specification language / design protocol} that makes a QR's standard explicit---what kind of meaning-making it claims to commit, and what kind of model it claims to produce.
Under this protocol, qualitative automation can be organized as a system of \textbf{high-cohesion, low-coupling} modules that \textbf{compose flexibly} to support diverse QR species.

To identify the recurring dimensions that organize qualitative research outcomes, we conducted an extensive analysis of hunndreds of prior QR papers, consulted domain experts, and converged on two condensed dimensions after $5$ months' iterations (App.~\ref{app:theory-dev}).
The first axis captures \textcolor{orange}{\textbf{level of meaning-making}}, from surface description to theory-mediated inference.
The second captures \textcolor{blue}{\textbf{level of modeling}}, from static themes and relational maps to explicit mechanism and system-dynamics models \citep{sterman2000business, meadows2008thinking}.
This yields our primary model of the design space: a \textbf{4$\times$4 landscape} defined by crossing the \textcolor{orange}{4 levels of meaning-making} with the \textcolor{blue}{4 levels of modeling}.

Using this landscape as an operational annotation scheme, we \textit{manually} mapped \textbf{300} QR items onto the space (Fig.~\ref{fig:landscape}).
The resulting distribution makes a clear empirical point: most existing systems cluster in regions associated with \emph{shallow meaning-making} and \emph{low-commitment modeling}---even when the motivating task would naturally call for deeper interpretation or richer structural explanations \citep{barros2024llmqr,parfenova-etal-2024-automating,schroeder2025llmqr}.
This gap is not merely a matter of degree. Progressing from surface themes to interpretive claims requires explicit warrants for latent inferences, while moving from relational sketches to richer models demands concrete modeling commitments that specify how phenomena evolve, not just which concepts are linked.

Building on this diagnosis, we use the landscape not only as a descriptive map but as a forward-looking design and governance scaffold.
In Sec.~\ref{sec:agenda}, we translate the observed skew into concrete implications for both qualitative researchers and NLP/LLM researchers -- providing suggestions on how QR analytic commitments should be specified and governed, which representational moves are needed to move beyond modeling-shallow outputs, and what kinds of artifacts and infrastructures are required to boost LLM-assisted QR research.

\section{From Unstructured Data to Structured Insights and Models}
\label{sec:traces-to-structure}

A central reason qualitative research (QR) is powerful yet difficult is that many target phenomena are \emph{not directly observable as variables}.  
In organizational, sociotechnical, and institutional settings, we often care about norms, coordination regimes, failure patterns, interaction dynamics, adaptation, and multi-party sensemaking that are system-level and latent.  
What we typically observe are \emph{partial unstructured datum}: interviews, documents, fieldnotes, meeting minutes, chat transcripts, incident reports, and (in AI settings) interaction logs and tool traces \citep{cemri2025multiagentllmsystemsfail}.  
These traces are perspectival, incomplete, and context-bound.  
Like the ``blind men and the elephant,'' each trace touches only part of an underlying mechanism, and apparent inconsistencies can become coherent only after within-frame interpretation and integration.

This challenge is especially salient for modern LLM deployments. 
LLM systems become multi-agent, tool-mediated, and long-horizon, the object of analysis is less a single input--output mapping and more an evolving interaction process \citep{wu2023autogen,qian-etal-2024-chatdev,hong2023metagpt}.  
Consequently, diagnosis often takes the form of qualitative analysis over long traces (who believed what, when coordination broke, why verification failed).  
The outputs are frequently QR artifacts such as failure typologies, phase narratives, and feedback explanations \citep{cemri2025mast}.  
Methodologically, ``conducting QR'' is therefore not merely \emph{summarizing text}.  
It is a disciplined attempt to reconstruct latent structure from fragmentary evidence.

\subsection{A quantitative--qualitative analogy}
\label{subsec:quant-qual-analogy}

To make this reconstruction problem legible to an NLP/ML audience, it helps to contrast QR with a classical quantitative modeling pipeline.  
In quantitative modeling, the analyst often begins with a specified observation space (variables/features) and a model family, and success is anchored by an explicit fit criterion (likelihood, loss, $R^2$) \citep{box1976science}.  
In QR, by contrast, the observation space is often \emph{textual} and the latent variables are not given.  
Analysts must \emph{induce} semantic units and justify the inferences that link excerpts to claims about the underlying phenomenon \citep{miles2014qualitative,saldana2021coding,lincoln1985naturalistic}.  
This is why QR emphasizes traceability (showing how claims connect to excerpts), credibility/trustworthiness, and iterative refinement of codes and concepts \citep{krippendorff2018contentanalysis,graneheim2004trustworthiness}.

The analogy is imperfect, but it highlights a computational point. 
QR requires making two kinds of commitments explicit: what meanings are inferred \emph{from and across traces}, and what global structure is asserted \emph{over those meanings}.
In case-based theory building, for example, researchers emphasize disciplined within-case interpretation and careful cross-case synthesis, rather than optimization of a scalar objective \citep{george2005case}.

\begin{table}[t]
\small
\centering
\footnotesize
\begin{tabular}{p{0.21\linewidth}p{0.3\linewidth}p{0.35\linewidth}}
\toprule
\textbf{Feature} & \textbf{Quantitative data analysis} & \textbf{Qualitative data analysis} \\
\midrule
Input data & Numeric/tabular observations & Unstrucutred data (interviews, docs, logs) \\
Primary inference & Parameter estimation, hypothesis testing & Meaning inference + structural reconstruction and integration \\
Model products & Regression, classifiers, ODEs & Themes, stage models, causal graphs, feedback systems \\
Eval target & Fittness / predictive loss & Traceability and consistency to corpora \\
\bottomrule
\end{tabular}
\caption{A motivating analogy between quantitative and qualitative analysis. Both involve modeling under uncertainty, but differ in input modality, inferential focus, representational products, and evaluation criteria.}
\label{tab:quant-qual-analogy}
\vspace{-6mm}
\end{table}

\subsection{Two coupled inference problems}
\label{subsec:two-inference-problems}

Because the underlying phenomenon is only indirectly visible through traces, qualitative analysis must solve two coupled inference problems—one about meaning (both within and across traces) and one about structure (how inferred meanings are organized into system-level claims):

\paragraph{(1) Meaning inference within and across traces.}
Given an excerpt—and often by triangulating multiple excerpts across people, times, or artifacts — what can we \emph{legitimately infer} beyond surface form?
This includes deciding whether we are describing what is said, organizing it into categories, inferring implicit commitments entailed by the situation, or applying an external conceptual lens \citep{berelson1952contentanalysis,hsieh2005threeapproaches,gioia2013rigor,krippendorff2018contentanalysis}.
This meaning-making can happen cross-trace (e.g., resolving apparent contradictions, refining category boundaries, or semantically aggregating shared patterns).

\paragraph{(2) Across-instance structural reconstruction.}
Given many such inferences, what \emph{kind of system-level structure} are we claiming?  
A study may synthesize traces into a thematic map, a phase model, a mechanism diagram, or a dynamical feedback explanation, and these end products impose different validity demands \citep{miles2014qualitative,george2005case}.  
Adjacent traditions make the same point with different formalisms, including process tracing (mechanism evidence), mechanism-based explanation (generative processes), and configurational approaches such as QCA (combinations of conditions) \citep{bennett2015processtracing,hedstrom1998socialmechanisms,ragin1987comparative}.  

\subsection{Why we factorize into meaning-making and modeling}
\label{subsec:why-factorize}

These two inference problems motivate a coordinate system rather than a single catch-all label of ``analysis.''  
\textit{(i)} We need a \textbf{level of meaning-making} axis to state how strong the semantic leap is at the level of evidence interpretation: paraphrase vs.\ categorization vs.\ within-frame interpretation vs.\ theory-mediated reframing within or across traces.
Without this axis, surface-level reorganization can be conflated with analysis involving substantive interpretation or theoretical commitment.
\textit{(ii)} We need a \textbf{level of modeling} axis to state what form of system claim is made at the \emph{phenomenon level}: static relational configuration, staged process, causal dependency structure, or evolving feedback system.  
Without this axis, ``building theory'' collapses into one bucket, even though a taxonomy, a phase model, a causal graph, and a dynamical system model are qualitatively different products with different criteria for warrant and evaluation \citep{zeigler2000theory,law2015simulation,sterman2000business}.  

Taken together, the two axes reduce category errors (e.g., mistaking thematic coding for a mechanism explanation), align evaluation with the type of claim being made, and sharpen where current LLM capabilities reliably lie.  
We therefore introduce the four levels of meaning-making in Sec.~\ref{sec:meaning-making} and the four levels of modeling in Sec.~\ref{sec:level-modeling}, and use their cross-product as the organizing map for LLM-driven qualitative analysis.

\section{Landscape Model: a 4$\times$4 Conceptual Framework for Qualitative Outputs}
\label{sec:landscape}

Qualitative analysis reconstructs latent phenomena from \emph{partial textual traces} (interviews, documents, logs, incident reports, transcripts) by making two kinds of commitments: (i) what is inferred from any given excerpt, and (ii) what global structure is asserted across excerpts \citep{miles2014qualitative,saldana2021coding,lincoln1985naturalistic}.
We factorize these commitments into two analytically separable axes. 
The first axis is a \textbf{level of meaning-making} (how far the analysis moves beyond surface-faithful restatement toward interpretation and theory-mediated reframing), drawing on classic manifest/latent and emic/etic distinctions \citep{berelson1952contentanalysis,krippendorff2018contentanalysis,graneheim2004trustworthiness,pike1954language}. The second axis is a \textbf{level of modeling} (what explicit representational structure is produced: static maps, stage models, causal mechanisms, or dynamical feedback systems), drawing on system science and simulation traditions \citep{box1976science,zeigler2000theory,law2015simulation}.

Importantly, two axes are analytically separable: the same interpretive claim can be expressed as a thematic map, a phase narrative, a causal diagram, or a feedback model. Their cross-product yields a \textbf{4$\times$4 landscape} that makes QR products comparable across methodological ``species'' and clarifies why evaluation (and LLM-based automation) must be aligned to the kind of claim being made.

\subsection{\textcolor{orange}{Dimension 1: Level of Meaning-Making}}
\label{sec:meaning-making}

We use \emph{meaning-making} to describe \textbf{what an analysis adds beyond the literal statements}.
At low levels, analytic outcomes stay trace-faithful. 
At higher levels, outcomes introduce stronger \textit{semantic commitments}: they infer latent intent and causations, and may further re-interpret the case through an external theoretical lens.
We distinguish $4$ levels of meaning-making by whether the output commits to
(i) \textbf{implicit inference} beyond what is explicitly stated, and
(ii) \textbf{external conceptual resources} beyond the case's own frame.
We only briefly discuss our definition here and give a more rigorous articulation in App.~\ref{app:meaing} due to page limit.


\paragraph{\mean{1}: Descriptive (manifest, surface-faithful).}
\mean{1} outputs restate, summarize, or extract \emph{explicit} content with minimal abstraction and minimal inference. This aligns with ``manifest'' content analysis: what is directly observable in the record and checkable against the source \citep{berelson1952contentanalysis,holsti1969contentanalysis,krippendorff2018contentanalysis}. In computational terms, \mean{1} corresponds to grounded summarization and information extraction when the output remains trace-faithful.

\paragraph{\mean{2}: Categorical (themes, topics, patterned organization).}
\mean{2} outputs introduce \emph{organizational abstraction}: they group observations into recurring categories/themes/types (a reusable label system) while remaining primarily \emph{frame-faithful}---they organize what appears in the corpus more than they assert deeper latent ``why'' claims \citep{braun2006thematic,miles2014qualitative,saldana2021coding}. In computational terms, \mean{2} aligns with thematic summarization, taxonomy induction, and clustering/topic-organization when the result is an organization of surface content rather than an inference about implicit commitments.

\paragraph{\mean{3}: Interpretive (implicit meaning entailed by the frame).}
\mean{3} outputs articulate \emph{implicit or latent meaning} that is not explicitly stated but is plausibly \emph{entailed by the situation's semantic frame}---e.g., underlying goals, obligations, causal mechanisms, hidden patterns, pragmatic intent, or unstated constraints that make the episode coherent \citep{graneheim2017challenges}. 
This level is grounded in (i) \textbf{frame-based inference} \citep{fillmore1982framesemantics,baker1998framenet}, (ii) \textbf{pragmatics and implied commitments} \citep{grice1975logic,searle1969speechacts,stalnaker2002commonground}, and (iii) interpretivist accounts of sensemaking and thick description \citep{goffman1974frameanalysis,weick1995sensemaking,geertz1973interpretation}.
For NLP readers, \mean{3} is closest to entailment-style inference in spirit \citep{dagan2006rte,bowman2015snli}: the claim is \emph{defeasible} but should be warranted by cues and shared background assumptions appropriate to the case.

\paragraph{\mean{4}: Theoretical (external reframing; etic constructs).}
\mean{4} outputs re-situate the material inside an \emph{external theoretical or conceptual system} that is not licensed by the frame alone (etic vocabulary, theoretical coding, sensitizing concepts). The defining criterion is that the imported framework functions as an \emph{analytic engine} shaping what counts as evidence and how claims are constructed \citep{hsieh2005threeapproaches,gioia2013rigor,saldana2021coding}. This includes theory-guided seeing and abductive/theory-mediated reinterpretation \citep{blumer1954wrongtheory,tavory2014abductive}, as well as productive interpretation in hermeneutic traditions \citep{gadamer1989truthmethod,ricoeur1976interpretationtheory}. Because \mean{4} introduces stronger conceptual scope, it raises higher demands for warrant and explicit limits.



\subsection{\textcolor[HTML]{2567E9}{Dimension 2: Level of Modeling}}
\label{sec:level-modeling}

Here, \emph{modeling} refers to the \textbf{explicit representational structure} produced to describe a phenomenon: what the units are, what relations hold among them, and what kinds of reasoning the representation supports (models as representations, not necessarily predictive estimators) \citep{box1976science,zeigler2000theory,law2015simulation}. We distinguish $4$ levels by whether the model commits to (i) \textbf{time}, (ii) \textbf{causality/mechanism}, and (iii) \textbf{system dynamics} (loopy iterative state updates and feedback). 
An expanded discussion can be found in App.~\ref{app:modeling}.

\paragraph{\model{1}: Static taxonomy and relational configuration models.}
\model{1} models are \textbf{static}: they represent themes/constructs/topologies and relationships without committing to temporal evolution or causal production. Typical outputs include thematic maps/networks \citep{braun2006thematic,attridestirling2001thematic}, cross-sectional relational/network representations \citep{wasserman1994social,borgatti2009network}, and concept maps used to organize relations without time/causality semantics \citep{trochim1989concept,novak2008conceptmaps}. Signature: ``what relates to what'' in a snapshot.

\paragraph{\model{2}: Stage/phase/timeline models (time without causality).}
\model{2} models make \textbf{time} first-class: the phenomenon is organized as stages, episodes, or a timeline, but edges primarily mean ``next/then'' rather than ``produces/changes'' \citep{vandevenpoole1995explaining,langley1999strategies,pettigrew1990longitudinal}. 
Related sequence-analytic descriptions belong here when the primary commitment is ordering and segmentation rather than mechanism \citep{abbott1995sequence,abbott2001timematters}. Signature: ``what happens when.''

\paragraph{\model{3}: Causal dependency and mechanism models (directed influence).}
\model{3} models explicitly represent \textbf{directed influence}: nodes are constructs/states/actors and edges encode causal/mechanism-like dependencies (enable/constraint, produce/change, mediate) \citep{pearl2009causality,spirtes2000cps}. This includes qualitative causal maps and fuzzy cognitive maps when used as influence structures \citep{kosko1986fuzzy}. Signature: ``why/how'' as directed production, typically still read as a (possibly qualitative) dependency graph rather than an iterated dynamical system.

\paragraph{\model{4}: Dynamical systems / feedback / complex systems models (state + update + iteration).}
\model{4} models treat the phenomenon as an \textbf{evolving system} whose behavior depends on \textbf{iterative state change} (state at $t$ shapes state at $t{+}1$), often via feedback, delays, and nonlinear responses \citep{forrester1961industrial,sterman2000business,meadows2008thinking}. This includes qualitative system-dynamics reasoning (reinforcing/balancing loopy machinery), coupled state-transition accounts with endogenous evolution, and simulation models (agent-based, discrete-event) when used to explain emergent trajectories \citep{bonabeau2002agent,macywiller2002actors,epstein2006generative,gilbert2005simulation}. Signature: the core explanatory claim depends on iteration and feedback, not just a static pathway diagram.


\begin{figure}
    \centering
    \includegraphics[width=1.05\linewidth]{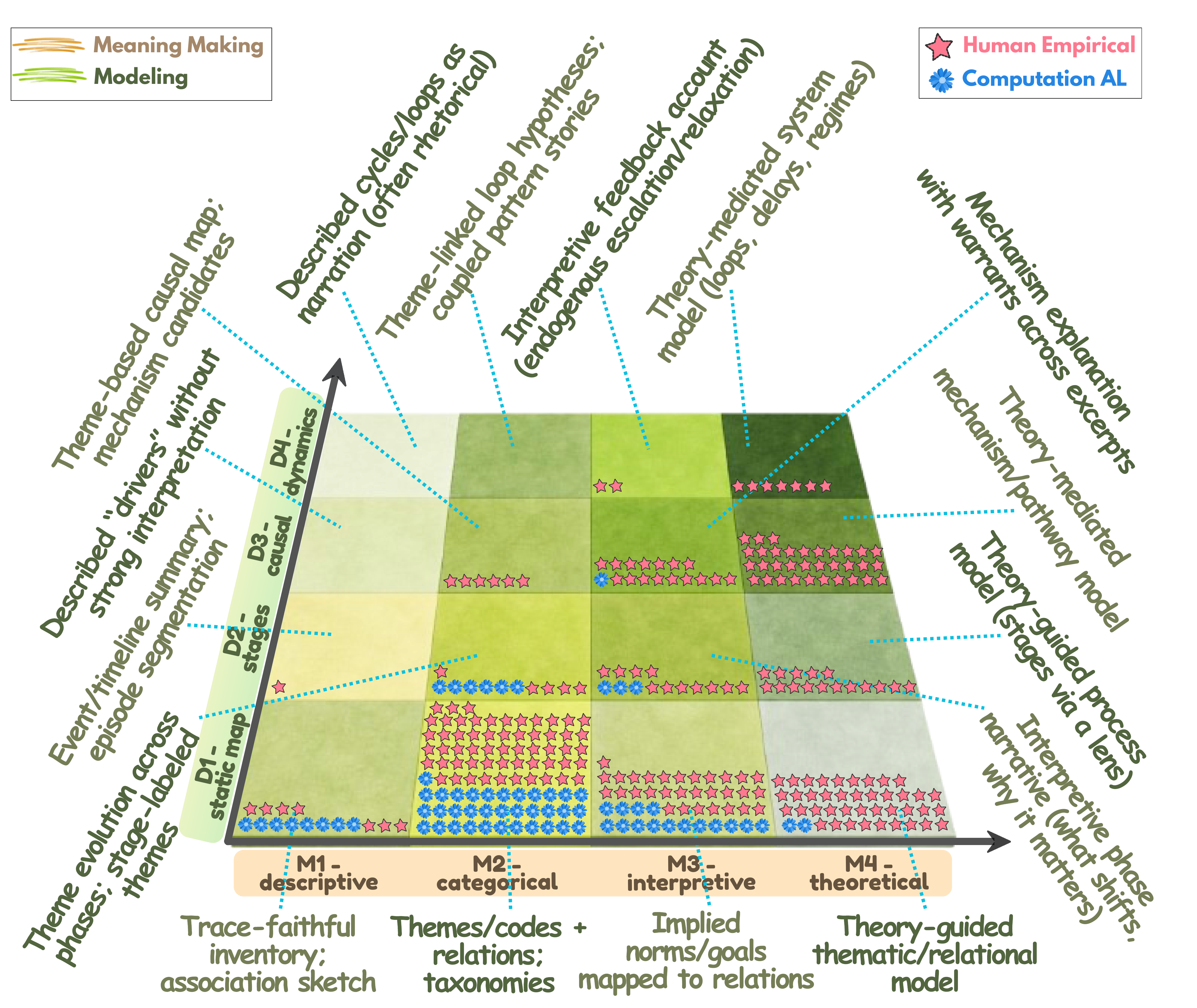}
    \caption{\textbf{QR Landscape}: Our $4 \times 4$ landscape model and annotated paper distribution.}
    \label{fig:landscape}
    \vspace{-6mm}
\end{figure}

\subsection{The integrated 4$\times$4 conceptual landscape}
\label{sec:landscape-agenda}

Figure~\ref{fig:landscape} integrates our two axes---\textbf{meaning-making} (rows) and \textbf{modeling} (columns)---into a $4\times4$ landscape that functions as a compact coordinate system for qualitative outputs.
Each cell corresponds to a distinct \emph{level of semantic commitment} and a distinct \emph{level of representation}:
moving right  increases semantic commitment (from surface-faithful description to external-lens theorizing), while moving upward increases representational commitment (from static relational structure to time, causal dependence, and system dynamics).
With this decomposition of qualitative research commitments, we can situate different qualitative research traditions (e.g., content analysis, thematic analysis, grounded theory, IPA) within the landscape.
We present detailed methodological mappings later in Appendix~\ref{app:method-mapping} due to page limit.

\paragraph{Why the landscape matters for automation and evaluation.}
The landscape prevents category errors (e.g., mistaking strong thematic coding for a mechanism explanation) and makes explicit that ``good'' is cell-dependent: \mean{1} is judged primarily by trace fidelity; \mean{2} by category coherence and coverage; \mean{3} by inference warrant and stability under alternative readings; \mean{4} by appropriateness and disciplined application of the external lens \citep{lincoln1985naturalistic,graneheim2004trustworthiness}. 
On the modeling axis, \model{1} is judged by structural organization; \model{2} by temporal fidelity and boundary clarity; \model{3} by mechanism evidence and directional warrant; \model{4} by whether feedback/iteration is genuinely supported (loops ``close'' with reciprocal evidence) \citep{sterman2000business,pearl2009causality}.

\subsection{Operationalizability Experiment}
\label{sec:operationalizability-exp}

As a complementary validation of the clarity and operationalizability of our theoretical framework, we conducted an independent automatic annotation pass using \texttt{GPT-5.2} on the \textit{human empirical papers} used in grounding study (paper set and protocol in Sec.~\ref{sec:annotation-study}).
The model was provided only with the definitions of our meaning-making and modeling levels and was asked to assign labels without access to human annotations.
We then compared the model’s assignments against human judgments using exact-match agreement. 
The resulting alignment reached $\mathbf{85\%}$ for meaning-making levels and $\mathbf{94\%}$ for modeling levels ((App.\ref{app:op-exp-details})).
Importantly, this evaluation is not intended to assess the model’s analytical capability, but rather to probe whether our classification criteria are sufficiently explicit, internally coherent, and decision-consistent to be applied by an independent agent. 
The high agreement suggests the rubric is sufficiently explicit to be applied consistently, supporting their suitability for scalable analysis and computational downstreams.

\section{Empirical Grounding: An Annotation Study on Existing Qualitative Research}
\label{sec:annotation-study}

\paragraph{Paper Collection Processes}
To empirically ground the proposed landscape, we conduct a manual annotation study that places representative qualitative-research papers into the 4$\times$4 grid. 
The goal is not to exhaustively review the QR literature, but to test whether our two axes -- levels of meaning-making and levels of modeling -- are operational, discriminative, and faithful to real research practice. 
This empirical check prevents the landscape from becoming a purely conceptual taxonomy and provides an evidence-backed view of where existing work concentrates.
We curated two complementary paper sets that reflect the two communities our framework connects -- the  \textbf{empirical QR papers} and \textbf{computational papers}:

\paragraph{Empirical QR papers} refer to classical qualitative research in which human researchers manually code, conduct narrative synthesis, and build models from empirical records. 
We curated papers using a \textbf{PRISMA-style staged screening procedure} (identification, deduplication/screening, eligibility assessment, and inclusion). 
We compile \textbf{$4{,}493$} domain keywords across \textbf{$51$} domains and paire them with \textbf{$10$} QR methodology terms (e.g., grounded theory, thematic analysis, narrative analysis), yielding \textbf{$44{,}930$} query combinations of the form \texttt{<domain keyword> AND <QR method>}.
Using Google Scholar, we retrieve the first \textbf{3 pages} of results per query and obtained \textbf{$664{,}244$} records (identification stage).
We then remove duplicates and retaine only entries with downloadable full texts, resulting in \textbf{$138{,}683$} downloaded papers (screening stage).
Next, we perform an eligibility triage using \textbf{Qwen3-32B} to retain papers that report empirical qualitative studies with no heavy reliance on AI automation (lightweight HCI tools such as NVivo were allowed); we exclud purely computational papers, theoretical papers, and literature reviews, yielding \textbf{$32{,}321$} eligible candidates. 
Finally, we draw a \textbf{simple random sample} of \textbf{$231$} papers from this pool for manual annotation (inclusion stage). Additional details—including the full keyword list, exact prompts/filters, and exclusion breakdown—are provided in the Appendix.

\paragraph{Computational papers} employ computational techniques to automate some major part of QR processes.
In our analysis, we focus exclusively on the meaning-making and modeling responsibilities assumed by LLM-based components, abstracting away from human analytical steps outside the automated pipeline.
Unlike empirical QR, few NLP or AI papers explicitly self-identify as automating qualitative research, which makes purely keyword-driven systematic retrieval insufficient.
Accordingly, our curation process for computational papers adopts an \textbf{exploratory, human-in-the-loop literature expansion strategy based on snowball sampling} \citep{biernacki1981snowball}.
We begin with a set of seed queries (e.g., \texttt{LLM Thematic Analysis}, \texttt{Automatic Schema Induction}, \texttt{Automatic Taxonomy Construction}, \texttt{Event Causal Graph Construction}) and iteratively expand the corpus through backward and forward citation tracing.
Papers are manually assessed and retained only if they propose \emph{novel automation pipelines} for inducing patterns or structures from unstructured data, rather than merely applying off-the-shelf tools.
Following this process, we identify and retain \textbf{$69$} eligible computational papers for annotation.
We acknowledge that our corpora is conditioned on query terms and visibility, and the distributions might not be perfectly unbiased population estimates.

\paragraph{The Paper Annotation Process}
We manually annotated all sampled papers according to the operational definitions of the meaning-making axis (\mean{1}--\mean{4}) and the modeling axis (\model{1}--\model{4}) introduced in Sec.~\ref{sec:landscape}.
Three authors served as annotators.
Each paper was independently annotated by at least two annotators on \emph{both} axes, using a shared rubric that specifies level definitions, boundary tests, and decision rules (Appendix~\ref{app:annotation-details}).
Annotators were instructed to judge the paper's \emph{dominant analytic output} and \emph{dominant substantive model of the target phenomenon} (rather than incidental examples, background discussion, or the computational pipeline itself), and to ground their decisions in concrete representational features (e.g., what nodes/edges correspond to; whether structure is source$\rightarrow$sink pathway-like vs.\ feedback-coupled; whether meaning claims remain surface-faithful vs.\ introduce implicit commitments vs.\ import external theoretical constructs).
Disagreements were resolved through discussion among the annotators, resulting in a single consensus label per paper for each axis.
We report the final consensus annotations in Fig.~\ref{fig:landscape}, and release the full rubric, prompts, and boundary heuristics used for annotation in the Appendix~\ref{app:annotation-details} to support reproducibility.

\paragraph{Observations}
Three consistent patterns can be observed from Fig. ~\ref{fig:landscape}, which together reveals the gap between LLM-based and traditional QR practices.
\textbf{(O1) Computational work is modeling-shallow:} They concentrate heavily in \model{1}--\model{2}, with only sparse coverage of \model{3}--\model{4}.
Many systems do progress along the \emph{meaning-making} axis (from descriptive toward categorical and occasionally interpretive outputs), but this semantic enrichment rarely translates into stronger \emph{representational commitments}.
Even when outputs contain richer semantics, they are typically encoded as static artifacts (e.g., themes, relational maps) rather than mechanism-bearing models.
\textbf{(O2) Human QR is meaning-deep but often representation-light}.
Empirical QR papers are markedly right-shifted on \emph{meaning-making}, with most studies occupying \textbf{\mean{3}--\mean{4}}.
However, many of these papers still instantiate their claims in \textbf{\model{1}--\model{3}} rather than \textbf{\model{4}}. 
In other words, human QR frequently makes strong interpretive or theory-mediated commitments without necessarily formalizing them into explicit feedback- or state-update representations.
\textbf{(O3) Meaning and modeling are coupled in high-end human QR}: In the empirical corpus, high-level modeling almost never appears with low-level meaning-making: the upper-left region (high modeling, low meaning) is essentially empty. 
When papers reach \model{3} (causations) or \model{4} (dynamics), they almost always also make \mean{3}--\mean{4} commitments, suggesting that richer representational forms typically require (and are justified by) deeper interpretive grounding.

\section{Agenda}
\label{sec:agenda}

The observed gaps from Fig.~\ref{fig:landscape} motivates an agenda for both qualitative and NLP/LLM researchers. 
Agenda operationalization details are in App.~\ref{app:agenda_operationalization}.

\subsection{Implications for qualitative researchers}
For qualitative researchers and domain experts, the landscape should be treated less as a ``taxonomy'' and more as a \textbf{specification language for analytic contracts}: it clarifies what kind of output is being requested, what counts as acceptable warrant, and which risks are tolerable. We highlight the following directions: 
\textbf{(1) Specify QR analytic commitments} by explicitly specifying the intended \textcolor{orange}{meaning-making} and \textcolor{blue}{modeling} levels for each analytic step (e.g., \model{1} theme condensation versus \model{3} mechanism proposal), thereby clearly distinguishing descriptive organization from interpretive or theoretical claims and improving transparency, comparability, and accountability across studies and tools.
\textbf{(2) Design governance rules to mitigate risks of LLM-assisted QR.}
Empirical studies and practitioners report consistently emphasize the risks of LLM-assisted qualitative research, including overconfident fabrication, erosion of contextual grounding, and misalignment with qualitative values and human judgment \citep{schroeder2025llmqr, barros2024llmqr, parfenova-etal-2024-automating, ubellacker2024academiaos, reject-reflexivea-qr}.
However, one plausible hypothesis is that these risks might not distribute uniformly across all kinds of qualitative work. 
Not all meaning-making and modeling operations are equally demanding or equally fragile.
It is plausible that LLMs can perform lower-level tasks (e.g., \mean{1}–\mean{2}, \model{1}–\model{2}) with relatively high reliability, while error rates, epistemic drift, and value misalignment increase as semantic and representational commitments deepen.
By explicitly delineating which regions of the landscape are appropriate for automation (often \mean{1}–\mean{2}, \model{1}–\model{2}) and which require sustained human adjudication and positionality-aware interpretation (often \mean{3}–\mean{4}, \model{3}–\model{4}), the question shifts from a binary “should we use LLMs” to a scoped governance decision about when and where automation might be appropriate.
Our framework could conceptually scaffold systematic benchmarking to establish  more rigorous governance rules for applying LLMs in QR.
\textbf{(3) Create shareable gold artifacts that make qualitative standards testable}—including curated codebooks \textit{(the collection of corpora-induced meaning units)}, explicitly documented boundary and counterexamples, and trace-linked rationales that connect excerpts to analytic claims.
Primary qualitative materials (e.g., interviews, transcripts) and their associated codebooks are often closed due to privacy, confidentiality, and ethical constraints, which limits direct reuse for evaluation. 
As a result, the lack of standardized benchmarks that pair raw corpora with aligned, authoritative codebooks forces computational works to rely on bespoke datasets and derived codebooks for evaluation \citep{pi2025logos, Zhong2025HICodeHI, Qiao2025ThematicLMAL}.
Without human ground truth interpretation, the robustness of LLM self-evaluation for \textcolor{orange}{meaning-making} is questionable.
Moreover, QR outcomes are typically narrative and discursive, making them poorly suited for evaluating \textcolor{blue}{model fitness or structural adequacy}.
Converting core narrative claims into crisper representational forms—such as relational graphs, mechanism maps can enable systematic comparison across different QR pipelines.

\vspace{-2mm}
\subsection{Implications for NLP/LLM researchers}
For NLP and LLM researchers building agentic systems for qualitative automation, the landscape suggests four concrete directions:

\textbf{(1) design code relation induction algorithms} that go beyond semantic aggregation of codes  by making explicit the transition from \emph{coding individual data points} to \emph{modeling relationships among codes}. 
In QR terms, this corresponds to moving from open coding (meaning-making over individual instances) to axial coding, where analysts examine how codes relate through causal, temporal, logical, or interactive relations. 
Most existing computational work effectively stops at \model{1} because they treat codes as independent labels or clusters and does not attempt to induce structured relationships among them.
Advancing to \model{2}--\model{4} therefore requires methods that explicitly reason over inter-code relations rather than merely condensing or summarizing code semantics. 
\textbf{(2) enable adaptive level selection}, i.e., mechanisms that let the system decide which level of meaning-making and modeling is warranted by a given corpus, with calibrated abstention or fallback to lower-commitment outputs when evidence is insufficient or ambiguity is high;
\textbf{(3) create evaluation methodologies for qualitative modeling} that operationalize ``fitness-to-corpus'' for \model{2}/\model{3}/\model{4} structures (e.g., evidence coverage, contradiction rate, pattern matching and semantic consistency), since humans perform such assessment intuitively but systematic approximations remain unclear.
Beyond alingment with expert-created ground truth models, automatic goodness-of-fit metrics for qualitative models will be a key missing ingredient for scalable automation and rapid research iterations.

\vspace{-1mm}
\section{Conclusion}

We present a 4×4 landscape that disentangles \textcolor{orange}{meaning-making} from \textcolor{blue}{modeling} in qualitative research.
An annotation study reveals a systematic gap: current LLM-based systems enrich semantics without making stronger representational commitments.
The framework clarifies where automation aligns with qualitative practice, where it fails, and what representational and evaluative advances are needed for reliable QR in the LLM era.



\section{Limitations and Risks}
\label{sec:limitations}

Researchers should remain cautious about using LLMs to automate or accelerate qualitative research (QR). Because LLMs are trained on large-scale text corpora with uneven social coverage, they can reproduce and amplify social biases, stereotypes, and dominant interpretive frames. More fundamentally, LLMs do not possess lived experience, stable positionality, or access to participants' phenomenological and interactional context; consequently, they may produce fluent but ungrounded interpretations, especially when asked to infer motives, norms, or latent meanings beyond what is explicitly supported by the data.

These limitations are particularly salient for paradigms that treat meaning-making as inherently situated and reflexive (e.g., reflexive thematic analysis and interpretative phenomenological analysis): since an LLM lacks a reliable social identity and cannot engage in reflexive accountability in the human sense, it is not generally appropriate to treat it as a replacement for the human analyst \citep{reject-reflexivea-qr}. Persona prompting may change style or surface perspective, but it is not a substitute for identity formation rooted in long-term cultural participation, symbolic interaction, and tacit social learning. As a result, different groups' divergent interpretations of the same statement cannot be resolved by prompting alone, and the model may still default to normative or majority assumptions.

Beyond these conceptual limits, LLM-based pipelines introduce practical risks that affect rigor and trustworthiness:
\begin{enumerate}
    \item \textbf{Over-interpretation and hallucination.} LLM outputs can appear coherent and theory-like while inventing rationales, mechanisms, or “implicit commitments” that are weakly warranted by the trace (a form of meaning-mimicking rather than meaning-making).
    \item \textbf{Loss of traceability.} Without strict evidence-linking, multi-step summarization and synthesis can break the audit trail from excerpts to claims, making it difficult to verify how a code, edge, or mechanism was induced.
    \item \textbf{Reproducibility and drift.} Results can be sensitive to prompting, sampling, context windowing, and model/version changes, complicating replication and longitudinal comparability.
    \item \textbf{Automation bias.} Fluent outputs can overweight model-generated interpretations in analysts' judgment, reducing critical scrutiny and potentially narrowing alternative readings.
    \item \textbf{Data governance and confidentiality.} QR data often contain sensitive information; using external model services can introduce privacy, consent, and data-retention risks unless de-identification, secure deployment, and appropriate agreements are in place.
\end{enumerate}

Importantly, these risks are not limited to socially rich data. Even in relatively mechanistic settings (e.g., operational logs), LLMs can make epistemic errors in pattern recognition and in mapping evidence to structure, producing plausible but incorrect models. Therefore, we view LLMs as most appropriate for rapid prototyping and iterative sensemaking, rather than as an end-to-end replacement for expert analysis. Best practice requires explicit uncertainty, routine error checking, and iterative alignment between the tool's outputs and the researcher's analytic intent.

Moreover, the empirical distributions reported in this paper are derived from a curated sample of qualitative and computational works rather than from an exhaustive or statistically representative census of the field. 
Our collection process relies on keyword-based retrieval from scholarly indices, visibility in major venues, and manual inclusion decisions, which may systematically under-represent certain disciplines, methodological traditions, languages, or forms of qualitative practice (e.g., grey literature, monographs, or non-English venues). 
As a result, the observed skew toward lower-commitment meaning-making and modeling should be interpreted as descriptive of our sampled corpus, not as a population-level estimate of all qualitative research or all LLM-assisted QR efforts. 
Nevertheless, we argue that the concentration patterns observed are informative for diagnosing prevailing design tendencies in currently visible LLM-based tooling and for motivating the need for clearer representational commitments.

\section{Ethical Considerations}
This work addresses LLM-assisted qualitative data analysis, which is often applied to sensitive human,
organizational, or institutional data. 
Ethical risks primarily arise in downstream use: fluent model outputs may be mistaken for warranted interpretation, leading to misattribution of motives, norms, or responsibility, especially in high-stakes contexts.
Qualitative corpora frequently contain confidential information; responsible use requires appropriate consent, de-identification, and secure data handling. 
In addition, tools that scale qualitative modeling may be misused for surveillance or managerial control in asymmetric power settings.
We emphasize that LLM outputs should remain evidence-linked analytic aids rather than automated judgments,
with human accountability retained throughout.

We used LLM (GPT, Gemini) to improve the sentence and grammar for writing the abstract, introduction, analysis, and conclusion sections of the paper.
All text are mannually written and LLM only helps with word choice and grammar.

\bibliography{custom}

\begin{thebibliography}{169}
\providecommand{\natexlab}[1]{#1}

\bibitem[{Abbott(1995)}]{abbott1995sequence}
Andrew Abbott. 1995.
\newblock \href {https://doi.org/10.1146/annurev.so.21.080195.000521} {Sequence analysis: New methods for old ideas}.
\newblock \emph{Annual Review of Sociology}, 21:93--113.

\bibitem[{Abbott(2001)}]{abbott2001timematters}
Andrew Abbott. 2001.
\newblock \emph{Time Matters: On Theory and Method}.
\newblock University of Chicago Press, Chicago, IL.

\bibitem[{Agyeman(2014)}]{agyeman2014role}
Yaw~Boakye Agyeman. 2014.
\newblock The role of local knowledge in sustaining ecotourism livelihood as an adaptation to climate change.

\bibitem[{Anggraini and Alim(2023)}]{anggraini2023crypto}
Devi~Ferdina Anggraini and Mohammad~Nizarul Alim. 2023.
\newblock Crypto investment: Phenomenology study on investor behaviour.
\newblock In \emph{Proceeding International Conference on Economy, Management, and Business (Volume 1, 2023)}, volume~1, pages 307--317.

\bibitem[{Attride-Stirling(2001)}]{attridestirling2001thematic}
Jennifer Attride-Stirling. 2001.
\newblock \href {https://doi.org/10.1177/146879410100100307} {Thematic networks: An analytic tool for qualitative research}.
\newblock \emph{Qualitative Research}, 1(3):385--405.

\bibitem[{Attwell and Hannah(2022)}]{attwell2022convergence}
Katie Attwell and Adam Hannah. 2022.
\newblock Convergence on coercion: functional and political pressures as drivers of global childhood vaccine mandates.
\newblock \emph{International Journal of Health Policy and Management}, 11(11):2660.

\bibitem[{Austin(1962)}]{austin1962how}
J.~L. Austin. 1962.
\newblock \emph{How to Do Things with Words}.
\newblock Clarendon Press, Oxford.

\bibitem[{Baker et~al.(1998)Baker, Fillmore, and Lowe}]{baker1998framenet}
Collin~F. Baker, Charles~J. Fillmore, and John~B. Lowe. 1998.
\newblock \href {https://doi.org/10.3115/980845.980860} {The {B}erkeley {F}rame{N}et project}.
\newblock In \emph{Proceedings of the 36th Annual Meeting of the Association for Computational Linguistics and 17th International Conference on Computational Linguistics (Volume 1)}, pages 86--90, Montreal, Quebec, Canada. Association for Computational Linguistics.

\bibitem[{Barany et~al.(2024)Barany, Nasiar, Porter, Zambrano, Andres, Bright, Shah, Liu, Gao, Zhang, Mehta, Choi, Giordano, and Baker}]{Barany2024ChatGPTFE}
Amanda Barany, Nidhi Nasiar, Chelsea Porter, Andres~Felipe Zambrano, Juliana Ma. Alexandra~L. Andres, Dara Bright, Mamta Shah, Xiner Liu, Sabrina Gao, Jiayi Zhang, Shruti Mehta, Jaeyoon Choi, Camille Giordano, and Ryan~S. Baker. 2024.
\newblock \href {https://api.semanticscholar.org/CorpusID:271303020} {Chatgpt for education research: Exploring the potential of large language models for qualitative codebook development}.
\newblock In \emph{International Conference on Artificial Intelligence in Education}.

\bibitem[{Barke et~al.(2022)Barke, James, and Polikarpova}]{Barke2022GroundedCH}
Shraddha Barke, Michael~B. James, and Nadia Polikarpova. 2022.
\newblock \href {https://api.semanticscholar.org/CorpusID:250144196} {Grounded copilot: How programmers interact with code-generating models}.
\newblock \emph{Proceedings of the ACM on Programming Languages}, 7:85 -- 111.

\bibitem[{Barros et~al.(2024)Barros, Azevedo, Graciano~Neto, Kassab, Kalinowski, do~Nascimento, and Bandeira}]{barros2024llmqr}
Cau{\~a}~Ferreira Barros, Bruna~Borges Azevedo, Valdemar~Vicente Graciano~Neto, Mohamad Kassab, Marcos Kalinowski, Hugo Alexandre~D. do~Nascimento, and Michelle C. G. S.~P. Bandeira. 2024.
\newblock \href {https://arxiv.org/abs/2411.14473} {Large language model for qualitative research -- a systematic mapping study}.
\newblock \emph{Preprint}, arXiv:2411.14473.

\bibitem[{Barros et~al.(2025{\natexlab{a}})Barros, Azevedo, Neto, Kassab, Kalinowski, Do~Nascimento, and Bandeira}]{barros2025large}
Cau{\~a}~Ferreira Barros, Bruna~Borges Azevedo, Valdemar Vicente~Graciano Neto, Mohamad Kassab, Marcos Kalinowski, Hugo Alexandre~D Do~Nascimento, and Michelle~CGSP Bandeira. 2025{\natexlab{a}}.
\newblock Large language model for qualitative research: A systematic mapping study.
\newblock In \emph{2025 IEEE/ACM International Workshop on Methodological Issues with Empirical Studies in Software Engineering (WSESE)}, pages 48--55. IEEE.

\bibitem[{Barros et~al.(2025{\natexlab{b}})Barros, Azevedo, Neto, Kassab, Kalinowski, do~Nascimento, and Bandeira}]{barros2025largelanguagemodelqualitative}
Cauã~Ferreira Barros, Bruna~Borges Azevedo, Valdemar Vicente~Graciano Neto, Mohamad Kassab, Marcos Kalinowski, Hugo Alexandre~D. do~Nascimento, and Michelle C. G. S.~P. Bandeira. 2025{\natexlab{b}}.
\newblock \href {https://arxiv.org/abs/2411.14473} {Large language model for qualitative research -- a systematic mapping study}.
\newblock \emph{Preprint}, arXiv:2411.14473.

\bibitem[{Beattie et~al.(2004)Beattie, Fearnley, and Brandt}]{Beattie2004AGT}
Vivien~A. Beattie, Stella Fearnley, and Richard Brandt. 2004.
\newblock \href {https://api.semanticscholar.org/CorpusID:59405517} {A grounded theory model of auditor-client negotiations}.
\newblock \emph{Social Science Research Network}.

\bibitem[{Bekoulis et~al.(2020)Bekoulis, Papagiannopoulou, and Deligiannis}]{Bekoulis2020ARO}
Giannis Bekoulis, Christina Papagiannopoulou, and N.~Deligiannis. 2020.
\newblock \href {https://api.semanticscholar.org/CorpusID:237276903} {A review on fact extraction and verification}.
\newblock \emph{ACM Computing Surveys (CSUR)}, 55:1 -- 35.

\bibitem[{Bennett and Checkel(2015)}]{bennett2015processtracing}
Andrew Bennett and Jeffrey~T. Checkel, editors. 2015.
\newblock \href {https://doi.org/10.1017/CBO9781139858472} {\emph{Process Tracing: From Metaphor to Analytic Tool}}.
\newblock Cambridge University Press, Cambridge, UK.

\bibitem[{Berelson(1952)}]{berelson1952contentanalysis}
Bernard Berelson. 1952.
\newblock \emph{Content Analysis in Communication Research}.
\newblock Free Press, Glencoe, IL.

\bibitem[{Berg(2001)}]{berg2001qualitative}
Bruce~L Berg. 2001.
\newblock \emph{QUALITATIVE RESEARCH METHODS FOR THE SOCIAL SCIENCES 4th}.
\newblock Allyn and bacon.

\bibitem[{Biernacki and Waldorf(1981)}]{biernacki1981snowball}
Patrick Biernacki and Dan Waldorf. 1981.
\newblock Snowball sampling: Problems and techniques of chain referral sampling.
\newblock \emph{Sociological Methods \& Research}, 10(2):141--163.

\bibitem[{Binder and Edwards(2010)}]{Binder2010UsingGT}
Mario Binder and John~S. Edwards. 2010.
\newblock \href {https://api.semanticscholar.org/CorpusID:54764179} {Using grounded theory for theory building in operations management research: a study on inter-firm relationship governance}.
\newblock \emph{International Journal of Operations \& Production Management}, 30:232--259.

\bibitem[{Blumer(1954)}]{blumer1954wrongtheory}
Herbert Blumer. 1954.
\newblock What is wrong with social theory?
\newblock \emph{American Sociological Review}, 19(1):3--10.

\bibitem[{Bonabeau(2002)}]{bonabeau2002agent}
Eric Bonabeau. 2002.
\newblock \href {https://doi.org/10.1073/pnas.082080899} {Agent-based modeling: Methods and techniques for simulating human systems}.
\newblock \emph{Proceedings of the National Academy of Sciences}, 99(suppl\_3):7280--7287.

\bibitem[{Borgatti et~al.(2009)Borgatti, Mehra, Brass, and Labianca}]{borgatti2009network}
Stephen~P. Borgatti, Ajay Mehra, Daniel~J. Brass, and Giuseppe Labianca. 2009.
\newblock \href {https://doi.org/10.1126/science.1165821} {Network analysis in the social sciences}.
\newblock \emph{Science}, 323(5916):892--895.

\bibitem[{Bowman et~al.(2015)Bowman, Angeli, Potts, and Manning}]{bowman2015snli}
Samuel~R. Bowman, Gabor Angeli, Christopher Potts, and Christopher~D. Manning. 2015.
\newblock \href {https://doi.org/10.18653/v1/D15-1075} {A large annotated corpus for learning natural language inference}.
\newblock In \emph{Proceedings of the 2015 Conference on Empirical Methods in Natural Language Processing}, pages 632--642.

\bibitem[{Box(1976)}]{box1976science}
George E.~P. Box. 1976.
\newblock \href {https://doi.org/10.1080/01621459.1976.10480949} {Science and statistics}.
\newblock \emph{Journal of the American Statistical Association}, 71(356):791--799.

\bibitem[{Bratianu(2020)}]{Bratianu2020TowardUT}
Constantin Bratianu. 2020.
\newblock \href {https://api.semanticscholar.org/CorpusID:226223526} {Toward understanding the complexity of the covid-19 crisis: a grounded theory approach}.
\newblock \emph{Management \& Marketing. Challenges for the Knowledge Society}, 15:410 -- 423.

\bibitem[{Braun and Clarke(2006{\natexlab{a}})}]{braun2006thematic}
Virginia Braun and Victoria Clarke. 2006{\natexlab{a}}.
\newblock \href {https://doi.org/10.1191/1478088706qp063oa} {Using thematic analysis in psychology}.
\newblock \emph{Qualitative Research in Psychology}, 3(2):77--101.

\bibitem[{Braun and Clarke(2006{\natexlab{b}})}]{braun2006using}
Virginia Braun and Victoria Clarke. 2006{\natexlab{b}}.
\newblock Using thematic analysis in psychology.
\newblock \emph{Qualitative research in psychology}, 3(2):77--101.

\bibitem[{Braun and Clarke(2019)}]{braun2019reflecting}
Virginia Braun and Victoria Clarke. 2019.
\newblock Reflecting on reflexive thematic analysis.
\newblock \emph{Qualitative research in sport, exercise and health}, 11(4):589--597.

\bibitem[{Braun and Clarke(2021{\natexlab{a}})}]{braun2021can}
Virginia Braun and Victoria Clarke. 2021{\natexlab{a}}.
\newblock Can i use ta? should i use ta? should i not use ta? comparing reflexive thematic analysis and other pattern-based qualitative analytic approaches.
\newblock \emph{Counselling and psychotherapy research}, 21(1):37--47.

\bibitem[{Braun and Clarke(2021{\natexlab{b}})}]{braun2021one}
Virginia Braun and Victoria Clarke. 2021{\natexlab{b}}.
\newblock One size fits all? what counts as quality practice in (reflexive) thematic analysis?
\newblock \emph{Qualitative research in psychology}, 18(3):328--352.

\bibitem[{Brown and Yule(1983)}]{brown1983discourse}
Gillian Brown and George Yule. 1983.
\newblock \emph{Discourse analysis}.
\newblock Cambridge university press.

\bibitem[{Bryda and Costa(2023)}]{bryda2023qualitative}
Grzegorz Bryda and Ant{\'o}nio~Pedro Costa. 2023.
\newblock Qualitative research in digital era: innovations, methodologies and collaborations.
\newblock \emph{Social Sciences}, 12(10):570.

\bibitem[{Carius and Teixeira(2024)}]{Carius2024ArtificialIA}
Ana~Carolina Carius and Alex~Justen Teixeira. 2024.
\newblock \href {https://api.semanticscholar.org/CorpusID:270539944} {Artificial intelligence and content analysis: the large language models (llms) and the automatized categorization}.
\newblock \emph{AI \& SOCIETY}, 40:2405 -- 2416.

\bibitem[{Carlson(2012)}]{carlson2012dialogue}
Lauri Carlson. 2012.
\newblock \emph{Dialogue games: An approach to discourse analysis}, volume~17.
\newblock Springer Science \& Business Media.

\bibitem[{Cemri et~al.(2025{\natexlab{a}})Cemri, Pan, Yang, Agrawal, Chopra, Tiwari, Keutzer, Parameswaran, Klein, Ramchandran, Zaharia, Gonzalez, and Stoica}]{cemri2025mast}
Mert Cemri, Melissa~Z Pan, Shuyi Yang, Lakshya~A Agrawal, Bhavya Chopra, Rishabh Tiwari, Kurt Keutzer, Aditya Parameswaran, Dan Klein, Kannan Ramchandran, Matei Zaharia, Joseph~E Gonzalez, and Ion Stoica. 2025{\natexlab{a}}.
\newblock \href {https://openreview.net/forum?id=fAjbYBmonr} {Why do multi-agent {LLM} systems fail?}
\newblock In \emph{NeurIPS 2025 Datasets and Benchmarks Track}.

\bibitem[{Cemri et~al.(2025{\natexlab{b}})Cemri, Pan, Yang, Agrawal, Chopra, Tiwari, Keutzer, Parameswaran, Klein, Ramchandran, Zaharia, Gonzalez, and Stoica}]{cemri2025multiagentllmsystemsfail}
Mert Cemri, Melissa~Z. Pan, Shuyi Yang, Lakshya~A. Agrawal, Bhavya Chopra, Rishabh Tiwari, Kurt Keutzer, Aditya Parameswaran, Dan Klein, Kannan Ramchandran, Matei Zaharia, Joseph~E. Gonzalez, and Ion Stoica. 2025{\natexlab{b}}.
\newblock \href {https://arxiv.org/abs/2503.13657} {Why do multi-agent llm systems fail?}
\newblock \emph{Preprint}, arXiv:2503.13657.

\bibitem[{Charmaz(2014)}]{charmaz2014constructing}
Kathy Charmaz. 2014.
\newblock \emph{Constructing Grounded Theory}, 2 edition.
\newblock SAGE, London.

\bibitem[{Charmaz and Thornberg(2021)}]{charmaz2021pursuit}
Kathy Charmaz and Robert Thornberg. 2021.
\newblock The pursuit of quality in grounded theory.
\newblock \emph{Qualitative research in psychology}, 18(3):305--327.

\bibitem[{Chen et~al.(2016)Chen, Kocielnik, Drouhard, Pe{\~n}a-Araya, Suh, Cen, Zheng, Aragon, and Pe{\~n}a-Araya}]{chen2016challenges}
Nan-chen Chen, Rafal Kocielnik, Margaret Drouhard, Vanessa Pe{\~n}a-Araya, Jina Suh, Keting Cen, Xiangyi Zheng, Cecilia~R Aragon, and V~Pe{\~n}a-Araya. 2016.
\newblock Challenges of applying machine learning to qualitative coding.
\newblock In \emph{ACM SIGCHI Workshop on Human-Centered Machine Learning}.

\bibitem[{Cheng et~al.(2024)Cheng, Dong, Shi, Liu, Hu, O'Connor, Hauptmann, and Whitefoot}]{Cheng2024SHIELDLS}
Zhi-Qi Cheng, Yifei Dong, Aike Shi, Wei Liu, Yuzhi Hu, Jason O'Connor, Alexander~G. Hauptmann, and Kate Whitefoot. 2024.
\newblock \href {https://api.semanticscholar.org/CorpusID:271855011} {Shield: Llm-driven schema induction for predictive analytics in ev battery supply chain disruptions}.
\newblock In \emph{Conference on Empirical Methods in Natural Language Processing}.

\bibitem[{Chopra(2019)}]{Chopra2019IndianSM}
Komal Chopra. 2019.
\newblock \href {https://api.semanticscholar.org/CorpusID:169237740} {Indian shopper motivation to use artificial intelligence}.
\newblock \emph{International Journal of Retail \& Distribution Management}.

\bibitem[{Cleland(2011)}]{Cleland2011PredictionAE}
Carol~E. Cleland. 2011.
\newblock \href {https://api.semanticscholar.org/CorpusID:145565077} {Prediction and explanation in historical natural science}.
\newblock \emph{The British Journal for the Philosophy of Science}, 62:551 -- 582.

\bibitem[{Coleman(2017)}]{Coleman2017TechnologicalIL}
Donnie~Steve Coleman. 2017.
\newblock \href {https://api.semanticscholar.org/CorpusID:151759683} {Technological immersion learning: A grounded theory}.

\bibitem[{Cortazzi(1994)}]{cortazzi1994narrative}
Martin Cortazzi. 1994.
\newblock Narrative analysis.
\newblock \emph{Language teaching}, 27(3):157--170.

\bibitem[{Dagan et~al.(2006)Dagan, Glickman, and Magnini}]{dagan2006rte}
Ido Dagan, Oren Glickman, and Bernardo Magnini. 2006.
\newblock The {PASCAL} recognising textual entailment challenge.
\newblock In \emph{Machine Learning Challenges. Evaluating Predictive Uncertainty, Visual Object Classification, and Recognising Textual Entailment}, pages 177--190. Springer.

\bibitem[{Dai et~al.(2023)Dai, Xiong, and Ku}]{Dai2023LLMintheloopLL}
Shih-Chieh Dai, Aiping Xiong, and Lun-Wei Ku. 2023.
\newblock \href {https://api.semanticscholar.org/CorpusID:264436526} {Llm-in-the-loop: Leveraging large language model for thematic analysis}.
\newblock In \emph{Conference on Empirical Methods in Natural Language Processing}.

\bibitem[{Du et~al.(2022)Du, Zhang, Li, Wang, Yu, Wang, Lai, Lin, Liu, Zhou, Wen, Li, Hannan, Lei, Kim, Dror, Wang, Regan, Zeng, Lyu, Yu, Edwards, Jin, Jiao, Kazeminejad, Wang, Callison-Burch, Vondrick, Bansal, Roth, Han, Chang, Palmer, and Ji}]{Du2022RESIN11SE}
Xinya Du, Zixuan Zhang, Sha Li, Ziqi Wang, Pengfei Yu, Hongwei Wang, T.~Lai, Xudong Lin, Iris Liu, Ben Zhou, Haoyang Wen, Manling Li, Darryl Hannan, Jie Lei, Hyounghun Kim, Rotem Dror, Haoyu Wang, Michael Regan, Qi~Zeng, and 15 others. 2022.
\newblock \href {https://api.semanticscholar.org/CorpusID:249010869} {Resin-11: Schema-guided event prediction for 11 newsworthy scenarios}.
\newblock \emph{Proceedings of the 2022 Conference of the North American Chapter of the Association for Computational Linguistics: Human Language Technologies: System Demonstrations}.

\bibitem[{Edge et~al.(2024)Edge, Trinh, Cheng, Bradley, Chao, Mody, Truitt, and Larson}]{Edge2024FromLT}
Darren Edge, Ha~Trinh, Newman Cheng, Joshua Bradley, Alex Chao, Apurva~N. Mody, Steven Truitt, and Jonathan Larson. 2024.
\newblock \href {https://api.semanticscholar.org/CorpusID:269363075} {From local to global: A graph rag approach to query-focused summarization}.
\newblock \emph{ArXiv}, abs/2404.16130.

\bibitem[{Epstein(2006)}]{epstein2006generative}
Joshua~M. Epstein. 2006.
\newblock \emph{Generative Social Science: Studies in Agent-Based Computational Modeling}.
\newblock Princeton University Press, Princeton, NJ.

\bibitem[{Fairclough(2023)}]{fairclough2023critical}
Norman Fairclough. 2023.
\newblock Critical discourse analysis.
\newblock In \emph{The Routledge handbook of discourse analysis}, pages 11--22. Routledge.

\bibitem[{Fang et~al.(2025)Fang, Li, and Lu}]{Fang2025DecodingCI}
Hanming Fang, Ming Li, and Guangli Lu. 2025.
\newblock \href {https://api.semanticscholar.org/CorpusID:276597414} {Decoding china's industrial policies}.
\newblock \emph{SSRN Electronic Journal}.

\bibitem[{Farley(2021)}]{farley2021modern}
David Farley. 2021.
\newblock \emph{Modern Software Engineering: Doing what works to build better software faster}.
\newblock Addison-Wesley Professional.

\bibitem[{Fellahi(2021)}]{fellahi2021rural}
Anis Fellahi. 2021.
\newblock \emph{Rural youth outmigration choices in light of government development policies in Algeria}.
\newblock Ph.D. thesis, Universit{\"a}ts-und Landesbibliothek Bonn.

\bibitem[{Ferreira~Barros et~al.(2024)Ferreira~Barros, Borges~Azevedo, Graciano~Neto, Kassab, Kalinowski, do~Nascimento, and Bandeira}]{ferreira2024large}
Cau{\~a} Ferreira~Barros, Bruna Borges~Azevedo, Valdemar~Vicente Graciano~Neto, Mohamad Kassab, Marcos Kalinowski, Hugo Alexandre~D do~Nascimento, and Michelle~CGSP Bandeira. 2024.
\newblock Large language model for qualitative research--a systematic mapping study.
\newblock \emph{arXiv e-prints}, pages arXiv--2411.

\bibitem[{Fillmore(1982)}]{fillmore1982framesemantics}
Charles~J. Fillmore. 1982.
\newblock Frame semantics.
\newblock In \emph{Linguistics in the Morning Calm: Selected Papers from {SICOL-1981}}, pages 111--137. Hanshin Publishing Company, Seoul.

\bibitem[{Fletcher and Sarkar(2012)}]{Fletcher2012AGT}
David Fletcher and Mustafa Sarkar. 2012.
\newblock \href {https://api.semanticscholar.org/CorpusID:41307063} {A grounded theory of psychological resilience in olympic champions}.
\newblock \emph{Psychology of Sport and Exercise}, 13:669--678.

\bibitem[{Flick(2013)}]{flick2013sage}
Uwe Flick. 2013.
\newblock The sage handbook of qualitative data analysis.

\bibitem[{Forrester(1961)}]{forrester1961industrial}
Jay~W. Forrester. 1961.
\newblock \emph{Industrial Dynamics}.
\newblock MIT Press, Cambridge, MA.

\bibitem[{Gadamer(1989)}]{gadamer1989truthmethod}
Hans-Georg Gadamer. 1989.
\newblock \emph{Truth and Method}, 2 edition.
\newblock Continuum, New York, NY.

\bibitem[{Gao et~al.(2023)Gao, Choo, Cao, Lee, and Perrault}]{gao2023coaicoder}
Jie Gao, Kenny Tsu~Wei Choo, Junming Cao, Roy Ka-Wei Lee, and Simon Perrault. 2023.
\newblock Coaicoder: Examining the effectiveness of ai-assisted human-to-human collaboration in qualitative analysis.
\newblock \emph{ACM Transactions on Computer-Human Interaction}, 31(1):1--38.

\bibitem[{Gao et~al.(2025)Gao, Shu, and Yeo}]{Gao2025EfficiencyWR}
Jie Gao, Zoey Shu, and Shun~Yi Yeo. 2025.
\newblock \href {https://api.semanticscholar.org/CorpusID:275212618} {Efficiency with rigor! a trustworthy llm-powered workflow for qualitative data analysis}.

\bibitem[{Gebreegziabher et~al.(2023)Gebreegziabher, Zhang, Tang, Meng, Glassman, and Li}]{Gebreegziabher2023PaTATHC}
Simret~Araya Gebreegziabher, Zheng Zhang, Xiaohang Tang, Yihao Meng, Elena~L. Glassman, and Toby Jia-Jun Li. 2023.
\newblock \href {https://api.semanticscholar.org/CorpusID:258216675} {Patat: Human-ai collaborative qualitative coding with explainable interactive rule synthesis}.
\newblock \emph{Proceedings of the 2023 CHI Conference on Human Factors in Computing Systems}.

\bibitem[{Gee(2025)}]{gee2025introduction}
James~Paul Gee. 2025.
\newblock \emph{An introduction to discourse analysis: Theory and method}.
\newblock routledge.

\bibitem[{Geertz(1973)}]{geertz1973interpretation}
Clifford Geertz. 1973.
\newblock \emph{The Interpretation of Cultures: Selected Essays}.
\newblock Basic Books, New York, NY.

\bibitem[{George and Bennett(2005)}]{george2005case}
Alexander~L. George and Andrew Bennett. 2005.
\newblock \emph{Case Studies and Theory Development in the Social Sciences}.
\newblock MIT Press, Cambridge, MA.

\bibitem[{Gilbert and Troitzsch(2005)}]{gilbert2005simulation}
Nigel Gilbert and Klaus~G. Troitzsch. 2005.
\newblock \emph{Simulation for the Social Scientist}, 2 edition.
\newblock Open University Press, Maidenhead, UK.

\bibitem[{Gioia et~al.(2013)Gioia, Corley, and Hamilton}]{gioia2013rigor}
Dennis~A. Gioia, Kevin~G. Corley, and Aimee~L. Hamilton. 2013.
\newblock \href {https://doi.org/10.1177/1094428112452151} {Seeking qualitative rigor in inductive research: Notes on the {G}ioia methodology}.
\newblock \emph{Organizational Research Methods}, 16(1):15--31.

\bibitem[{Glaser and Strauss(1967)}]{glaser1967discovery}
Barney~G. Glaser and Anselm~L. Strauss. 1967.
\newblock \emph{The Discovery of Grounded Theory: Strategies for Qualitative Research}.
\newblock Sociology Press, Mill Valley, CA.

\bibitem[{Goffman(1974)}]{goffman1974frameanalysis}
Erving Goffman. 1974.
\newblock \emph{Frame Analysis: An Essay on the Organization of Experience}.
\newblock Harper \& Row, New York, NY.

\bibitem[{Goldsmith(2021)}]{goldsmith2021using}
Laurie~J Goldsmith. 2021.
\newblock Using framework analysis in applied qualitative research.
\newblock \emph{Qualitative report}, 26(6).

\bibitem[{Goulet et~al.(2023)Goulet, Lessard-Desch{\^e}nes, Pariseau-Legault, Breton, and Crocker}]{Goulet2023CommunityTO}
Marie-H{\'e}l{\`e}ne Goulet, Clara Lessard-Desch{\^e}nes, Pierre Pariseau-Legault, Richard Breton, and Anne~G. Crocker. 2023.
\newblock \href {https://api.semanticscholar.org/CorpusID:259130456} {Community treatment orders: A qualitative study of stakeholder perspectives.}
\newblock \emph{International journal of law and psychiatry}, 89:101901.

\bibitem[{Graneheim et~al.(2017)Graneheim, Lindgren, and Lundman}]{graneheim2017challenges}
Ulla~H. Graneheim, Britt-Marie Lindgren, and Berit Lundman. 2017.
\newblock \href {https://doi.org/10.1016/j.nedt.2017.06.002} {Methodological challenges in qualitative content analysis: A discussion paper}.
\newblock \emph{Nurse Education Today}, 56:29--34.

\bibitem[{Graneheim and Lundman(2004)}]{graneheim2004trustworthiness}
Ulla~H. Graneheim and Berit Lundman. 2004.
\newblock \href {https://doi.org/10.1016/j.nedt.2003.10.001} {Qualitative content analysis in nursing research: Concepts, procedures and measures to achieve trustworthiness}.
\newblock \emph{Nurse Education Today}, 24(2):105--112.

\bibitem[{Grice(1975)}]{grice1975logic}
H.~Paul Grice. 1975.
\newblock Logic and conversation.
\newblock In Peter Cole and Jerry~L. Morgan, editors, \emph{Syntax and Semantics, Volume 3: Speech Acts}, pages 41--58. Academic Press, New York, NY.

\bibitem[{Hammersley(2006)}]{hammersley2006ethnography}
Martyn Hammersley. 2006.
\newblock Ethnography: problems and prospects.
\newblock \emph{Ethnography and education}, 1(1):3--14.

\bibitem[{Hedstr{\"o}m and Swedberg(1998)}]{hedstrom1998socialmechanisms}
Peter Hedstr{\"o}m and Richard Swedberg, editors. 1998.
\newblock \emph{Social Mechanisms: An Analytical Approach to Social Theory}.
\newblock Cambridge University Press, Cambridge, UK.

\bibitem[{Holsti(1969)}]{holsti1969contentanalysis}
Ole~R. Holsti. 1969.
\newblock \emph{Content Analysis for the Social Sciences and Humanities}.
\newblock Addison-Wesley, Reading, MA.

\bibitem[{Hong et~al.(2023)Hong, Zhuge, Chen, Zheng, Cheng, Zhang, Wang, Wang, Yau, Lin, Zhou, Ran, Xiao, Wu, and Schmidhuber}]{hong2023metagpt}
Sirui Hong, Mingchen Zhuge, Jiaqi Chen, Xiawu Zheng, Yuheng Cheng, Ceyao Zhang, Jinlin Wang, Zili Wang, Steven Ka~Shing Yau, Zijuan Lin, Liyang Zhou, Chenyu Ran, Lingfeng Xiao, Chenglin Wu, and J{\"u}rgen Schmidhuber. 2023.
\newblock \href {https://arxiv.org/abs/2308.00352} {Metagpt: Meta programming for a multi-agent collaborative framework}.
\newblock \emph{Preprint}, arXiv:2308.00352.

\bibitem[{Hou et~al.(2024)Hou, Zhu, Zheng, Zhang, Huang, Zhong, Li, Du, and Ker}]{Hou2024PromptbasedAF}
Chenyu Hou, Gaoxia Zhu, Juan Zheng, Lishan Zhang, Xiaoshan Huang, Tianlong Zhong, Shan Li, Hanxiang Du, and Chin~Lee Ker. 2024.
\newblock \href {https://api.semanticscholar.org/CorpusID:268733742} {Prompt-based and fine-tuned gpt models for context-dependent and -independent deductive coding in social annotation}.
\newblock \emph{Proceedings of the 14th Learning Analytics and Knowledge Conference}.

\bibitem[{Hsieh and Shannon(2005)}]{hsieh2005threeapproaches}
Hsiu-Fang Hsieh and Sarah~E. Shannon. 2005.
\newblock \href {https://doi.org/10.1177/1049732305276687} {Three approaches to qualitative content analysis}.
\newblock \emph{Qualitative Health Research}, 15(9):1277--1288.

\bibitem[{Huang et~al.(2025)Huang, Reyna, Lerner, Xia, and Hempel}]{huang2025professional}
Ruanqianqian Huang, Avery Reyna, Sorin Lerner, Haijun Xia, and Brian Hempel. 2025.
\newblock Professional software developers don't vibe, they control: Ai agent use for coding in 2025.
\newblock \emph{arXiv preprint arXiv:2512.14012}.

\bibitem[{J{\"a}ppinen et~al.(2022)J{\"a}ppinen, Muurinen, and K{\"a}{\"a}ri{\"a}inen}]{jappinen2022theory}
Maija J{\"a}ppinen, Heidi Muurinen, and Aino K{\"a}{\"a}ri{\"a}inen. 2022.
\newblock Theory-driven supervision as a method of strengthening the emerging professional identity of social work students.
\newblock \emph{Relational Social Work}, 6(1):3--18.

\bibitem[{Jin et~al.(2022)Jin, Li, and Ji}]{jin-etal-2022-event}
Xiaomeng Jin, Manling Li, and Heng Ji. 2022.
\newblock Event schema induction with double graph autoencoders.
\newblock In \emph{Proceedings of the 2022 Conference of NAACL}, pages 2013--2025, Seattle, United States. Association for Computational Linguistics.

\bibitem[{Jowsey et~al.(2025)Jowsey, Braun, Clarke, Lupton, and Fine}]{reject-reflexivea-qr}
Tanisha Jowsey, Virginia Braun, Victoria Clarke, Deborah Lupton, and Michelle Fine. 2025.
\newblock \href {https://doi.org/10.1177/10778004251401851} {We reject the use of generative artificial intelligence for reflexive qualitative research}.
\newblock \emph{Qualitative Inquiry}, page 10778004251401851.

\bibitem[{Kang et~al.(2025)Kang, Han, Tian, Zhang, and Rzeszotarski}]{themeViz}
Daye Kang, Zhuolun Han, Jiahe Tian, Muhan Zhang, and Jeffrey~M Rzeszotarski. 2025.
\newblock \href {https://doi.org/10.1145/3757675} {Themeviz: Understanding the effect of human-ai collaboration in theme development with an llm-enhanced interactive visual system}.
\newblock 9(7).

\bibitem[{Karakas and Sarigollu(2019)}]{karakas2019spirals}
Fahri Karakas and Emine Sarigollu. 2019.
\newblock Spirals of spirituality: A qualitative study exploring dynamic patterns of spirituality in turkish organizations.
\newblock \emph{Journal of business ethics}, 156(3):799--821.

\bibitem[{Keith et~al.(2017)Keith, Handler, Pinkham, Magliozzi, McDuffie, and O'Connor}]{Keith2017IdentifyingCK}
Katherine~A. Keith, Abram Handler, Michael Pinkham, Cara Magliozzi, J.~A. McDuffie, and Brendan~T. O'Connor. 2017.
\newblock \href {https://api.semanticscholar.org/CorpusID:22248025} {Identifying civilians killed by police with distantly supervised entity-event extraction}.
\newblock \emph{ArXiv}, abs/1707.07086.

\bibitem[{Kendellen and Camir{\'e}(2019)}]{Kendellen2019ApplyingIL}
Kelsey Kendellen and Martin Camir{\'e}. 2019.
\newblock \href {https://api.semanticscholar.org/CorpusID:149624460} {Applying in life the skills learned in sport: A grounded theory}.
\newblock \emph{Psychology of Sport and Exercise}, 40:23–32.

\bibitem[{Khan et~al.(2024)Khan, Kegalle, D'Silva, Watt, Whelan-Shamy, Ghahremanlou, and Magee}]{khan2024automating}
Awais~Hameed Khan, Hiruni Kegalle, Rhea D'Silva, Ned Watt, Daniel Whelan-Shamy, Lida Ghahremanlou, and Liam Magee. 2024.
\newblock Automating thematic analysis: how llms analyse controversial topics.
\newblock \emph{arXiv preprint arXiv:2405.06919}.

\bibitem[{Kleinheksel et~al.(2020)Kleinheksel, Rockich-Winston, Tawfik, and Wyatt}]{kleinheksel2020demystifying}
A.~J. Kleinheksel, Nicole Rockich-Winston, Huda Tawfik, and Tasha~R. Wyatt. 2020.
\newblock \href {https://doi.org/10.5688/ajpe7113} {Demystifying content analysis}.
\newblock \emph{American Journal of Pharmaceutical Education}, 84(1):7113.

\bibitem[{Kosko(1986)}]{kosko1986fuzzy}
Bart Kosko. 1986.
\newblock \href {https://doi.org/10.1016/S0020-7373(86)80040-2} {Fuzzy cognitive maps}.
\newblock \emph{International Journal of Man-Machine Studies}, 24(1):65--75.

\bibitem[{Krippendorff(2018)}]{krippendorff2018contentanalysis}
Klaus Krippendorff. 2018.
\newblock \href {https://doi.org/10.4135/9781071878781} {\emph{Content Analysis: An Introduction to Its Methodology}}, 4 edition.
\newblock SAGE Publications, Thousand Oaks, CA.

\bibitem[{Kristelstein-H{\"a}nninen(2022)}]{kristelstein2022brain}
Ket Kristelstein-H{\"a}nninen. 2022.
\newblock The brain waste phenomenon in finland: factors preventing inclusion of highly educated women with immigrant background into finnish workforce.

\bibitem[{Lam et~al.(2024)Lam, Teoh, Landay, Heer, and Bernstein}]{Lam2024ConceptIA}
Michelle~S. Lam, Janice Teoh, James~A. Landay, Jeffrey Heer, and Michael~S. Bernstein. 2024.
\newblock \href {https://api.semanticscholar.org/CorpusID:269214633} {Concept induction: Analyzing unstructured text with high-level concepts using lloom}.
\newblock \emph{Proceedings of the 2024 CHI Conference on Human Factors in Computing Systems}.

\bibitem[{Langley(1999)}]{langley1999strategies}
Ann Langley. 1999.
\newblock \href {https://doi.org/10.5465/amr.1999.2553248} {Strategies for theorizing from process data}.
\newblock \emph{Academy of Management Review}, 24(4):691--710.

\bibitem[{Law(2015)}]{law2015simulation}
Averill~M. Law. 2015.
\newblock \emph{Simulation Modeling and Analysis}, 5 edition.
\newblock McGraw-Hill Education, New York, NY.

\bibitem[{Li et~al.(2018)Li, Sun, Han, and Li}]{Li2018ASO}
J.~Li, Aixin Sun, Jianglei Han, and Chenliang Li. 2018.
\newblock \href {https://api.semanticscholar.org/CorpusID:56895382} {A survey on deep learning for named entity recognition}.
\newblock \emph{IEEE Transactions on Knowledge and Data Engineering}, 34:50--70.

\bibitem[{Li et~al.(2022{\natexlab{a}})Li, Li, Wang, Huang, Cho, Ji, Han, and Voss}]{li2022futureonedimensionalcomplexevent}
Manling Li, Sha Li, Zhenhailong Wang, Lifu Huang, Kyunghyun Cho, Heng Ji, Jiawei Han, and Clare Voss. 2022{\natexlab{a}}.
\newblock \href {https://arxiv.org/abs/2104.06344} {The future is not one-dimensional: Complex event schema induction by graph modeling for event prediction}.
\newblock \emph{Preprint}, arXiv:2104.06344.

\bibitem[{Li et~al.(2022{\natexlab{b}})Li, Reddy, Wang, Chiang, Lai, Yu, Zhang, and Ji}]{Li2022COVID19CR}
Manling Li, Revanth~Gangi Reddy, Ziqi Wang, Yi-Shyuan Chiang, T.~Lai, Pengfei Yu, Zixuan Zhang, and Heng Ji. 2022{\natexlab{b}}.
\newblock \href {https://api.semanticscholar.org/CorpusID:247127078} {Covid-19 claim radar: A structured claim extraction and tracking system}.
\newblock In \emph{Annual Meeting of the Association for Computational Linguistics}.

\bibitem[{Li et~al.(2020)Li, Zeng, Lin, Cho, Ji, May, Chambers, and Voss}]{Li2020ConnectingTD}
Manling Li, Qi~Zeng, Ying Lin, Kyunghyun Cho, Heng Ji, Jonathan May, Nathanael Chambers, and Clare~R. Voss. 2020.
\newblock \href {https://api.semanticscholar.org/CorpusID:226262197} {Connecting the dots: Event graph schema induction with path language modeling}.
\newblock In \emph{Conference on Empirical Methods in Natural Language Processing}.

\bibitem[{Li et~al.(2023)Li, Zhao, Zhao, Li, Ji, Callison-Burch, and Han}]{Li2023OpenDomainHE}
Sha Li, Ruining Zhao, Rui Zhao, Manling Li, Heng Ji, Chris Callison-Burch, and Jiawei Han. 2023.
\newblock \href {https://api.semanticscholar.org/CorpusID:259342736} {Open-domain hierarchical event schema induction by incremental prompting and verification}.
\newblock In \emph{Annual Meeting of the Association for Computational Linguistics}.

\bibitem[{Lincoln and Guba(1985)}]{lincoln1985naturalistic}
Yvonna~S. Lincoln and Egon~G. Guba. 1985.
\newblock \emph{Naturalistic Inquiry}.
\newblock SAGE Publications, Newbury Park, CA.

\bibitem[{Loosemore(1999)}]{Loosemore1999AGT}
Martin Loosemore. 1999.
\newblock \href {https://api.semanticscholar.org/CorpusID:108504714} {A grounded theory of construction crisis management}.
\newblock \emph{Construction Management and Economics}, 17:9--19.

\bibitem[{Lunney(2025)}]{lunney2025resilience}
Edward Lunney. 2025.
\newblock \emph{RESILIENCE AMIDST CHAOS: THE AUTOETHNOGRAPHIC JOURNEY OF A DOCTORAL PRACTITIONER NAVIGATING EXTREME UNCERTAINTY}.
\newblock Ph.D. thesis, University of Liverpool.

\bibitem[{Macy and Willer(2002)}]{macywiller2002actors}
Michael~W. Macy and Robert Willer. 2002.
\newblock \href {https://doi.org/10.1146/annurev.soc.28.110601.141117} {From factors to actors: Computational sociology and agent-based modeling}.
\newblock \emph{Annual Review of Sociology}, 28:143--166.

\bibitem[{Manuj and Pohlen(2012)}]{manuj2012reviewer}
Ila Manuj and Terrance~L Pohlen. 2012.
\newblock A reviewer's guide to the grounded theory methodology in logistics and supply chain management research.
\newblock \emph{International Journal of Physical Distribution \& Logistics Management}, 42(8/9):784--803.

\bibitem[{McKeown et~al.(2015)McKeown, Roy, and Spandler}]{McKeown2015YoullNW}
Mick McKeown, Alastair Roy, and Helen Spandler. 2015.
\newblock \href {https://api.semanticscholar.org/CorpusID:29766103} {'you'll never walk alone': Supportive social relations in a football and mental health project.}
\newblock \emph{International journal of mental health nursing}, 24 4:360--9.

\bibitem[{Meadows(2008)}]{meadows2008thinking}
Donella~H. Meadows. 2008.
\newblock \emph{Thinking in Systems: A Primer}.
\newblock Chelsea Green Publishing, White River Junction, VT.

\bibitem[{Miles et~al.(2014)Miles, Huberman, and Salda{\~n}a}]{miles2014qualitative}
Matthew~B. Miles, A.~Michael Huberman, and Johnny Salda{\~n}a. 2014.
\newblock \emph{Qualitative Data Analysis: A Methods Sourcebook}, 3 edition.
\newblock SAGE, Thousand Oaks, CA.

\bibitem[{Montes et~al.(2025)Montes, Feldt, Martos, Ouhbi, Premanandan, and Graziotin}]{Montes2025LargeLM}
Cristina~Martinez Montes, Robert Feldt, Cristina~Miguel Martos, Sofia Ouhbi, Shweta Premanandan, and Daniel Graziotin. 2025.
\newblock \href {https://api.semanticscholar.org/CorpusID:282246500} {Large language models in thematic analysis: Prompt engineering, evaluation, and guidelines for qualitative software engineering research}.
\newblock \emph{ArXiv}, abs/2510.18456.

\bibitem[{Nasar et~al.(2021)Nasar, Jaffry, and Malik}]{article}
Zara Nasar, Syed~Waqar Jaffry, and Muhammad Malik. 2021.
\newblock \href {https://doi.org/10.1145/3445965} {Named entity recognition and relation extraction: State of the art}.
\newblock \emph{ACM Computing Surveys}, 54.

\bibitem[{Novak and Cañas(2008)}]{novak2008conceptmaps}
Joseph~D. Novak and Alberto~J. Cañas. 2008.
\newblock The theory underlying concept maps and how to construct and use them.
\newblock Technical Report IHMC CmapTools 2006-01 Rev 01-2008, Florida Institute for Human and Machine Cognition.

\bibitem[{Nowell et~al.(2017)Nowell, Norris, White, and Moules}]{nowell2017thematic}
Lorelli~S Nowell, Jill~M Norris, Deborah~E White, and Nancy~J Moules. 2017.
\newblock Thematic analysis: Striving to meet the trustworthiness criteria.
\newblock \emph{International journal of qualitative methods}, 16(1):1609406917733847.

\bibitem[{Nyaaba et~al.(2025)Nyaaba, SungEun, Apam, Acheampong, and Dwamena}]{Nyaaba2025OptimizingGA}
Matthew Nyaaba, Min SungEun, Mary~Abiswin Apam, Kwame~O. Acheampong, and Emmanuel Dwamena. 2025.
\newblock \href {https://api.semanticscholar.org/CorpusID:277244800} {Optimizing generative ai's accuracy and transparency in inductive thematic analysis: A human-ai comparison}.
\newblock \emph{ArXiv}, abs/2503.16485.

\bibitem[{{OpenAI}(2023)}]{openai2023gpt4}
{OpenAI}. 2023.
\newblock \href {https://arxiv.org/abs/2303.08774} {{GPT}-4 technical report}.
\newblock \emph{Preprint}, arXiv:2303.08774.

\bibitem[{Padmakumar et~al.(2025)Padmakumar, Chang, Lo, Downey, and Naik}]{padmakumar2025intent}
Vishakh Padmakumar, Joseph~Chee Chang, Kyle Lo, Doug Downey, and Aakanksha Naik. 2025.
\newblock Intent-aware schema generation and refinement for literature review tables.
\newblock \emph{Findings of the Association for Computational Linguistics: EMNLP 2025}, pages 23450--23472.

\bibitem[{Pan et~al.(2025)Pan, Arabzadeh, Cogo, Zhu, Xiong, Agrawal, Mao, Shen, Pallerla, Patel et~al.}]{pan2025measuring}
Melissa~Z Pan, Negar Arabzadeh, Riccardo Cogo, Yuxuan Zhu, Alexander Xiong, Lakshya~A Agrawal, Huanzhi Mao, Emma Shen, Sid Pallerla, Liana Patel, and 1 others. 2025.
\newblock Measuring agents in production.
\newblock \emph{arXiv preprint arXiv:2512.04123}.

\bibitem[{Paoli(2024)}]{performing}
Stefano~De Paoli. 2024.
\newblock Performing an inductive thematic analysis of semi-structured interviews with a large language model: An exploration and provocation on the limits of the approach.
\newblock \emph{Social Science Computer Review}, 42(4):997--1019.

\bibitem[{Parfenova et~al.(2024)Parfenova, Denzler, and Pfeffer}]{parfenova-etal-2024-automating}
Angelina Parfenova, Alexander Denzler, and J{\"o}rgen Pfeffer. 2024.
\newblock \href {https://doi.org/10.18653/v1/2024.acl-srw.17} {Automating qualitative data analysis with large language models}.
\newblock In \emph{Proceedings of the 62nd Annual Meeting of the Association for Computational Linguistics (Volume 4: Student Research Workshop)}, pages 83--91, Bangkok, Thailand. Association for Computational Linguistics.

\bibitem[{Parfenova et~al.(2025)Parfenova, Marfurt, Pfeffer, and Denzler}]{parfenova2025text}
Angelina Parfenova, Andreas Marfurt, J{\"u}rgen Pfeffer, and Alexander Denzler. 2025.
\newblock Text annotation via inductive coding: Comparing human experts to llms in qualitative data analysis.
\newblock In \emph{Findings of the Association for Computational Linguistics: NAACL 2025}, pages 6456--6469.

\bibitem[{Parkinson et~al.(2016)Parkinson, Eatough, Holmes, Stapley, and Midgley}]{parkinson2016framework}
Sally Parkinson, Virginia Eatough, Joshua Holmes, Emily Stapley, and Nick Midgley. 2016.
\newblock Framework analysis: a worked example of a study exploring young people’s experiences of depression.
\newblock \emph{Qualitative research in psychology}, 13(2):109--129.

\bibitem[{Pearl(2009)}]{pearl2009causality}
Judea Pearl. 2009.
\newblock \emph{Causality: Models, Reasoning, and Inference}, 2 edition.
\newblock Cambridge University Press, Cambridge, UK.

\bibitem[{Pecoraro and Uusitalo(2014)}]{pecoraro2014conflicting}
Maria~Grazia Pecoraro and Outi Uusitalo. 2014.
\newblock Conflicting values of ethical consumption in diverse worlds--a cultural approach.
\newblock \emph{Journal of Consumer Culture}, 14(1):45--65.

\bibitem[{Pettigrew(1990)}]{pettigrew1990longitudinal}
Andrew~M. Pettigrew. 1990.
\newblock \href {https://doi.org/10.1287/orsc.1.3.267} {Longitudinal field research on change: Theory and practice}.
\newblock \emph{Organization Science}, 1(3):267--292.

\bibitem[{Pi et~al.(2025)Pi, Yang, and Nguyen}]{pi2025logos}
Xinyu Pi, Qisen Yang, and Chuong Nguyen. 2025.
\newblock \href {https://arxiv.org/abs/2509.24294} {Logos: {LLM}-driven end-to-end grounded theory development and schema induction for qualitative research}.
\newblock \emph{Preprint}, arXiv:2509.24294.

\bibitem[{Pike(1954)}]{pike1954language}
Kenneth~L. Pike. 1954.
\newblock \emph{Language in Relation to a Unified Theory of the Structure of Human Behavior}.
\newblock Summer Institute of Linguistics, Glendale, CA.

\bibitem[{Qian et~al.(2024)Qian, Liu, Liu, Chen, Dang, Li, Yang, Chen, Su, Cong, Xu, Li, Liu, and Sun}]{qian-etal-2024-chatdev}
Chen Qian, Wei Liu, Hongzhang Liu, Nuo Chen, Yufan Dang, Jiahao Li, Cheng Yang, Weize Chen, Yusheng Su, Xin Cong, Juyuan Xu, Dahai Li, Zhiyuan Liu, and Maosong Sun. 2024.
\newblock \href {https://doi.org/10.18653/v1/2024.acl-long.810} {{C}hat{D}ev: Communicative agents for software development}.
\newblock In \emph{Proceedings of the 62nd Annual Meeting of the Association for Computational Linguistics (Volume 1: Long Papers)}, pages 15174--15186, Bangkok, Thailand. Association for Computational Linguistics.

\bibitem[{Qiao et~al.(2025)Qiao, Walker, Cunningham, and Koh}]{Qiao2025ThematicLMAL}
Tingrui Qiao, Caroline Walker, Chris Cunningham, and Yun~Sing Koh. 2025.
\newblock Thematic-lm: A llm-based multi-agent system for large-scale thematic analysis.
\newblock \emph{Proceedings of the ACM on Web Conference 2025}.

\bibitem[{Ragin(1987)}]{ragin1987comparative}
Charles~C. Ragin. 1987.
\newblock \emph{The Comparative Method: Moving Beyond Qualitative and Quantitative Strategies}.
\newblock University of California Press, Berkeley, CA.

\bibitem[{Rao et~al.(2024)Rao, Agarwal, Dalal, Calacci, and Monroy-Hern'andez}]{Rao2024QuaLLMAL}
Varun~Nagaraj Rao, Eesha Agarwal, Samantha Dalal, Dan Calacci, and Andr'es Monroy-Hern'andez. 2024.
\newblock \href {https://api.semanticscholar.org/CorpusID:269635588} {Quallm: An llm-based framework to extract quantitative insights from online forums}.
\newblock \emph{ArXiv}, abs/2405.05345.

\bibitem[{Rempel et~al.(2013)Rempel, Ravindran, Rogers, and Magill-Evans}]{Rempel2013ParentingUP}
Gwen~R. Rempel, Vinitha~Paul Ravindran, Laura~G. Rogers, and Joyce Magill-Evans. 2013.
\newblock \href {https://api.semanticscholar.org/CorpusID:28344321} {Parenting under pressure: a grounded theory of parenting young children with life-threatening congenital heart disease.}
\newblock \emph{Journal of advanced nursing}, 69 3:619--30.

\bibitem[{Ricoeur(1976)}]{ricoeur1976interpretationtheory}
Paul Ricoeur. 1976.
\newblock \emph{Interpretation Theory: Discourse and the Surplus of Meaning}.
\newblock Texas Christian University Press, Fort Worth, TX.

\bibitem[{Rietz and Maedche(2021)}]{Rietz2021CodyAA}
Tim Rietz and Alexander Maedche. 2021.
\newblock \href {https://api.semanticscholar.org/CorpusID:233987399} {Cody: An ai-based system to semi-automate coding for qualitative research}.
\newblock \emph{Proceedings of the 2021 CHI Conference on Human Factors in Computing Systems}.

\bibitem[{Ritchie et~al.(2013)Ritchie, Ormston, McNaughton~Nicholls, and Lewis}]{ritchie2013qualitative}
Jane Ritchie, Rachel Ormston, Carol McNaughton~Nicholls, and Jane Lewis. 2013.
\newblock Qualitative research practice: A guide for social science students and researchers.

\bibitem[{Safa et~al.(2024)Safa, Adib-Hajbaghery, and Rezaei}]{Safa2024AMF}
Azade Safa, Mohsen Adib-Hajbaghery, and Mahboubeh Rezaei. 2024.
\newblock \href {https://api.semanticscholar.org/CorpusID:267778836} {A model for older adults’ coping with the death of their child: a grounded theory study}.
\newblock \emph{BMC Psychiatry}, 24.

\bibitem[{Salda{\~n}a(2021)}]{saldana2021coding}
Johnny Salda{\~n}a. 2021.
\newblock \emph{The Coding Manual for Qualitative Researchers}.
\newblock SAGE, Thousand Oaks, CA.

\bibitem[{Schroeder et~al.(2025)Schroeder, Aubin Le~Qu{\'e}r{\'e}, Randazzo, Mimno, and Schoenebeck}]{schroeder2025llmqr}
Hope Schroeder, Marianne Aubin Le~Qu{\'e}r{\'e}, Casey Randazzo, David Mimno, and Sarita Schoenebeck. 2025.
\newblock \href {https://doi.org/10.1145/3706598.3713120} {Large language models in qualitative research: Uses, tensions, and intentions}.
\newblock In \emph{Proceedings of the CHI Conference on Human Factors in Computing Systems (CHI 2025)}.

\bibitem[{Searle(1969)}]{searle1969speechacts}
John~R. Searle. 1969.
\newblock \emph{Speech Acts: An Essay in the Philosophy of Language}.
\newblock Cambridge University Press, Cambridge.

\bibitem[{Serkina and Logvinova(2019)}]{serkina2019administrative}
Yana~I Serkina and Anastasia~V Logvinova. 2019.
\newblock Administrative management of universities: Background and consequences.
\newblock \emph{Amazonia Investiga}, 8(22):673--683.

\bibitem[{Sharma and Wallace(2025)}]{sharma2025details}
Ansh Sharma and James~R Wallace. 2025.
\newblock Details: Deep thematic analysis with iterative llm support.
\newblock In \emph{Proceedings of the 7th ACM Conference on Conversational User Interfaces}, pages 1--7.

\bibitem[{Singh et~al.(2025)Singh, Chang, Anastasiades, Haddad, Naik, Tanaka, Zamarron, Nguyen, Hwang, Dunkleberger, Latzke, Rao, Lochner, Evans, Kinney, Weld, Downey, and Feldman}]{Singh2025Ai2SQ}
Amanpreet Singh, Joseph~Chee Chang, Chloe Anastasiades, Dany Haddad, Aakanksha Naik, Amber Tanaka, Angele Zamarron, Cecile Nguyen, Jena~D. Hwang, Jason Dunkleberger, Matt Latzke, Smita Rao, Jaron Lochner, Rob Evans, Rodney Kinney, Daniel~S. Weld, Doug Downey, and Sergey Feldman. 2025.
\newblock \href {https://api.semanticscholar.org/CorpusID:277786810} {Ai2 scholar qa: Organized literature synthesis with attribution}.
\newblock \emph{ArXiv}, abs/2504.10861.

\bibitem[{Skryabina et~al.(2016)Skryabina, Morris, Byrne, Harkin, Rook, and Stallard}]{skryabina2016child}
Elena Skryabina, Joanna Morris, Danielle Byrne, Nicola Harkin, Sarah Rook, and Paul Stallard. 2016.
\newblock Child, teacher and parent perceptions of the friends classroom-based universal anxiety prevention programme: A qualitative study.
\newblock \emph{School Mental Health}, 8(4):486--498.

\bibitem[{Smith et~al.(2009)Smith, Flowers, and Larkin}]{smith2009ipa}
Jonathan~A. Smith, Paul Flowers, and Michael Larkin. 2009.
\newblock \emph{Interpretative Phenomenological Analysis: Theory, Method and Research}.
\newblock SAGE, Thousand Oaks, CA.

\bibitem[{Spirtes et~al.(2000)Spirtes, Glymour, and Scheines}]{spirtes2000cps}
Peter Spirtes, Clark Glymour, and Richard Scheines. 2000.
\newblock \emph{Causation, Prediction, and Search}, 2 edition.
\newblock MIT Press, Cambridge, MA.

\bibitem[{Stalnaker(2002)}]{stalnaker2002commonground}
Robert Stalnaker. 2002.
\newblock \href {https://doi.org/10.1023/A:1020867916902} {Common ground}.
\newblock \emph{Linguistics and Philosophy}, 25(5):701--721.

\bibitem[{Sterman(2000)}]{sterman2000business}
John~D. Sterman. 2000.
\newblock \emph{Business Dynamics: Systems Thinking and Modeling for a Complex World}.
\newblock Irwin/McGraw-Hill, Boston.

\bibitem[{Sun et~al.(2020)Sun, Wu, and Yin}]{Sun2020GreenIR}
Yingying Sun, Lei-Yu Wu, and Shi Yin. 2020.
\newblock \href {https://api.semanticscholar.org/CorpusID:230549247} {Green innovation risk identification of the manufacturing industry under global value chain based on grounded theory}.
\newblock \emph{Sustainability}.

\bibitem[{Szajnfarber and Gralla(2017)}]{szajnfarber2017qualitative}
Zoe Szajnfarber and Erica Gralla. 2017.
\newblock Qualitative methods for engineering systems: Why we need them and how to use them.
\newblock \emph{Systems Engineering}, 20(6):497--511.

\bibitem[{Tavory and Timmermans(2014)}]{tavory2014abductive}
Iddo Tavory and Stefan Timmermans. 2014.
\newblock \emph{Abductive Analysis: Theorizing Qualitative Research}.
\newblock University of Chicago Press, Chicago, IL.

\bibitem[{Tian et~al.(2019)Tian, Zhang, Yu, and Cao}]{Tian2019ExploringTF}
Qingfeng Tian, Shuo Zhang, Hui-Yun Yu, and Guangming Cao. 2019.
\newblock \href {https://api.semanticscholar.org/CorpusID:96424944} {Exploring the factors influencing business model innovation using grounded theory: The case of a chinese high-end equipment manufacturer}.
\newblock \emph{Sustainability}.

\bibitem[{Tinder(2022)}]{tinder2022rewards}
Galen Tinder. 2022.
\newblock The rewards of oral narration.
\newblock \emph{The International Journal of Reminiscence and Life Review}, 9(1):25--33.

\bibitem[{Treude(2024)}]{treude2024qualitative}
Christoph Treude. 2024.
\newblock Qualitative data analysis in software engineering: Techniques and teaching insights.
\newblock In \emph{Handbook on Teaching Empirical Software Engineering}, pages 155--176. Springer.

\bibitem[{Trochim(1989)}]{trochim1989concept}
William M.~K. Trochim. 1989.
\newblock \href {https://doi.org/10.1016/0149-7189(89)90016-5} {An introduction to concept mapping for planning and evaluation}.
\newblock \emph{Evaluation and Program Planning}, 12(1):1--16.

\bibitem[{{\"U}bellacker(2024)}]{ubellacker2024academiaos}
Thomas {\"U}bellacker. 2024.
\newblock \href {https://arxiv.org/abs/2403.08844} {Academiaos: Automating grounded theory development in qualitative research with large language models}.
\newblock \emph{Preprint}, arXiv:2403.08844.

\bibitem[{Van~de Ven and Poole(1995)}]{vandevenpoole1995explaining}
Andrew~H. Van~de Ven and Marshall~Scott Poole. 1995.
\newblock \href {https://doi.org/10.5465/amr.1995.9508080329} {Explaining development and change in organizations}.
\newblock \emph{Academy of Management Review}, 20(3):510--540.

\bibitem[{Wang et~al.(2025)Wang, Colby, Okwara, Rao, Liu, and Monroy-Hern{\'a}ndez}]{Wang2025PolicyPulseLT}
Maggie Wang, Ella Colby, Jennifer Okwara, Varun~Nagaraj Rao, Yuhan Liu, and Andr{\'e}s Monroy-Hern{\'a}ndez. 2025.
\newblock \href {https://api.semanticscholar.org/CorpusID:278051126} {Policypulse: Llm-synthesis tool for policy researchers}.
\newblock \emph{Proceedings of the Extended Abstracts of the CHI Conference on Human Factors in Computing Systems}.

\bibitem[{Wasserman and Faust(1994)}]{wasserman1994social}
Stanley Wasserman and Katherine Faust. 1994.
\newblock \href {https://doi.org/10.1017/CBO9780511815478} {\emph{Social Network Analysis: Methods and Applications}}.
\newblock Cambridge University Press, Cambridge, UK.

\bibitem[{Weick(1995)}]{weick1995sensemaking}
Karl~E. Weick. 1995.
\newblock \emph{Sensemaking in Organizations}.
\newblock SAGE Publications, Thousand Oaks, CA.

\bibitem[{Wen et~al.(2021)Wen, Lin, Lai, Pan, Li, Lin, Zhou, Li, Wang, Zhang, Yu, Dong, Wang, Fung, Mishra, Lyu, Sur{\'i}s, Chen, Brown, Palmer, Callison-Burch, Vondrick, Han, Roth, Chang, and Ji}]{Wen2021RESINAD}
Haoyang Wen, Ying Lin, T.~Lai, Xiaoman Pan, Sha Li, Xudong Lin, Ben Zhou, Manling Li, Haoyu Wang, Hongming Zhang, Xiaodong Yu, Alexander Dong, Zhenhailong Wang, Yi~Ren Fung, Piyush Mishra, Qing Lyu, D{\'i}dac Sur{\'i}s, Brian Chen, Susan~Windisch Brown, and 7 others. 2021.
\newblock \href {https://api.semanticscholar.org/CorpusID:235097376} {Resin: A dockerized schema-guided cross-document cross-lingual cross-media information extraction and event tracking system}.
\newblock In \emph{North American Chapter of the Association for Computational Linguistics}.

\bibitem[{Wiebe et~al.(2025)Wiebe, Khan, Burns, and Slotta}]{wiebe2025qualitative}
Joel~P Wiebe, Rubaina Khan, Samantha Burns, and James~D Slotta. 2025.
\newblock Qualitative research in the age of llms: A human-in-the-loop approach to hybrid thematic analysis.
\newblock In \emph{Proceedings of the 19th International Conference of the Learning Sciences-ICLS 2025, pp. 1123-1131}. International Society of the Learning Sciences.

\bibitem[{Wu et~al.(2023)Wu, Bansal, Zhang, Wu, Li, Zhu, Zhang, Zhang, Wang et~al.}]{wu2023autogen}
Qingyun Wu, Gagan Bansal, Jieyu Zhang, Yiran Wu, Beibin Li, Erkang Zhu, Li~Zhang, Ruoyu Zhang, Xiaoyun Wang, and 1 others. 2023.
\newblock \href {https://arxiv.org/abs/2308.08155} {Autogen: Enabling next-gen {LLM} applications via multi-agent conversation}.
\newblock \emph{Preprint}, arXiv:2308.08155.

\bibitem[{Xu et~al.(2025)Xu, Yi, Lim, Xu, Well, Mery, Zhang, Zhang, Ji, Pingali et~al.}]{tama}
Huimin Xu, Seungjun Yi, Terence Lim, Jiawei Xu, Andrew Well, Carlos Mery, Aidong Zhang, Yuji Zhang, Heng Ji, Keshav Pingali, and 1 others. 2025.
\newblock Tama: A human-ai collaborative thematic analysis framework using multi-agent llms for clinical interviews.
\newblock \emph{arXiv preprint arXiv:2503.20666}.

\bibitem[{Yi et~al.(2025)Yi, Nguyen, Xu, Lim, Skrovan, Beri, Modi, Well, Leqi, Markey, and Ding}]{Yi2025SFTTASF}
Seungjun Yi, Joakim Nguyen, Hui-Rong Xu, Terence Lim, Joseph Skrovan, Mehak Beri, Hitakshi Modi, Andrew Well, Liu Leqi, Mia~K. Markey, and Ying Ding. 2025.
\newblock \href {https://api.semanticscholar.org/CorpusID:281421777} {Sft-ta: Supervised fine-tuned agents in multi-agent llms for automated inductive thematic analysis}.
\newblock \emph{ArXiv}, abs/2509.17167.

\bibitem[{Yi et~al.()Yi, Nguyen, Xu, Lim, Well, Markey, and Ding}]{autota}
Seungjun Yi, Joakim Nguyen, Huimin Xu, Terence Lim, Andrew Well, Mia Markey, and Ying Ding.
\newblock Auto-ta: Towards scalable automated thematic analysis (ta) via multi-agent large language models with reinforcement learning.
\newblock In \emph{ACL 2025 Student Research Workshop}.

\bibitem[{Zeigler et~al.(2000)Zeigler, Praehofer, and Kim}]{zeigler2000theory}
Bernard~P. Zeigler, Herbert Praehofer, and Tag~Gon Kim. 2000.
\newblock \emph{Theory of Modeling and Simulation: Integrating Discrete Event and Continuous Complex Dynamic Systems}, 2 edition.
\newblock Academic Press, San Diego, CA.

\bibitem[{Zhong et~al.(2023)Zhong, Wu, Li, Peng, and Wu}]{Zhong2023ACS}
Lingfeng Zhong, Jia Wu, Qian Li, Hao Peng, and Xindong Wu. 2023.
\newblock \href {https://api.semanticscholar.org/CorpusID:256808669} {A comprehensive survey on automatic knowledge graph construction}.
\newblock \emph{ACM Computing Surveys}, 56:1 -- 62.

\bibitem[{Zhong et~al.(2025)Zhong, Wang, and Field}]{Zhong2025HICodeHI}
Mian Zhong, Pristina Wang, and Anjalie Field. 2025.
\newblock \href {https://api.semanticscholar.org/CorpusID:281421574} {Hicode: Hierarchical inductive coding with llms}.
\newblock \emph{ArXiv}, abs/2509.17946.

\bibitem[{Zhou et~al.(2016)Zhou, Gao, and He}]{Zhou2016JointlyEE}
Deyu Zhou, Tianmeng Gao, and Yulan He. 2016.
\newblock \href {https://api.semanticscholar.org/CorpusID:11998198} {Jointly event extraction and visualization on twitter via probabilistic modelling}.
\newblock In \emph{Annual Meeting of the Association for Computational Linguistics}.

\end{thebibliography}
\appendix
\label{sec:appendix}

\section{Related work}
\label{relatedwork}

\paragraph{Computational Qualitative Data Analysis Development}
Augmenting and automating qualitative data analysis has been investigated for a long time \citep{chen2016challenges, bryda2023qualitative}.
Existing works move from keyword matching to fully automation trial. 
CoAICoder \citep{gao2023coaicoder} leverages AI to enhance human-to-human collaboration in qualitative data analysis.
Cody \citep{Rietz2021CodyAA} and PaTAT \citep{Gebreegziabher2023PaTATHC} serve as a code recommender by learning from patterns in human-generated codes. 
These efforts first show chances in AI-assisted qualitative analysis. After that, researchers focus more on semantic ability and quality of computational methods.

Recently, Large Language Models have been used to facilitate qualitative research \citep{Hou2024PromptbasedAF, barros2025large}.
ThemeViz \citep{themeViz} and \cite{montes2025largelanguagemodelsthematic} demonstrate that LLM can scaffold human efforts. 
At this stage, people primarily treat the LLM as a conversational partner. 
Building on this, several works aim to establish efficient human-AI collaborative frameworks for qualitative analysis \citep{Rao2024QuaLLMAL, sharma2025details, wiebe2025qualitative, tama}. 
Similarly, MindCoder \citep{} introduces a framework designed to produce transparent analytical traces.
From another branch, researchers also explored fully automated qualitative analysis workflow, performing initial trials on LLM-based automatic thematic analysis\citep{braun2006using, performing, khan2024automating}.
LLOOM \citep{Lam2024ConceptIA}, Thematic-LM \citep{Qiao2025ThematicLMAL}, HICode \citep{Zhong2025HICodeHI} and Auto-TA \citep{autota} provide fully-automated pipelines that convert an unannotated corpus to a comprehensive codebook.
Consistent with the thematic-analysis paradigm, their goal is to summarize themes across the corpus, not to construct an explicit hierarchy of concepts or rigidly model logical relationships among codes \citep{braun2021can}. 
For methods with higher level modeling, they require explicitly hierarchical, relational, and iterative (e.g., grounded theory). 
LOGOS \cite{pi2025logos} tried to fully automated grounded-theory development, mirroring standard grounded-theory practice and support deeper and finer-grained interpretive analyses in qualitative research.

\paragraph{Domain Specific Computational Qualitative Data Analysis}

Considerable attention has been devoted to qualitative analysis in recent years in computer science community \citep{Li2020ConnectingTD, Li2023OpenDomainHE, li2022futureonedimensionalcomplexevent, jin-etal-2022-event, Wen2021RESINAD, Du2022RESIN11SE}. 
A major line of work in the AI community represents events as standard knowledge graphs, where actions and named entities are modeled as nodes with predefined edge structures, and graph neural networks are used to perform reasoning over them.
Adopting a comparable formalism, \cite{Cheng2024SHIELDLS} specifically addresses the domain of the electric vehicle battery supply chain. 
Software Engineering is also another area where qualitative analysis has been used \cite{treude2024qualitative, Montes2025LargeLM}.
Moreover, the AI community also sees the potential of qualitative research in the trending area, like multi-agent systems \cite{cemri2025mast, pan2025measuring}, and vibe-coding \cite{huang2025professional}.
In the context of broader qualitative data analysis application, scholars across diverse disciplines are increasingly incorporating LLMs into their analytic pipelines. 
For example, policy analysts have utilized LLMs to distill themes from extensive text archives \citep{Fang2025DecodingCI, Wang2025PolicyPulseLT}. 
Literature reviews, a cornerstone of qualitative inquiry, offer another fertile ground for LLM-driven analysis, specifically for detecting and structuring themes across large bodies of publication \citep{ubellacker2024academiaos, Singh2025Ai2SQ, padmakumar2025intent}.

\paragraph{Evaluating Qualitative Data Analysis Results} 

Existing frameworks for evaluating qualitative analysis emphasize a multi-dimensional approach designed to ensure both methodological rigor and theoretical depth \cite{nowell2017thematic, manuj2012reviewer, charmaz2021pursuit, ritchie2013qualitative, gee2025introduction, smith2009ipa}.
As synthesized in the literature, valid qualitative assessment relies on six primary pillars.
(1) The analysis must demonstrate fidelity and empirical grounding, ensuring that the resulting theory is "true" to the data, captures the participants' voices, and possesses sufficient explanatory power.
(2) The reasoning must be logical and sound, characterized by confirmability, analytic rigor, and interpretations that are strongly supported by evidence.
(3) Consistency and stability are paramount; this requires dependability and traceability (an audit trail) to establish the transparency and reliability of the research process.
(4) The reusability of the findings is evaluated through their transferability, allowing readers to judge applicability to other contexts via thick description.
(5) The framework mandates sufficient reflexivity, requiring researchers to acknowledge their assumptions and maintain theoretical openness to modification.
(6) The evaluation prioritizes the originality and contribution of the results, seeking novel conceptualizations that extend beyond local descriptions to offer generalizable theoretical understanding.

From the computational perspective, researchers also adpot different ways to evaluate qualitative analysis results.
Traditional assessment typically depends on event/theme prediction tasks (e.g., edge type classification). 
People use traditional statistical scores, like F1, Recall, Precision, to measure how well the trained model can predict the event/theme \cite{Li2020ConnectingTD, Zhong2025HICodeHI}.
When it comes to semantic level, some classic NLP measurements are used. BLUE, ROUGE, BERTScore are used to compare the overlap or similarisy between ground truth Model generated results \cite{parfenova2025text, Yi2025SFTTASF,, Qiao2025ThematicLMAL}.
However, given the rareness of human ground truth, research also explore more flexible way to judge the results, either using human judgement \cite{Montes2025LargeLM} or LLM-as-a-judge \cite{pi2025logos}. But these evaluation methods are still in an early stage.

\section{How We Developed Our Theoretical Framework}
\label{app:theory-dev}

\subsection{Overview and Design Stance}
We developed the 4$\times$4 landscape through an iterative theory-building process aimed at producing a framework that is (i) \textit{conceptually clean} (non-overlapping commitments), (ii) \textit{operational} (annotators can reliably apply it), and (iii) \textit{interdisciplinarily grounded} (linked to established distinctions across qualitative methodology and modeling traditions). The process is best described as \textit{grounded-theory-like}, but extended by the need to reconcile and anchor concepts across multiple foundational literatures (linguistics, pragmatics, interpretive social science, philosophy of science, systems science, and engineering design).

\subsection{Iterative Development Cycle}
Across approximately five months, we conducted roughly ten rounds of iteration following a repeated cycle:

\begin{enumerate}
    \item \textbf{Incremental paper collection and landscape exposure.}
    We expanded a seed set of representative works spanning (a) qualitative research methods and exemplars and (b) NLP/LLM-based qualitative automation and related modeling traditions. This expansion was iterative: newly encountered formalisms and analytic ``moves'' frequently revealed missing distinctions and edge cases, prompting further targeted collection.

    \item \textbf{Pattern recognition and empirical grounding.}
    We repeatedly asked: \emph{what kinds of meaning-making commitments the researchers attempt, and what kind of model form does the paper treat as its primary explanation?}
    We recorded recurring output commitments and representational primitives (e.g., surface-faithful restatement vs.\ categorical organization vs.\ within-frame inference vs.\ external theoretical reframing; and, for modeling, stage narratives vs.\ directed pathways vs.\ coupled-feedback systems).

    \item \textbf{Pattern condensation into discriminative axes.}
    We progressively condensed observations into two orthogonal axes---\textbf{meaning-making level} and \textbf{modeling level}---with explicit boundary tests intended to reduce overlap and ambiguity. A key design goal was to classify \emph{what the analysis commits to}, rather than \emph{what the method is called}.

    \item \textbf{Framework proposal, boundary testing, and revision.}
    We iteratively proposed definitions, then stress-tested them against challenging cases where boundaries are known to fail (e.g., L2 vs.\ L3 ``themes vs.\ implied commitments''; L3 vs.\ L4 ``within-frame inference vs.\ etic theoretical import''; modeling L3 vs.\ L4 ``pathway-like directed dependence vs.\ mutually coupled feedback systems''). When definitions failed, we revised: (i) operational heuristics, (ii) decision rules, and (iii) failure-mode clarifications.

    \item \textbf{Application-driven validation (``fitness'' checks).}
    We repeatedly applied the evolving framework to diverse papers to check \textbf{coverage} (can it classify what exists?), \textbf{stability} (do similar papers land similarly?), and \textbf{operationality} (can others follow rules and reach comparable judgments?). These fitness checks served the same function as constant comparison in grounded theory: disagreements and hard cases were treated as signals that the framework required refinement rather than as annotator error.
\end{enumerate}

\subsection{Interdisciplinary Theoretical Anchoring}
Unlike single-paradigm framework construction, our development required explicit anchoring across disciplines to avoid purely rhetorical distinctions.
For meaning-making levels, we aligned boundaries with established separations such as manifest/latent content, emic/etic distinctions, frame-based inference, and pragmatics (implicature, presupposition, and speech acts), as well as interpretivist sensemaking and thick description.
For modeling levels, we grounded distinctions in the representational commitments of stage narratives, directed pathway models (e.g., DAG-like causal accounts, workflows/FSMs), and complex-systems interdependency models (feedback-centered coupling).
This anchoring served two purposes: (i) to justify why each level is qualitatively distinct rather than ``just deeper,'' and (ii) to supply decision rules that are communicable to both QR and NLP audiences.

\subsection{Social Proof and Community Feedback}
To ensure the framework is not only internally coherent but also intuitively usable by its intended communities, we repeatedly sought external feedback during development.
We presented intermediate versions in a research seminar attended by 10+ graduate-level researchers (including cognitive science PhDs) and conducted informal expert consultations with more than 5 faculties and domain experts spanning philosophy, linguistics, and sociology.
Across these interactions, feedback consistently emphasized that the framework is handy and intuitive as a shared vocabulary for describing (a) what an analysis output \emph{commits to} and (b) what kind of model form it \emph{privileges}.
We used this feedback as a practical check on communicability: when experts found a boundary confusing or a criterion too implicit, we revised definitions and boundary tests accordingly.

\section{Mapping Qualitative Research Methods into the Landscape}
\label{app:method-mapping}

\begin{table*}[t]
\centering \small
\begin{tabular}{p{0.1\linewidth}p{0.18\linewidth}p{0.20\linewidth}p{0.20\linewidth}p{0.20\linewidth}}
\toprule
 & \textbf{Model L1} \newline (static map) & \textbf{Model L2} \newline (stages/timeline) & \textbf{Model L3} \newline (causal pathways) & \textbf{Model L4} \newline (feedback dynamics) \\
\toprule
\textbf{Meaning L1} \newline (descriptive) &
Trace-faithful inventory; association sketch &
Event/timeline summary; episode segmentation &
Described ``drivers'' without strong interpretation &
Described cycles/loops as narration (often rhetorical) \\
\textbf{Meaning L2} \newline (categorical) &
Themes/codes + relations; taxonomies &
Theme evolution across phases; stage-labeled themes &
Theme-based causal map; mechanism candidates &
Theme-linked loop hypotheses; coupled pattern stories \\
\textbf{Meaning L3} \newline (interpretive) &
Implied norms/goals mapped to relations &
Interpretive phase narrative (what shifts, why it matters) &
Mechanism explanation with warrants across excerpts &
Interpretive feedback account (endogenous escalation/relaxation) \\
\textbf{Meaning L4} \newline (theoretical) &
Theory-guided thematic/relational model &
Theory-guided process model (stages via a lens) &
Theory-mediated mechanism/pathway model &
Theory-mediated system model (loops, delays, regimes) \\
\hline
\end{tabular}
\caption{The 4$\times$4 landscape as a conceptual coordinate system. Rows specify the semantic commitments made from traces; columns specify the representational commitments made about the phenomenon. Each cell implies different validity demands and evaluation targets.}
\label{tab:landscape-conceptual}
\end{table*}

The qualitative research community has developed a relatively comprehensive taxonomy of methods, addressing questions such as analytical targets, appropriate conditions of use, and conventions for data collection, processing and analyzing.
Here, we map established qualitative research methods onto our analytical landscape. 
Table ~\ref{tab:landscape-conceptual} show a conceptual coordinate metric for each \textbf{(Meaning making, Modeling)} pair.
Fig.~\ref{fig:mapping} generally shows how existing QR methods fit in our landscape.
We emphasize that the placements below are \textbf{high-level and illustrative rather than exhaustive}.
Method labels do not uniquely determine a single cell in the landscape; however, many research traditions exhibit typical or canonical regions of emphasis, arranged here generally from lower to higher degrees of interpretive and structural complexity:

\begin{figure}
    \centering
    \includegraphics[width=1\linewidth]{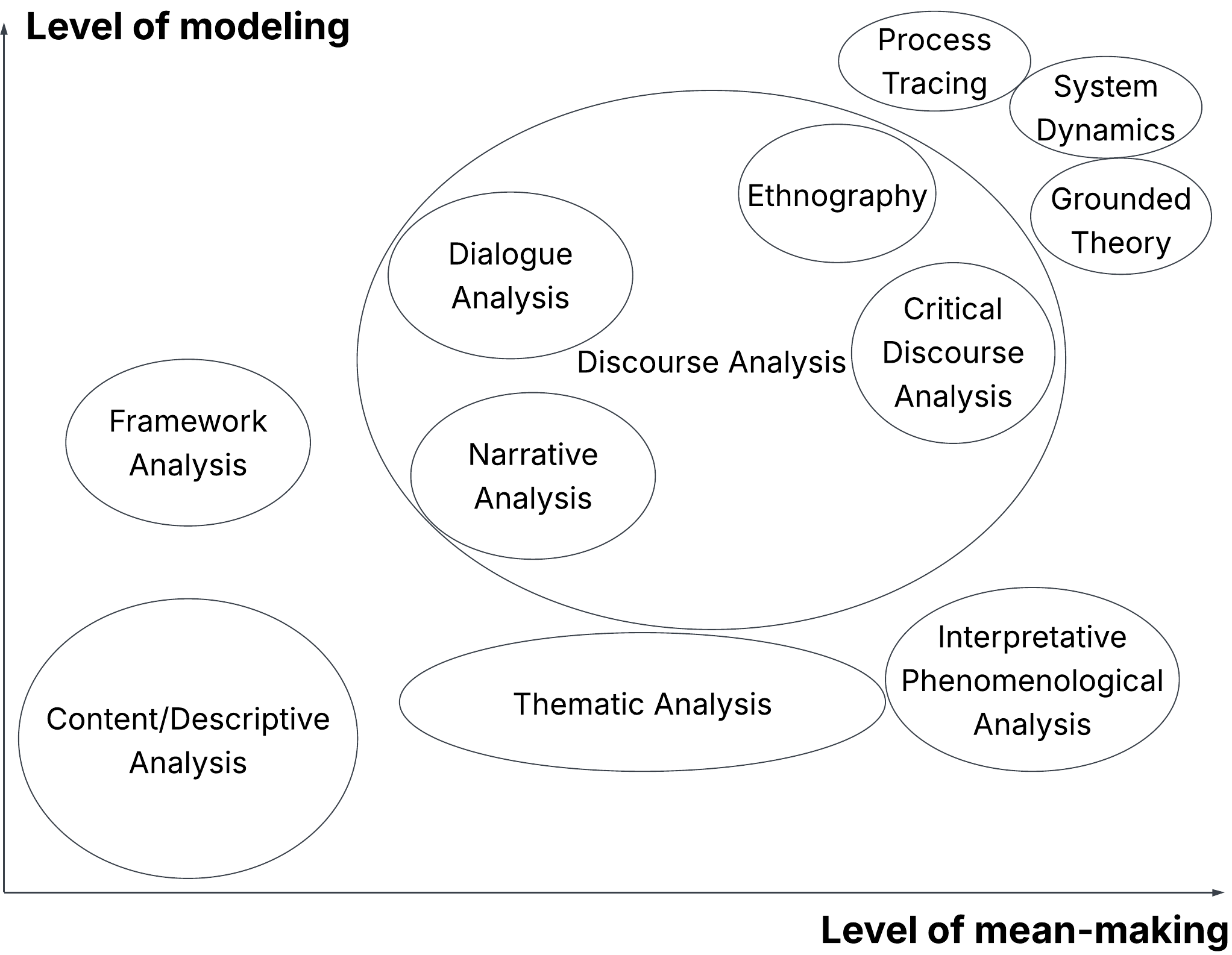}
    \caption{Mapping QR methods to our landscape}
    \label{fig:mapping}
    \vspace{-1mm}
\end{figure}

\begin{itemize}
    \item \textbf{Qualitative content analysis / descriptive reporting} typically emphasizes \mean{1}--\mean{2} with \model{1}, prioritizing faithful summaries, coding schemes, and relatively static thematic organization \citep{berelson1952contentanalysis,hsieh2005threeapproaches,krippendorff2018contentanalysis}.

    \item \textbf{Framework analysis} occupies a distinct position emphasizing \model{2} with \mean{1}--\mean{2}, offering a structured, matrix-based approach to data management that facilitates systematic comparison across cases while maintaining a descriptive focus \citep{parkinson2016framework, goldsmith2021using}.

    \item \textbf{Thematic analysis} commonly targets \mean{2} with \model{1}, focusing on the identification and organization of themes; some variants extend toward \model{2} when themes are explicitly sequenced or staged over time \citep{braun2006thematic,attridestirling2001thematic}. Highly reflexive thematic analysis can also go beyond to \mean{3} \citep{braun2019reflecting, braun2021one}.

    \item \textbf{Interpretative phenomenological analysis (IPA)} typically aims at \mean{3}, and sometimes \mean{4} when explicit theoretical or philosophical lenses are central. Representationally, IPA often relies on \model{1} or \model{2} structures to prioritize the depth of lived experience \citep{smith2009ipa}.

    \item \textbf{Discourse analysis and related traditions} (including \textbf{Narrative}, \textbf{Dialogue}, \textbf{Ethnography}, and \textbf{Critical Discourse Analysis}) form a central cluster in our landscape, frequently emphasizing \model{2}--\model{4} with \mean{2}--\mean{4}. These methods foreground the structures of communication, temporal bracketing, and cultural context \citep{brown1983discourse, langley1999strategies}. Within this cluster:
    \begin{itemize}
        \item \textbf{Narrative analysis} and \textbf{Dialogue Analysis} foregrounds episodes and plots \citep{cortazzi1994narrative, carlson2012dialogue}.
        \item \textbf{Critical Discourse Analysis (CDA)} and \textbf{Ethnography} often push toward higher mean-making (\mean{3}) by integrating broader social power dynamics and cultural systems \citep{fairclough2023critical, hammersley2006ethnography}.
    \end{itemize}

    \item \textbf{Process tracing and mechanism-centered explanation} align with \model{3} and \mean{3}--\mean{4}, focusing on warranted within-case causal and mechanistic claims, often mediated by explicit theoretical expectations \citep{bennett2015processtracing,hedstrom1998socialmechanisms}.

    \item \textbf{Grounded theory and case-based theorizing} generally push toward \mean{3}--\mean{4} and often toward \model{3}, emphasizing emergent mechanisms or explanatory logics, although the specific representational forms vary widely across traditions \citep{glaser1967discovery,charmaz2014constructing,george2005case}.

    \item \textbf{System dynamics and simulation-based explanation} align most directly with \model{4}, typically paired with \mean{3}--\mean{4}, as feedback-based and holistic explanations generally require substantial interpretive and theoretical commitments to model complex interdependencies \citep{forrester1961industrial,sterman2000business,meadows2008thinking}.
\end{itemize}

\section{Expanded Level of Meaning-Making}
\label{app:meaing}

Qualitative research is often described as “interpretive,” yet in practice different papers (and different LLM-based pipelines) construct meaning at markedly different depths and scopes. Some outputs remain close to what is explicitly said, some organize recurring topics, some infer implicit commitments and social/pragmatic implications, and some re-situate the same material inside external theoretical systems. Classic methods papers recognize related distinctions---for example, manifest versus latent content in content analysis \citep{berelson1952contentanalysis,holsti1969contentanalysis,krippendorff2018contentanalysis,graneheim2004trustworthiness,hsieh2005threeapproaches}, or informant-centric versus researcher-centric “orders” of coding \citep{gioia2013rigor,saldana2021coding}. However, there is limited consensus on a single, domain-general way to characterize \emph{what kind of meaning} an analysis produces, in a manner that is both (i) conceptually clean (non-overlapping levels) and (ii) operational (annotators can reliably distinguish levels).

We propose a four-level theory of meaning-making. Each level is defined by the \emph{semantic commitments} introduced by the output, not by the qualitative method named in the paper. Concretely, we distinguish levels along two coupled dimensions: \textbf{semantic depth} (how much implicit meaning is made explicit) and \textbf{conceptual scope} (whether meaning is constructed within the case’s own semantic frame, or by importing external conceptual resources). This yields four levels---Descriptive (L1), Categorical (L2), Interpretive (L3), and Theoretical (L4)---which we use as one axis of our 4$\times$4 landscape.

\subsection{Level 1: Descriptive meaning-making (manifest, surface-faithful)}
\label{subsec:mm-l1}
L1 outputs restate, summarize, or extract \emph{explicit} content with minimal abstraction and minimal inference. This corresponds to “manifest” content in classic content analysis: what is directly observable in the text or record, in a way that can be checked against the source \citep{berelson1952contentanalysis,holsti1969contentanalysis,krippendorff2018contentanalysis}. In qualitative content analysis terminology, L1 stays close to the surface form (e.g., who said what, what happened, what was mentioned), without interpreting intentions, unstated causes, or normative implications \citep{graneheim2004trustworthiness,kleinheksel2020demystifying}.

For NLP/LLM readers, L1 aligns with extractive/abstractive summarization and information extraction when the output remains strictly grounded in the provided text (e.g., generating a faithful synopsis, listing mentioned entities, enumerating reported events).

\subsection{Level 2: Categorical meaning-making (themes, topics, and patterning)}
\label{subsec:mm-l2}
L2 introduces \emph{organizational abstraction}: it groups surface observations into recurring categories, themes, domains, or types. This is the core analytic product of many thematic and coding-oriented workflows (e.g., themes as patterned responses within a dataset) \citep{braun2006thematic,saldana2021coding,miles2014qualitative}. Importantly, L2 abstraction is primarily \emph{frame-faithful}: categories summarize what appears in the corpus rather than asserting a deeper latent “why” that must be true.

In content analysis terms, L2 often corresponds to building a coding scheme that clusters manifest observations into higher-level labels \citep{krippendorff2018contentanalysis,hsieh2005threeapproaches}. For NLP/LLM readers, L2 aligns with clustering, topic modeling, weakly-supervised labeling, taxonomy induction, and thematic summarization---as long as the output remains an organization of surface content rather than an inference about implicit commitments.

\subsection{Level 3: Interpretive meaning-making (implicit meaning entailed by the frame)}
\label{subsec:mm-l3}
L3 makes a qualitative shift: it articulates \emph{implicit or latent meaning} that is not explicitly stated but is plausibly \emph{entailed by the semantic frame} of the situation. This is closely aligned with “latent” content analysis in qualitative research \citep{graneheim2004trustworthiness,graneheim2017challenges}, while retaining a key constraint: the interpretation is warranted by the case’s internal context and shared commonsense/genre knowledge, rather than by invoking a specialized external theory.

Three research traditions provide a principled foundation for this level:

\textbf{(i) Frame-based inference in linguistics and cognitive semantics.}
In Frame Semantics, understanding an utterance requires evoking a structured frame (a schematic situation with participants and roles) that licenses inferences beyond what is overtly said \citep{fillmore1982framesemantics}. FrameNet operationalizes this idea for computational linguistics by cataloging frames and frame elements that support systematic within-frame inference \citep{baker1998framenet}.

\textbf{(ii) Pragmatic inference and implied commitments.}
Classic pragmatics characterizes how speakers communicate more than they literally say, via implicature, presupposition, and speech-act force \citep{grice1975logic,austin1962how,searle1969speechacts}. Common-ground theory formalizes how discourse participants rely on shared background propositions that are not always explicit \citep{stalnaker2002commonground}. These mechanisms motivate L3 annotations such as inferring unstated goals, obligations, norms, or relational implications that make the discourse coherent.

\textbf{(iii) Interpretivist social science and sensemaking.}
Interpretive traditions emphasize that meaning is constructed by connecting cues to contextual frames to form a coherent account \citep{weick1995sensemaking,goffman1974frameanalysis}. “Thick description” explicitly treats interpretation as making implicit cultural/interactional context legible, without necessarily reducing it to external theory \citep{geertz1973interpretation}.

For NLP/LLM readers, L3 is the level most naturally connected to \emph{natural language inference} and pragmatic reasoning: producing statements that are not verbatim in the text but follow from it under shared background assumptions \citep{dagan2006rte,bowman2015snli}. L3 outputs are defeasible (interpretations can be challenged), but they are constrained: they should be defensible as “what must (or very likely) is going on” \emph{given} the situation described.

\subsection{Level 4: Theoretical meaning-making (external reframing and etic constructs)}
\label{subsec:mm-l4}
L4 expands conceptual scope: it re-situates the case within an \emph{external theoretical or conceptual system} that is not entailed by the semantic frame alone. The defining criterion is not whether a paper cites theory, but whether theory functions as an \emph{analytic engine} shaping the meaning that is produced \citep{hsieh2005threeapproaches,gioia2013rigor,saldana2021coding}. In qualitative methodology terms, L4 aligns with moves such as theoretical coding and second-order theorizing, where the analyst introduces constructs that are not native to the participants’ accounts \citep{gioia2013rigor,saldana2021coding}. This parallels the classic emic/etic distinction: L3 stays close to emic sense-making (within the case’s frame), whereas L4 introduces etic categories (external analytic vocabulary) \citep{pike1954language}.

Two additional literatures motivate why L4 should be treated as a distinct level rather than “just deeper interpretation”:

\textbf{Sensitizing concepts and theory-guided seeing.}
Interpretive social science often uses sensitizing concepts to guide what counts as evidence and how observations are connected; such concepts are explicitly not reducible to surface description \citep{blumer1954wrongtheory}. In modern qualitative theorizing, abductive analysis makes the theory step explicit: surprising observations motivate importing/constructing explanatory concepts that reorganize the case \citep{tavory2014abductive}.

\textbf{Hermeneutics and the productive role of interpretation.}
Hermeneutic traditions emphasize that interpretation can be “productive,” generating new understanding by applying horizons and conceptual resources that exceed what is explicitly said \citep{gadamer1989truthmethod,ricoeur1976interpretationtheory}. This supports treating L4 as a qualitatively distinct epistemic move: meaning is constructed \emph{through} an external lens.

For computational readers, L4 is the regime where the output’s validity depends on whether the imported framework is appropriate and correctly applied (e.g., mapping a case to an established theoretical model, typology, or mechanism vocabulary). This is also the level most sensitive to out-of-context hallucination: an LLM can easily “sound theoretical” by naming frameworks without the warrants that a human theorist would provide.

\subsection{Boundary tests and operational heuristics}
\label{subsec:mm-boundaries}
Because the boundary between L3 and L4 is the most failure-prone in both human annotation and LLM automation, we adopt explicit decision rules grounded in prior distinctions (manifest/latent; emic/etic; first-/second-order coding):

\paragraph{L1 vs.\ L2 (restatement vs.\ organization).}
If the output could be produced by rephrasing or extracting without introducing a reusable label system, it is L1. If it introduces categories/themes that summarize recurring elements across instances, it is L2 \citep{braun2006thematic,krippendorff2018contentanalysis}.

\paragraph{L2 vs.\ L3 (topics vs.\ implicit commitments).}
If the output only says \emph{what kinds of things} appear (themes/topics) it is L2. If it asserts \emph{what must be true to make the case coherent}---e.g., inferred goals, norms, implied obligations, latent tensions, unspoken constraints---it is L3, supported by pragmatic and frame-based inference \citep{grice1975logic,fillmore1982framesemantics,graneheim2017challenges}.

\paragraph{L3 vs.\ L4 (within-frame inference vs.\ theoretical reframing).}
A practical discriminator is an \textbf{external-lens test}:
\begin{quote}
If a competent reader could, using the situation’s ordinary frame and shared commonsense/pragmatic knowledge, infer the meaning without importing a specialized construct, it is L3. If the meaning depends on introducing an external framework/etic vocabulary (and the output would be incomplete or uninterpretable without that framework), it is L4.
\end{quote}
This test is consistent with emic/etic distinctions \citep{pike1954language} and with qualitative guidance that theory-driven (directed) analysis differs from inductive/latent interpretation \citep{hsieh2005threeapproaches,gioia2013rigor}.

\paragraph{Reliability and transparency.}
As semantic depth and conceptual scope increase, reliability is harder to achieve; qualitative methodology therefore emphasizes explicit code definitions, auditability of inference, and trustworthiness criteria \citep{lincoln1985naturalistic,graneheim2004trustworthiness,krippendorff2018contentanalysis}. In our annotation protocol (Sec.~\ref{sec:annotation-study}), we operationalize this by requiring short rationales and by calibrating annotators on boundary cases, especially L3/L4.

\paragraph{Why this matters for LLM-driven qualitative research.}
Current LLM tooling strongly supports L1--L2 outputs (summaries, clustering, thematic labels), but bridging to L3--L4 requires controlled inference: (i) making implicit meaning explicit without drifting beyond what the frame supports (L3), and (ii) applying external theories in a disciplined way, with warrants and limits (L4). Our four-level meaning-making axis isolates these distinct computational challenges, enabling more precise evaluation of “how far” an automated qualitative pipeline actually goes.

\section{Expanded Level of Modeling}
\label{app:modeling}

\paragraph{What we mean by ``modeling''.}
In this paper, \emph{modeling} refers to the \textbf{explicit representational structure} a study produces to describe a phenomenon---what the units are, what relations hold among them, and what kinds of reasoning the representation is intended to support.
This is closer to the notion of \emph{models as representations} in system science and simulation (rather than ``models'' as parameterized predictors in ML) \citep{box1976science, zeigler2000theory, law2015simulation}.
Importantly, a modeling level is \textbf{orthogonal} to a meaning-making level: the same interpretive claim can be expressed as (i) a static thematic map, (ii) a phase timeline, (iii) a causal mechanism graph, or (iv) a dynamical feedback system.

We operationalize \textbf{four levels of modeling} that recur across qualitative research and sociotechnical-system analysis.
The key discriminators are: (i) \textbf{time} (absent vs. present), (ii) \textbf{causality/mechanism} (absent vs. present), and (iii) \textbf{dynamical semantics} (static dependencies vs. iterative state updates with feedback).

\subsection{The four levels}
\label{subsec:modeling-levels}

\paragraph{Level 1: Static configuration and relational models (themes as structure).}
Level~1 models are \textbf{static} representations of a system’s structure.
The typical output is a set of \textbf{themes / constructs / entities} plus \textbf{relationships} that are \emph{not} interpreted as temporal evolution or causal production.
This includes (a) thematic maps and thematic networks that systematize how themes relate in a corpus \citep{braun2006thematic, attridestirling2001thematic}, (b) social network / relational representations (ties among actors, groups, or organizations) used as a cross-sectional structure \citep{wasserman1994social, borgatti2009network}, and (c) static knowledge/configuration layouts such as concept maps used to organize concepts and relations without committing to temporal or causal semantics \citep{trochim1989concept, novak2008conceptmaps}.

\noindent\textbf{Signature:} ``what relates to what'' (clusters, centrality, co-occurrence, association, configuration) in a snapshot.
Relations may be labeled and even directed (e.g., asymmetric ties), but the representation does not claim that the system \emph{evolves} or that A \emph{causes} B.
For CS/NLP readers, Level~1 is most naturally viewed as \emph{graph extraction / clustering / structural summarization} over qualitative material.

\paragraph{Level 2: Stage / phase / timeline models (time without causality).}
Level~2 models introduce \textbf{time} as a first-class organizing principle, but they still \textbf{do not} make the core representational commitment of causal/mechanistic production.
The output is a \textbf{sequence} of stages/phases (or a timeline of events/episodes), often produced via temporal bracketing, narrative sequencing, or process description \citep{vandevenpoole1995explaining, langley1999strategies, pettigrew1990longitudinal}.
A common form is a phase model of change (Phase~1 $\rightarrow$ Phase~2 $\rightarrow$ Phase~3) or a chronicle of key events, where the edges mean ``next / then / later'' rather than ``produces / changes''.
Sequence-analytic perspectives also fall naturally here when the modeling product is primarily an over-time patterning of states/events rather than a causal mechanism model \citep{abbott1995sequence, abbott2001timematters}.

\noindent\textbf{Signature:} ``what happens when'' (ordering, phases, episodes), with descriptive transition language.
For CS/NLP readers, Level~2 corresponds to \emph{segmentation and sequencing} problems (phase detection, timeline construction, episode modeling), where the representation encodes temporal order but not causal effect.

\paragraph{Level 3: Causal dependency and mechanism models (directed influence).}
Level~3 models explicitly represent \textbf{causal or mechanism-like dependencies}.
Nodes are constructs/states/actors, and edges indicate that one element \emph{changes/enables/blocks} another, potentially with conditions, mediators, or moderators.
This aligns with causal-graph traditions in social science and AI \citep{pearl2009causality, spirtes2000cps}, as well as qualitative/soft-systems artifacts that are intended as causal mechanism diagrams rather than mere association maps.
Related representational families include cognitive/causal maps and fuzzy cognitive maps when used as explicit influence structures \citep{kosko1986fuzzy}.

\noindent\textbf{Signature:} ``why / how'' in the sense of directed production (A $\rightarrow$ B as influence), but still often \textbf{read as a static dependency structure} rather than an iterated dynamical system.
The model may be qualitative and non-executable; what matters is the directional, mechanism-committing semantics.

\paragraph{Level 4: Dynamical systems / feedback / complex systems models (state + update + iteration).}
Level~4 models treat the phenomenon as an \textbf{evolving system} whose behavior unfolds through \textbf{iterative state change}.
The first-class units are \textbf{system components} (institutions, infrastructures, policy regimes, collective behaviors, capacities, constraints, resources) that have \textbf{states} which update over time (quantitatively or qualitatively).
The model specifies \textbf{transactions} (mechanism-like interactions) that update component states, and it uses \textbf{feedback}, \textbf{delays}, and often \textbf{nonlinear responses} to explain system-level patterns over time (waves, escalation/relaxation cycles, overshoot and correction, lock-in, tipping points, collapse/stabilization) \citep{forrester1961industrial, sterman2000business, meadows2008thinking}.

Crucially, Level~4 is \textbf{not restricted} to classical stock--flow diagrams.
A Level~4 model may be:
(i) qualitative system dynamics / feedback-system reasoning (reinforcing/balancing loops as explanatory machinery);
(ii) coupled regime/phase systems where components move among qualitative states and co-evolve; or
(iii) executable simulation models (agent-based, discrete-event) when the simulation is used to explain emergent trajectories \citep{bonabeau2002agent, macywiller2002actors, epstein2006generative, gilbert2005simulation, law2015simulation, zeigler2000theory}.

\noindent\textbf{Signature:} the paper’s core model claim depends on \textbf{endogenous time evolution} (state at $t$ shapes state at $t{+}1$) and uses feedback/iteration as the explanatory engine, not merely as metaphor.

\subsection{Operational boundaries and guardrails}
\label{subsec:modeling-boundaries}

\paragraph{Level 1 vs.\ Level 2: static structure vs.\ explicit temporality.}
A paper is Level~1 if the primary model is a \emph{snapshot} of themes/entities and their relations (association/ties/configuration) without encoding ordering, phases, or evolution.
It is Level~2 when the model explicitly organizes the phenomenon as \emph{over-time} stages, phases, or timelines \citep{langley1999strategies, pettigrew1990longitudinal}.

\paragraph{Level 2 vs.\ Level 3: temporal ordering vs.\ causal production.}
A paper remains Level~2 if arrows mean ``next'' (temporal succession) but the analysis does not commit to edges as causal/mechanistic (``produces/changes/enables'').
It becomes Level~3 when directed relations are used to explain how one construct generates changes in another, i.e., when the representation has causal semantics \citep{pearl2009causality, spirtes2000cps}.

\paragraph{Level 3 vs.\ Level 4: static causal graphs vs.\ dynamical system semantics.}
Level~3 models can include cycles, and they can talk about ``feedback'' rhetorically.
Level~4 requires something stronger: the paper must articulate \textbf{state + update semantics + iteration} so that the explanation of observed/predicted behavior depends on system evolution over time (often via feedback and delays) \citep{sterman2000business, forrester1961industrial}.

A practical diagnostic:
\begin{itemize}
  \item If you can remove time iteration and the core model claim still stands (it remains a story about dependencies or pathways), it is likely Level~3.
  \item If the core claim depends on iterative updating to produce macro behavior patterns (waves, oscillation, overshoot, collapse, lock-in), it is Level~4.
\end{itemize}

\paragraph{Why this matters for LLM-based qualitative automation.}
This modeling axis makes concrete what is being produced and therefore what should be evaluated.
Many current LLM pipelines naturally yield Level~1 products (themes + static relational maps) or Level~2 products (phase/timeline summaries).
Moving to Level~3 requires reliable causal semantics and defensible mechanism claims; moving to Level~4 requires explicit state/update/feedback commitments that support trajectory-level explanations and intervention reasoning \citep{pearl2009causality, sterman2000business, zeigler2000theory}.

\section{Annotation Study Examples}

\subsection{\textcolor{orange}{Dimension 1: level of Meaning-Making}}
\label{app:annotation-example}

\subsubsection{\mean{1}: Descriptive (manifest, surface-faithful).}
\begin{itemize}
    \item \textbf{\cite{Keith2017IdentifyingCK}:}
    \begin{quote}
        \textit{"...and extract entity mentions. Mentions are token spans that (1) were identified as “persons” by spaCy’s named entity recognizer, and (2) have a (firstname,lastname) pair as analyzed by the HAPNIS rule-based name parser,6 which extracts, for example, (John, Doe) from the string Mr. John A. Doe Jr.. To prepare sentence text for modeling, our pre-processor collapses the candidate mention span to a special TARGET symbol. To prevent overfitting, other person names are mapped to a different PERSON symbol; e.g. 'TARGET was killed in an encounter with police officer PERSON.'"}
    \end{quote}
    This paper proposes a framework of identifying "civilians killed by police" via "event-entity extraction". The framework is a mixture of Quantitative Analysis and Qualitative Analysis. However, Qualitative Analysis process only happens for event and entity extraction given unstructured data (news articles), serving as a preprocessing step for the proceeding Quantitative Analysis step. Hence, this is an example of \mean{1}.
    \item \textbf{\cite{article}:}
    \begin{quote}
        \textit{"Non-neural Approaches. Widely used supervised approaches for NER include HMM, MEMM, SVM, and CRF. Traditional approaches relied solely on underlying algorithm and initial training data."}
    \end{quote}
   This paper discusses different algorithms for Named Entity Recognition (NER) tasks, which involve learning underlying patterns of entities for extraction. Despite the requirement for global knowledge aggregation to form generalized understandings of entity patterns, such knowledge relies heavily on statistical distributions rather than semantic meaning. For example, if an entity is replaced with an arbitrary symbol (e.g., A) or masked, the overall abstracted knowledge remains unchanged because it is based on distributional statistics. However, this invariance does not hold for semantic knowledge: replacing the word “Mars” with “Atlantic Ocean” in a context of exploration changes the content, even though both are entities. Such substitutions substantially alter higher-level abstracted themes, shifting from “Space Exploration” in the case of Mars to “Treasure Hunting” in the latter case. Because no semantic abstraction is performed, this paper exemplifies \mean{1} meaning-making.
\end{itemize}

Below are a few more sample papers with \mean{1} complexity:
\citep{Li2018ASO, Zhong2023ACS, Bekoulis2020ARO, Cheng2024SHIELDLS, Zhou2016JointlyEE}

\subsubsection{\mean{2}: Categorical (themes, topics, patterned organization).}
\begin{itemize}
    \item \textbf{\cite{cemri2025mast}:}
    \begin{figure} [H]
        \centering
        \includegraphics[width=1\linewidth]{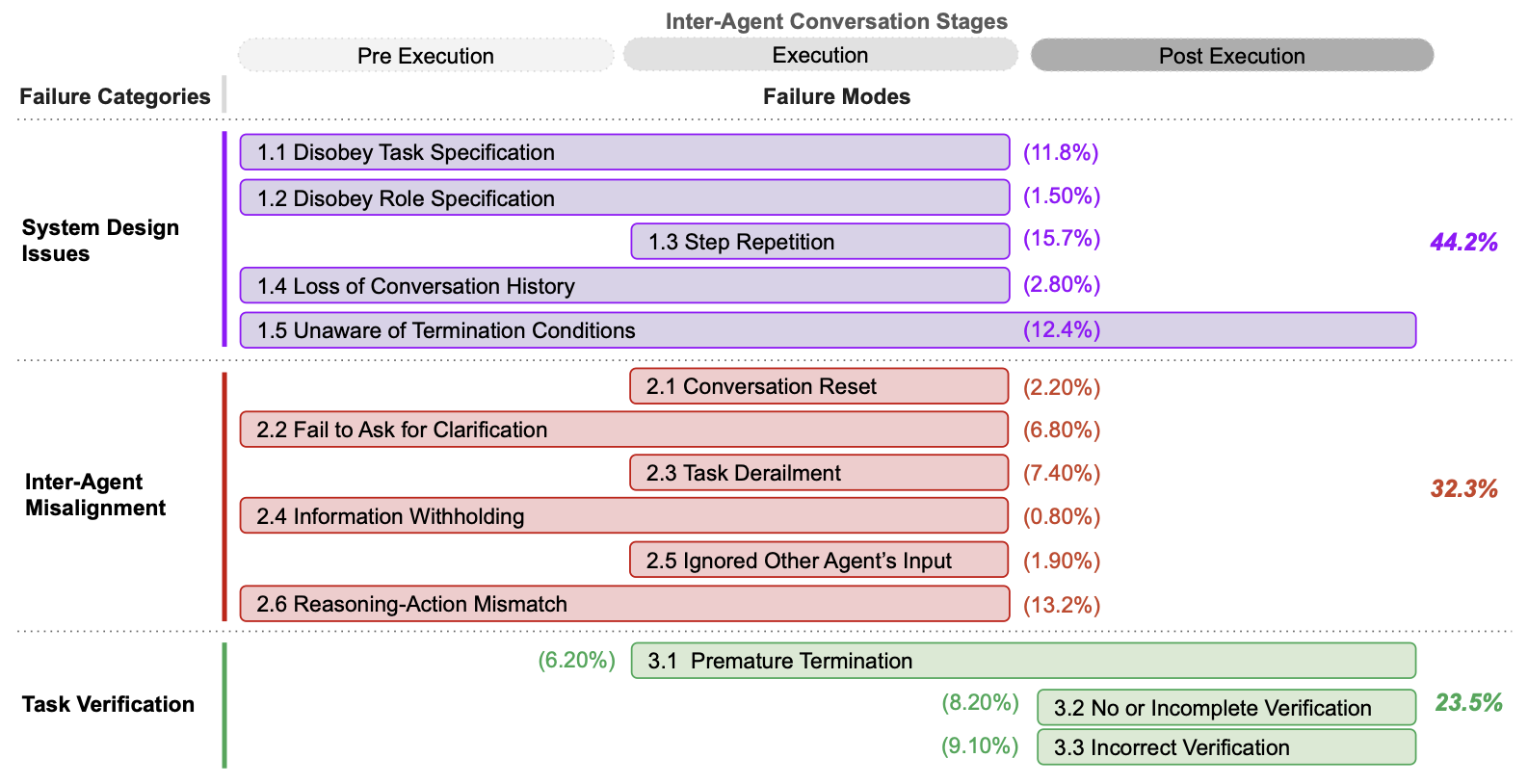}
        \label{fig:connectdots}
        \vspace{-5mm}
    \end{figure}
    \begin{quote}
        Figure 1: MAST: A Taxonomy of MAS Failure Modes. The inter-agent conversation stages indicate when a failure typically occurs within the end-to-end MAS execution pipeline. A failure mode spanning multiple stages signifies that the underlying issue can manifest or have implications across these different phases of operation. The percentages shown represent the prevalence of each failure mode and category as observed in our analysis of 1642 MAS execution traces. Detailed definitions for each failure mode and illustrative examples are available in Appendix A.
    \end{quote}
    This paper proposes some of the failures often encountered in Multi-Agent Systems (MAS), which were uncovered from QA on MAS run trajectories that failed to complete the intended tasks. These themes are the result of observing MAS trajectories and synthesizing general patterns to describe the common failures of MAS. Hence, this work exemplifies \mean{2}, as it involves global semantic knowledge aggregation across instances, but does not introduce interpretive explanations or underlying assumptions beyond those directly supported by the observed data.
    \item \textbf{\cite{Wang2025PolicyPulseLT}:}
        \begin{itemize}
            \item \textit{Parenting in the Digital Age}: Exploring the impact of digital technology on parenting practices and child development
            \item \textit{Work-Life Balance and Parenting}: Investigating the challenges faced by working parents and effective policies for achieving a healthy work-life balance.
            \item \textit{Mental Health Support for Parents}: Examining the importance of mental health resources for parents and strategies to improve access and support.
        \end{itemize}
   Similar to the MAS example, the high-level themes in this paper are also high-level, abstracted descriptions of what is observable in the dataset. Hence, this work is \mean{2}.
\end{itemize}

Below are a few more sample papers with \mean{2} complexity:
\citep{Fang2025DecodingCI, Li2022COVID19CR, Barany2024ChatGPTFE, Edge2024FromLT, Zhong2023ACS}

\subsubsection{\mean{3}: Interpretive (implicit meaning entailed by the frame).}
\begin{itemize}
    \item \textbf{\cite{Fletcher2012AGT}:}
    \begin{quote}
        \textit{\textbf{Positive personality}
            Olympic gold medalists possessed numerous positive person-
        ality characteristics, such as openness to new experiences, consci-
        entiousness, innovative, extraverted, emotionally stable, optimistic, and proactive, which influence the mechanisms of challenge
        appraisal and meta-cognition. The following quote illustrates how
        one champion evaluated missing out on selection for a major international competition in a positive manner, due to his opti-
        mistic and proactive nature:
            There were four of us challenging for these final two places .
        and I got told I was on the reserve list. And at the time it was
        devastating but it’s one of those things; if you don’t take a ticket
        in the raffle, you’re never going to win a prize. So you have to
        take the ticket . that’s part of life and it just makes you think
        'well, what can I do differently to make sure I do get success'?}
    \end{quote}
    This paper explores the common characteristics of being an “Olympic champion” through interviews with Olympic athletes. The interview data consist of narratives about the athletes’ daily lives and training practices. Themes such as “Positive Personality” cannot be derived through simple high-level textual summarization; instead, they require interpretive analysis of the interview material to address more complex questions such as “How is this useful?” or “What mindset characterizes an Olympic champion?” Because generating these higher level themes involves interpretive reasoning beyond descriptive aggregation, this work exemplifies \mean{3}.
    \item \textbf{\cite{Rao2024QuaLLMAL}:}
        \begin{itemize}
            \item \textit{Enhancing Transparency and Explainability}: Drivers are concerned about opaque fare calculations, unclear incentives, and uncertain criteria for earnings, surge pricing, and cancellations...
            \item \textit{Predictability and Worker Agency}: Drivers face unpredictable earnings from fluctuating surge pricing, algorithm changes, increased competition, complex incentive qualifications, and low compensation for long pickups and waits...
            \item \textit{Better Safety and More Time}: Drivers face navigation issues, support access difficulties, challenges with false complaints, and low compensation for additional tasks and wait times...
        \end{itemize}
        The data used in this paper were collected from Reddit forums and consist mainly of conversational snippets on diverse topics. One example from the data is: “I don’t understand what the hell Uber is thinking when giving us long pickups for short trips. NOBODY sane accepts a $\$4$ ride for someone half an hour away.” Mere descriptive abstraction would synthesize themes such as “User confusion with software algorithms,” which focus on what is being said rather than why it is being said. To further understand the underlying message conveyed in the quotation and synthesize meaningful themes such as “Predictability and Worker Agency” or “Better Safety and More Time,” interpretive analysis is required. Because such themes would not be possible without subjective interpretation, this paper provides a clear example of \mean{3}.
\end{itemize}

Below are a few more sample papers with \mean{3} complexity:
\citep{Hou2024PromptbasedAF, Montes2025LargeLM, McKeown2015YoullNW, Goulet2023CommunityTO, Nyaaba2025OptimizingGA}

\subsubsection{\mean{4}: Theoretical (external reframing; etic con-358 structs).}
\begin{itemize}
    \item \textbf{\cite{Carius2024ArtificialIA}:}
    \begin{quote}
        \begin{itemize}
            \item \textit{Health Barriers}: In addition to facing financial hardships, participants in our study expressed how these barriers affected their physical and mental well-being. Insufficient access to medical care further compounded their health concerns, making it challenging for them to achieve stability while dealing with mental and physical health issues. Many mentioned the absence of medical insurance or coverage, particularly among Black citizens...
            \item \textit{Insufficient Community Resources}: In interviews, participants also highlighted the significant challenges they face due to the lack of appropriate resources and infrastructure in their communities. They expressed concerns about the safety of their neighborhoods, the scarcity of well-paying job opportunities, and the need for more activities and community organizing efforts to improve their overall well-being...
            \item \textit{Public Assistance Barriers}: As we alluded to in earlier themes, participants in our study consistently emphasized the challenges they face with public assistance programs and the impact these barriers have on their journey toward financial stability...
        \end{itemize}
    \end{quote}
    This work clearly differentiates between \mean{3} and \mean{4}. While the themes are synthesized through interpretive analysis, the study also incorporates meta-contextual knowledge and draws on multiple domains to fully understand the operational mechanisms of a meta-system. For example, the work discusses the health barriers faced by many financially insufficient individuals, which prevent them from achieving financial stability. Furthermore, one contributing factor to this financial instability is the lack of infrastructure investment needed to attract businesses and provide economic opportunities. Thus, understanding the operation and impact of the meta-system requires recognizing the interconnectedness of multiple domains, including racial barriers, infrastructure development, and socioeconomic conditions. Consequently, this work serves as a strong example of \mean{4}.
    \item \textbf{\cite{Tian2019ExploringTF}:}
        \begin{quote}
            \textit{As a result, seven main categories of factors were identified to influence Shaanxi Blower’s business model innovation. These are market pressure, entrepreneurship, culture and strategy, technology, human resources, organizational capability and government policy. Market pressure and government policy are the direct external factors that promote business model innovation. Entrepreneurship is the direct internal factor that promotes business model innovation. Information technology and technological innovation, as the two technology-driven factors, directly affect business model innovation, but belong to different variables. Information technology together with government policy behavior and market pressure are exogenous variables, and technological innovation together with entrepreneurial spirit are endogenous variables. However, culture and strategy, human resources  and organizational capability are the guarantee factors for business model innovation.}
        \end{quote}
        Similar to the discussion in the first example, understanding the “business more” system in this paper requires analyzing the interconnections among "market pressure, entrepreneurship, culture and strategy, technology, human resources, organizational capability, and government policy". Each of these domains is complex on its own, and together they form a cohesive, well-functioning system whose understanding requires \mean{4}-level analysis. 
\end{itemize}

Below are a few more sample papers with \mean{4} complexity:
\citep{Binder2010UsingGT, Sun2020GreenIR, Beattie2004AGT, Barke2022GroundedCH, Rempel2013ParentingUP}

\subsection{\textcolor{blue}{Dimension 2: Level of Modeling}}

\subsubsection{\model{1}: Static taxonomy and relational configuration models.}
\begin{itemize}
    \item \textbf{\cite{cemri2025mast}:}
    \begin{figure} [H]
        \centering
        \includegraphics[width=1\linewidth]{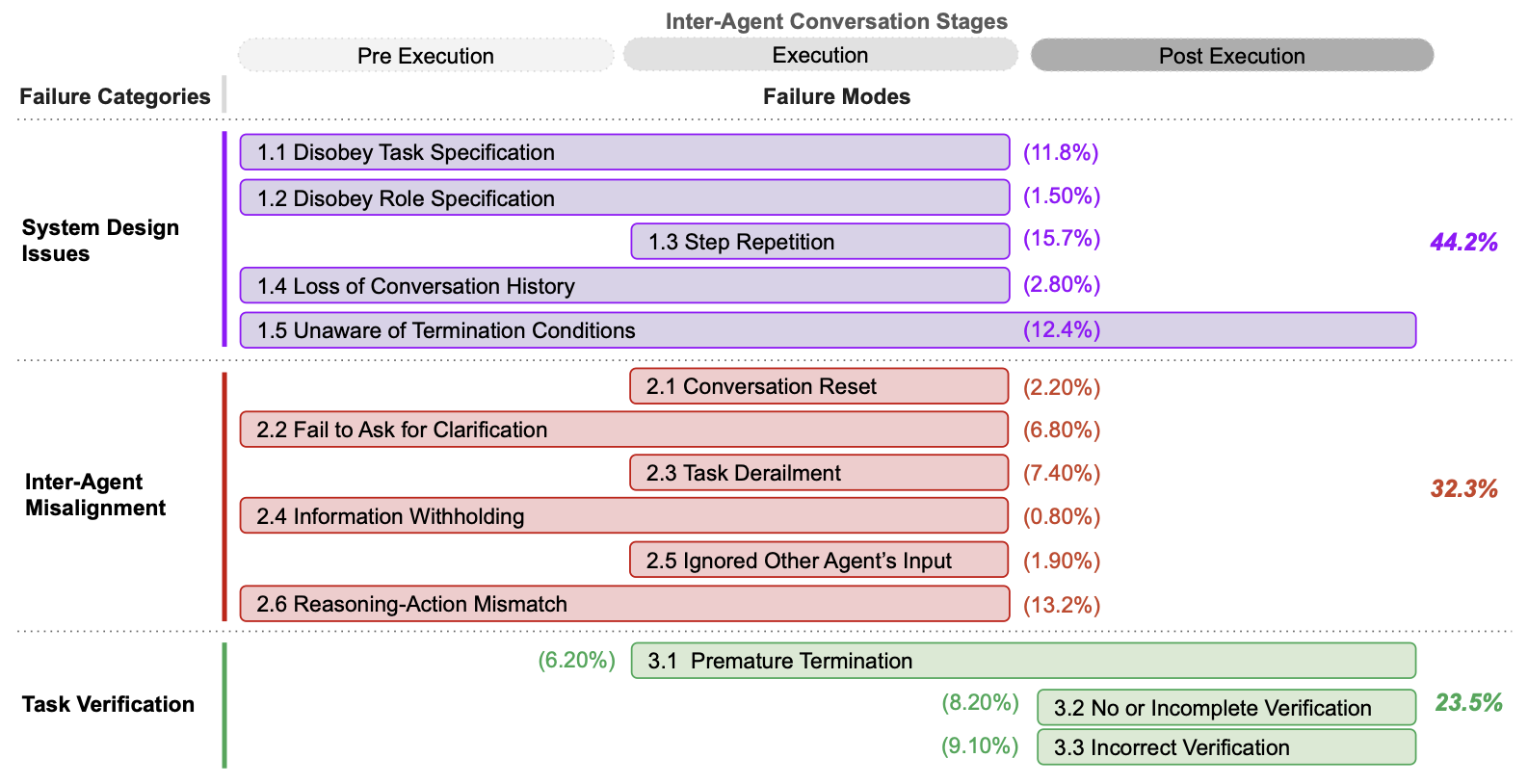}
        \label{fig:connectdots}
        \vspace{-5mm}
    \end{figure}
    \begin{quote}
        Figure 1: MAST: A Taxonomy of MAS Failure Modes. The inter-agent conversation stages indicate when a failure typically occurs within the end-to-end MAS execution pipeline. A failure mode spanning multiple stages signifies that the underlying issue can manifest or have implications across these different phases of operation. The percentages shown represent the prevalence of each failure mode and category as observed in our analysis of 1642 MAS execution traces. Detailed definitions for each failure mode and illustrative examples are available in Appendix A.
    \end{quote}
    The discovered themes are categorized into three main categories (“System Design Issues,” “Inter-Agent Misalignment,” and “Task Verification”) as specified in the paper. Since this constitutes a hierarchical organization of themes and patterns, this is an example of \model{1}.
    \item \textbf{\cite{Zhong2025HICodeHI}:}
        \begin{figure} [H]
            \centering
            \includegraphics[width=1\linewidth]{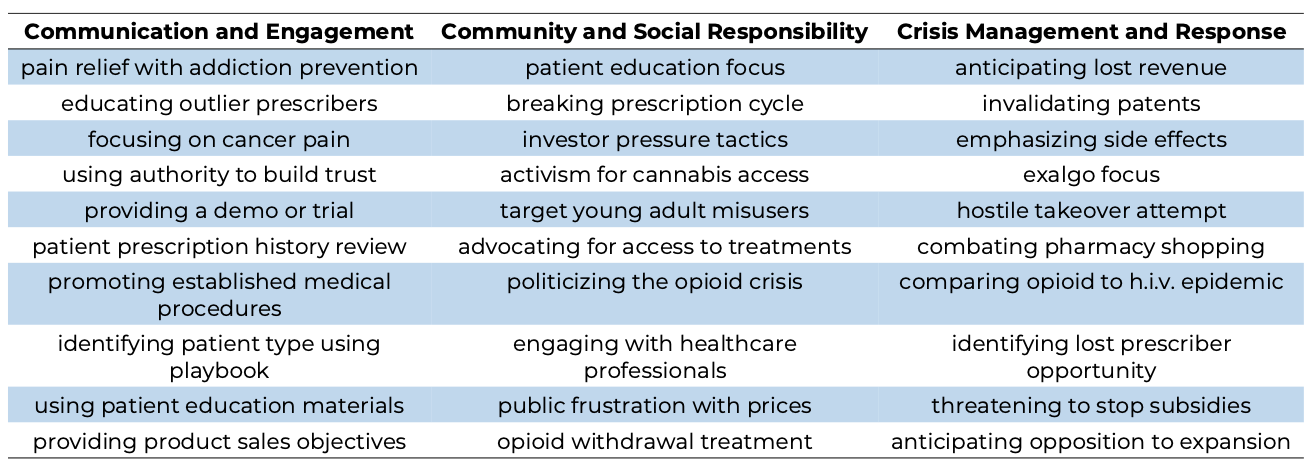}
            \label{fig:connectdots}
            \vspace{-5mm}
        \end{figure}
        \begin{quote}
            Figure 3: Three themes with randomly selecting cluster labels generated by HICode over OIDA.
        \end{quote}
        Similar to the discussion in the first example, the themes are organized in a hierarchical taxonomy, with one category being “Communication and Engagement".
\end{itemize}

Below are a few more sample papers with \model{1} complexity:
\citep{pecoraro2014conflicting, serkina2019administrative, jappinen2022theory, Gao2025EfficiencyWR, Lam2024ConceptIA}

\subsubsection{\model{2}: Stage/phase/timeline models (time without causality).}
\begin{itemize}
    \item \textbf{\cite{Li2023OpenDomainHE}:}
    \begin{figure} [H]
        \centering
        \includegraphics[width=0.7\linewidth]{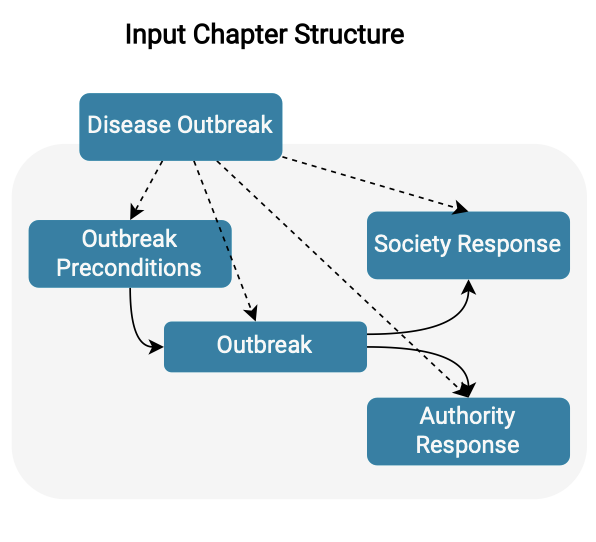}
        \label{fig:opendomains}
        \vspace{-5mm}
    \end{figure}
    \begin{quote}
        Figure 2: To create the schema for a given scenario, our model follows 3 rounds of operation: (1) event skeleton construction where we ask the LLM to list the important events; (2) event expansion to discover more related events for each existing event; event-event relation verification where we update the event-event relations based on the LLM’s answers to questions about each event pair.
    \end{quote}
    Figure 2 in this paper offers a glimpse of the output produced by the proposed QA framework, which organizes extracted events into a timeline with transition edges. This is an example of \model{2}, as it incorporates a notion of time by ordering events in a sequential flow. However, it does not reach \model{3}, because the relationships between events are temporal transitions rather than causal relationships.
    \item \textbf{\cite{Li2020ConnectingTD}:}
    \begin{figure} [H]
        \centering
        \includegraphics[width=0.9\linewidth]{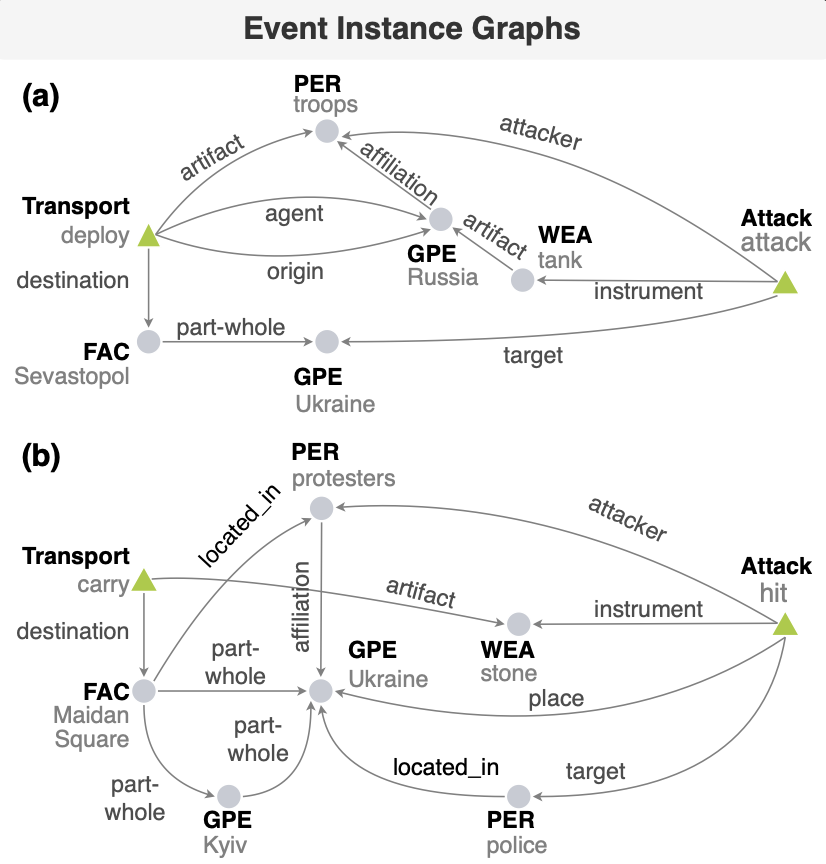}
        \label{fig:connectdots}
        \vspace{-5mm}
    \end{figure}
        \begin{quote}
            Figure 1: The framework of event graph schema induction. Given a news article, we construct an instance graph for every two event instances from information extraction (IE) results. In this example, instance graph (a) tells the story about Russia deploying troops to attack Ukraine using tanks from Russia; instance graph (b) is about Ukrainian protesters hit police using stones that are being carried to Maidan Square. We learn a path language model to select salient and coherent paths between two event types and merge them into a graph schema. The graph schema between ATTACK and TRANSPORT is an example output containing the top 20
        \end{quote}
        Figure 1 of this paper also provides a glimpse of the temporal structure of unfolding events, such as “Russia deploying troops to attack Ukraine” and “Ukrainian protesters hitting police.” For each unfolding event, the temporal graph illustrates how the event progresses and how subevents transition from one to another. However, no causal relationships are specified between the events or subevents. Hence, this is an example of \model{2}.
\end{itemize}

Below are a few more sample papers with \model{2} complexity:
\citep{lunney2025resilience, anggraini2023crypto, skryabina2016child, kristelstein2022brain, tinder2022rewards}

\subsubsection{\model{3}: Causal dependency and mechanism models (directed influence).}
\begin{itemize}
    \item \textbf{\cite{Du2022RESIN11SE}:}
    \begin{figure} [H]
        \vspace{-10mm}
        \centering
        \includegraphics[width=0.7\linewidth, angle=270]{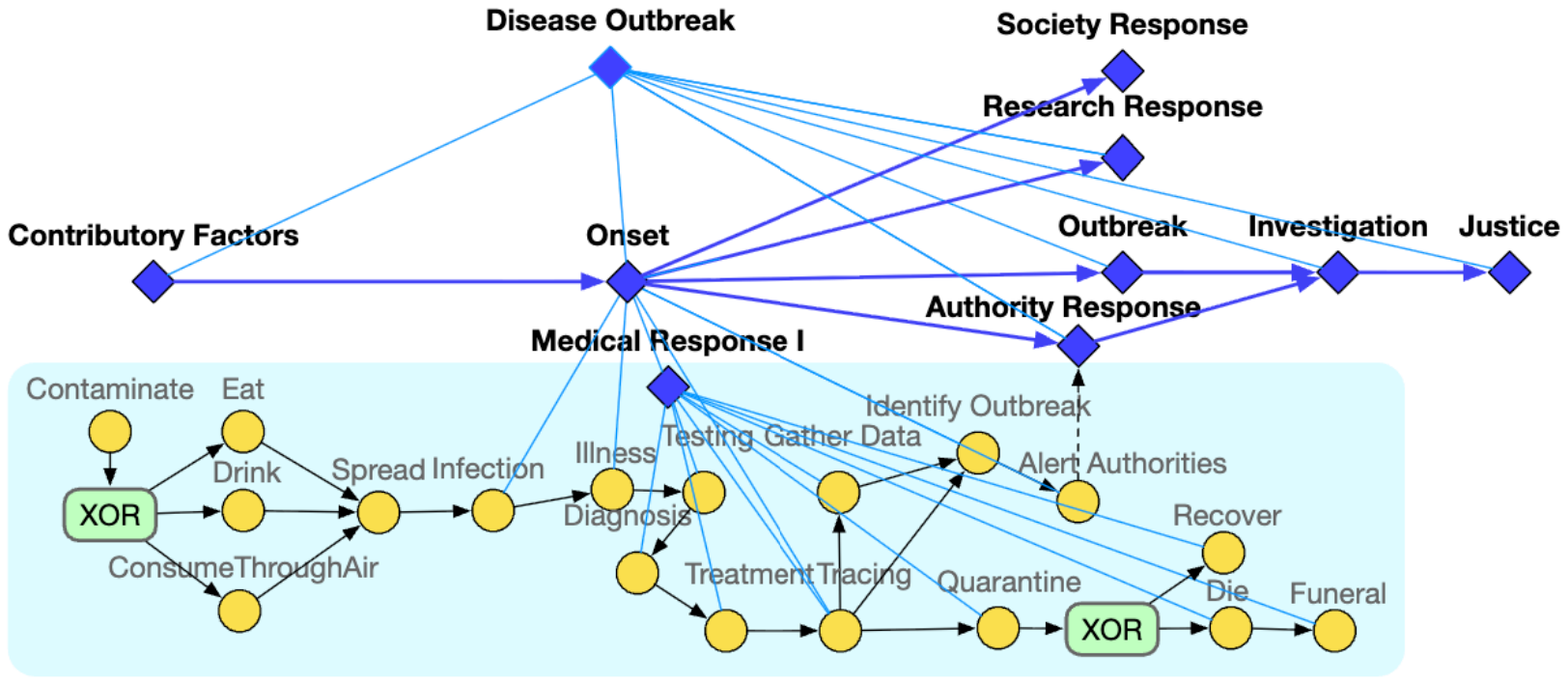}
        \label{fig:connectdots}
        \vspace{-10mm}
    \end{figure}
        \begin{quote}
            Figure 4: The curated schema for the disease outbreak scenario. Blue diamond shapes represent sub-schemas and yellow circles represent primitive events. Black arrows between primitive events represent temporal order, light blue lines between the primitive events and the sub-schema node represent event-subevent hierarchical relationship. Here we only show the primitive events under the Onset sub-schema
        \end{quote}
    This paper is about event schema extraction and prediction. Its figure 4 shows one example of extracted schema, in which the events are connected by both temporal order (black line) and causal relationship (XOR operation). We view this as a partial causal graph and label this paper as \model{3}.
    \item \textbf{\cite{Safa2024AMF}:}
    \begin{figure} [H]
        \vspace{-5mm}
        \centering
        \includegraphics[width=0.7\linewidth]{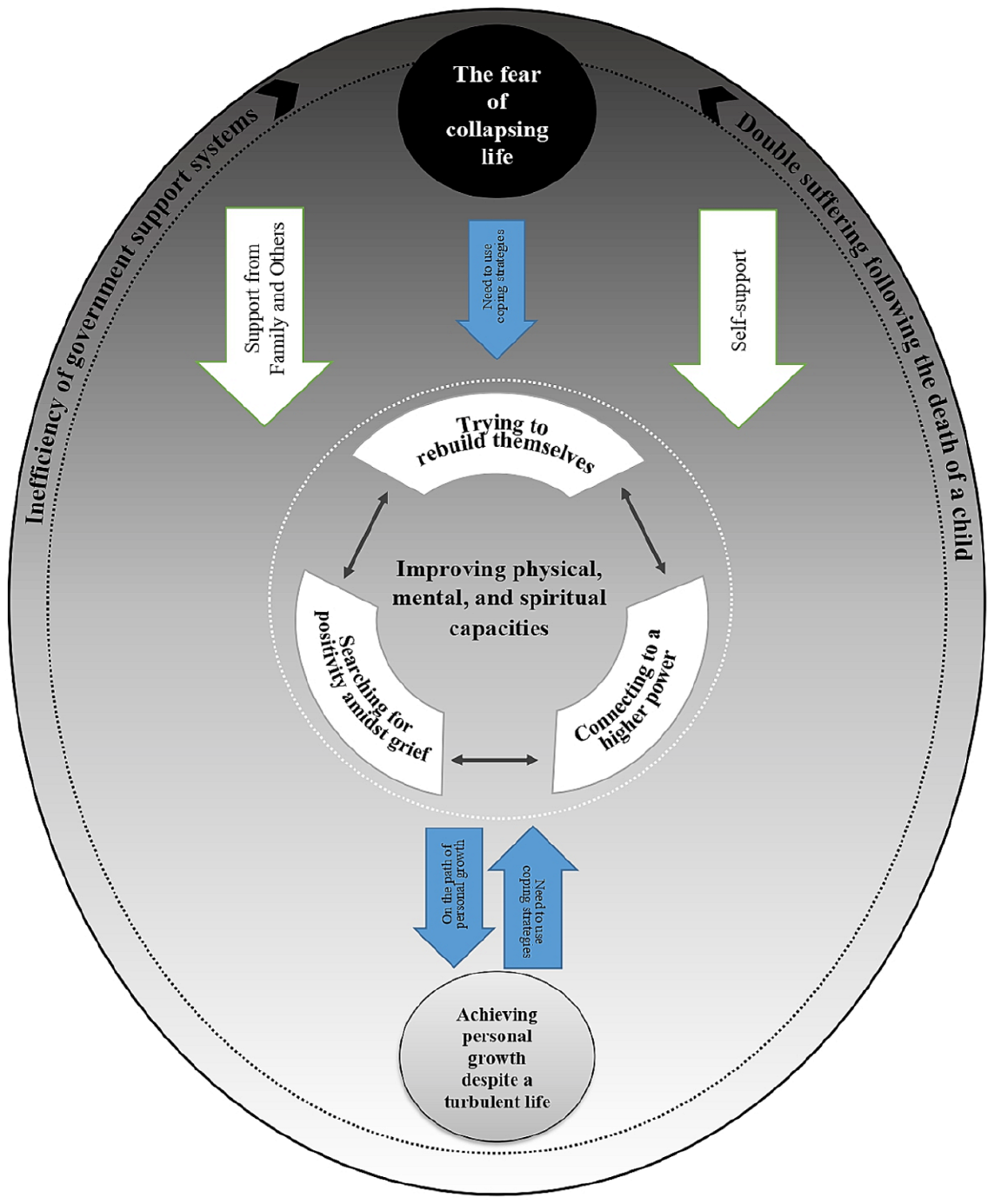}
        \label{fig:connectdots}
        \vspace{-5mm}
    \end{figure}
        \begin{quote}
            Figure 1 A schematic model “from fear of collapsing life to achieving personal growth”
        \end{quote}
        This paper represents a directed pathway structure centered on how older adults move from 'fear of life collapse'. The model is explicitly structured around directed dependence. Figure 1 and the textual description in the 'Explanation of the model' section make clear that nodes represent factors/conditions/strategies (not persistent system components), and edges encode how elements contribute to, enable, or lead to others. The model's explanatory force comes from clarifying which factors contribute to the main concern, which strategies address it, and what facilitates movement toward the outcome-classic \model{3} directed pathway logic, not \model{4} loopy component interdependency.
\end{itemize}

Below are a few more sample papers with \model{3} complexity:
\citep{Bratianu2020TowardUT, Kendellen2019ApplyingIL, agyeman2014role, attwell2022convergence, Chopra2019IndianSM}

\subsubsection{\model{4}: ynamical systems / feedback / complex systems models (state + update + iteration).}
\begin{itemize}
    \item \textbf{\cite{Kendellen2019ApplyingIL}:}
    \begin{figure} [H]
        \vspace{-1mm}
        \centering
        \includegraphics[width=0.7\linewidth, angle=270]{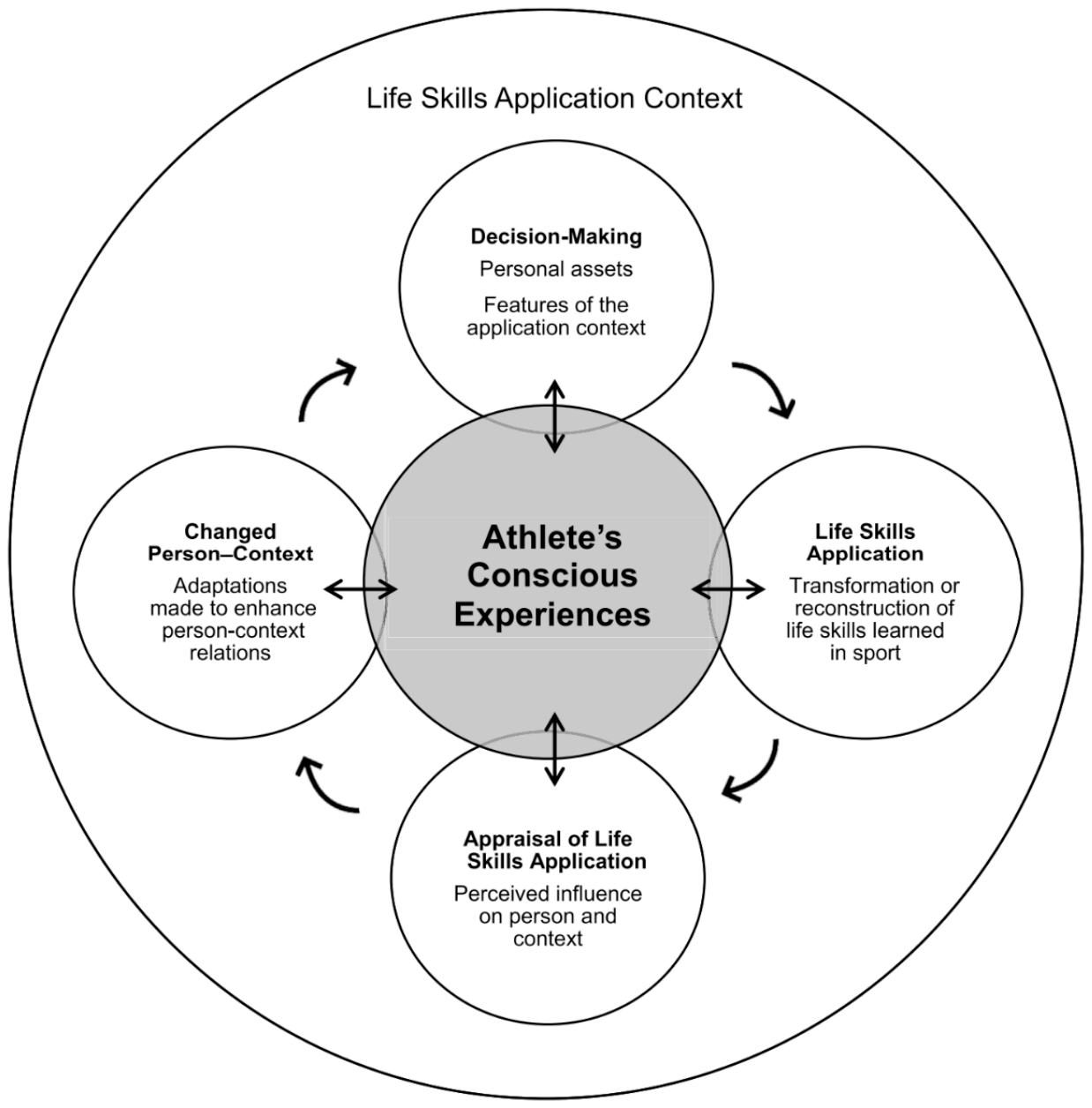}
        \label{fig:opendomains}
        \vspace{-5mm}
    \end{figure}
    \begin{quote}
        Figure 1. Grounded theory of life skills application.
    \end{quote}
    The paper produces a cyclical, evolving system of person-context regulations. The authors define the process as having 'no definitive end point,' instead functioning as a continuous loop of decision-making, application, appraisal, and adaptation. The topology is defined by mutual coupling where the 'changed person-context' statefully feeds back into future decision-making cycles. Unlike a \model{2} or \model{3} model which would move from sport (source) to life (sink), this model treats the athlete and their environment as a dynamic system where knowledge and assets are continuously re-regulated across domains. So we annotate this as \model{4}.
    \item \textbf{\cite{Bratianu2020TowardUT}}
    \begin{figure} [H]
        \vspace{-5mm}
        \centering
        \includegraphics[width=0.7\linewidth, angle=270]{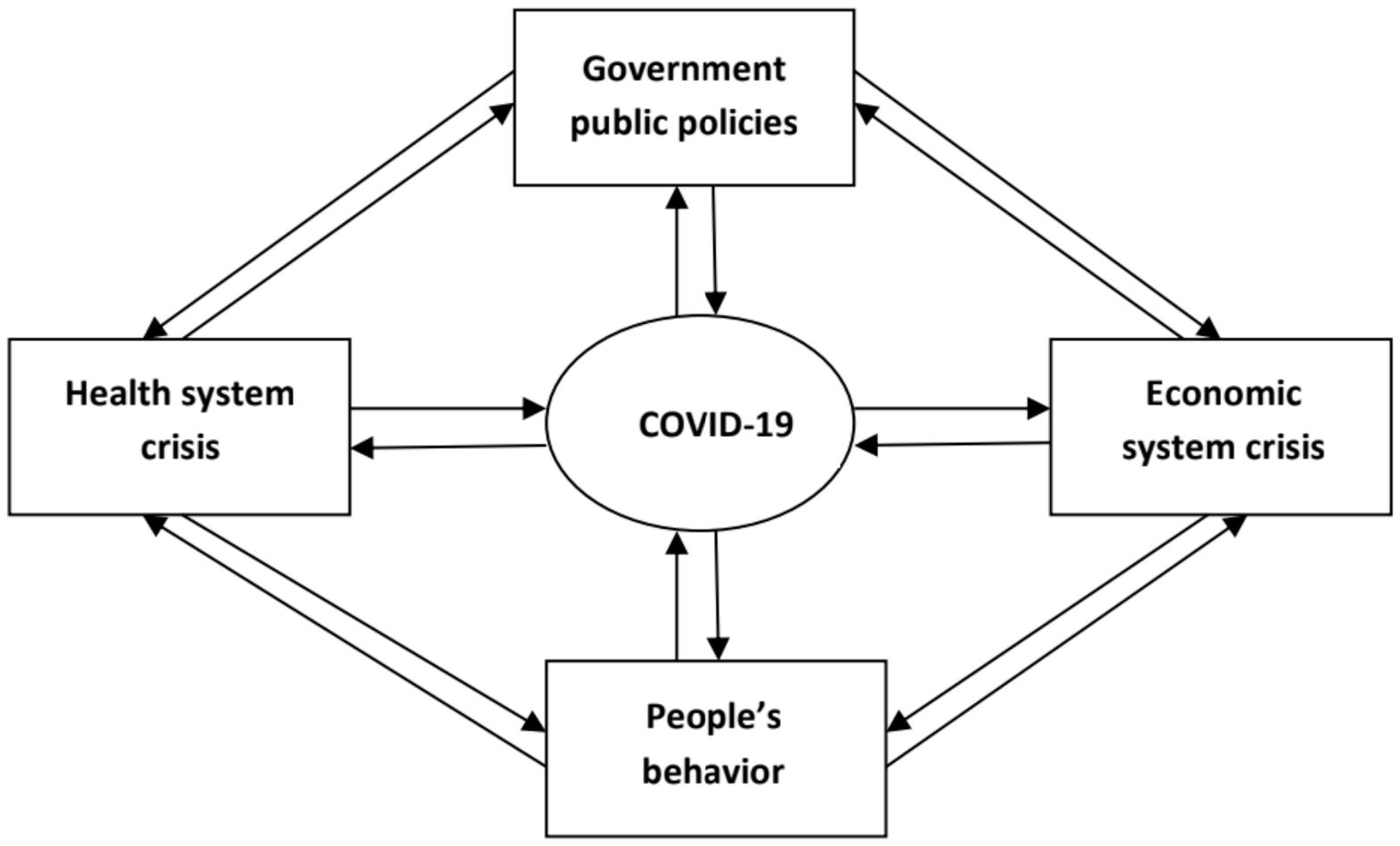}
        \label{fig:connectdots}
        \vspace{-8mm}
    \end{figure}
        \begin{quote}
            Figure 1. An integrated model of the COVID-19 pandemic dynamics concerning the health and economic systems Source: Author’s own research
        \end{quote}
    Based on the figure 1 of this paper, it explicitly defines the COVID-19 pandemic as a complex system of 'mutual coupling' and 'dense interdependencies' rather than a linear or directed sequence. The diagram shows correlations that 'act in all possible directions.' This violates the Source/Sink test (\model{3}) because there is no start-to-end flow. Instead, the model represents a state-based system where variables continuously influence and are influenced by the crises in a feedback web. The rationale for is rooted in this absence of a terminal state and the presence of iterative, stateful interdependency, which leads to \model{4} annotation.
\end{itemize}

Below are a few more sample papers with \model{4} complexity:
\citep{karakas2019spirals, Coleman2017TechnologicalIL, fellahi2021rural, Kendellen2019ApplyingIL, Loosemore1999AGT}

\section{Annotation Study Details}
\label{app:annotation-details}

\subsection{Research Paper Type Classification Prompt}
\begin{Verbatim}[breaklines,breakanywhere,fontsize=\small]
You are a careful research assistant. Classify papers using only the provided fields. Do not invent details. If evidence is insufficient, choose Unclassified. Follow the label schema exactly and output valid JSON only.
User message template:
You will be given a academic research paper.
 Task: classify the paper into exactly one of these top-level categories:
Conceptual & Theoretical
 Empirical - xxx (xxx must be one of the empirical method labels listed below)
 Computational
 Review & Synthesis
 Quantitative
 Unclassified
Definitions and decision rules (apply in order)
A. Review & Synthesis
 Choose if the primary contribution is synthesizing prior literature rather than reporting new primary data. Strong cues: “systematic review”, “scoping review”, “meta-synthesis”, “qualitative evidence synthesis”, “integrative review”, “realist review”, “narrative review” (when it summarizes literature).
B. Computational
 Choose if the primary contribution is building/evaluating computational methods/tools to automate or computationally support qualitative research (e.g., NLP for coding, LLM-assisted thematic analysis, automated discourse analysis pipelines). It may include experiments, but the core is the computational framework/tool.
C. Quantitative
 Choose if the paper’s primary methodology emphasizes statistical analysis and modeling rather than the qualitative methods listed below. Strong cues include: regression analysis, causal inference, time series analysis, statistical modeling, hypothesis testing, significance tests (p-values), confidence intervals, effect sizes, large-N survey econometrics, predictive modeling as the main analysis, randomized experiments analyzed primarily quantitatively, etc.
 If the paper uses interviews/focus groups only as minor context but the main results are statistical/model-based, still choose Quantitative.
D. Empirical - xxx
 Choose if the paper reports collecting and analyzing real-world data about a phenomenon (participants, field/site, texts, artifacts, records), and qualitative analysis is central. Label must be one of:
Empirical - Interpretative Phenomenological Analysis
 Subcategories (pick one in rationale): lived experience of illness/care; professional identity/role; sensemaking/transition; trauma/grief; other.
 Cues: “IPA”, “interpretative phenomenological”, idiographic focus, small samples.
Empirical - Grounded Theory
 Subcategories: classic/Glaserian; Straussian; constructivist (Charmaz); other/unspecified.
 Cues: “grounded theory”, “theory generation”, “constant comparative”, “theoretical sampling”, “saturation”.
Empirical - Thematic Analysis
 Subcategories: reflexive TA; codebook TA; framework-informed TA; other/unspecified.
 Cues: “thematic analysis”, “themes were identified”, Braun & Clarke.
Empirical - Narrative Analysis
 Subcategories: life story/biographical; illness narratives; organizational narratives; other/unspecified.
 Cues: “narrative analysis”, “stories”, “plot”, “narratives”, temporality.
Empirical - Discourse Analysis
 Subcategories: conversation analysis; critical discourse analysis (CDA); discursive psychology; other/unspecified.
 Cues: “discourse analysis”, “discursive”, “CDA”, “conversation analysis”, turn-taking, rhetoric.
Empirical - Dialogue Analysis
 Subcategories: interactional/dialogic analysis; clinical dialogue; collaborative/team dialogue; other/unspecified.
 Cues: “dialogue analysis”, “dialogic”, “interaction episodes”.
Empirical - Framework Analysis
 Subcategories: policy/service evaluation; health services; implementation/QI; other/unspecified.
 Cues: “framework analysis”, matrix-based charting, applied policy research.
Empirical - Ethnography
 Subcategories: classic ethnography; focused ethnography; digital/virtual ethnography; institutional ethnography; other/unspecified.
 Cues: “ethnography”, “participant observation”, fieldnotes, prolonged engagement, site culture.
Empirical - Content Analysis
 Subcategories: qualitative content analysis; directed content analysis; summative content analysis; other/unspecified.
 Cues: “content analysis”, coding categories, frequency counts may appear but still qualitative.
Empirical - OTHERS
 Subcategories: case study (qualitative); phenomenology (non-IPA); interpretive description; template analysis; rapid qualitative analysis; participatory/action research; mixed-methods (qual-dominant); other.
 Use only if none of the above methods fit but it is clearly qualitative empirical work.
E. Conceptual & Theoretical
 Choose if the paper’s primary contribution is conceptual/philosophical/methodological discussion (arguments, frameworks, critiques) without primary data collection and not a review/synthesis.
F. Unclassified
 Choose if the provided fields are too sparse/ambiguous to justify A–E.
Output format (STRICT)
Return valid JSON only, with:
 rationale (100~200 words, grounded in evidence from the provided fields; mention what text triggered the decision; note uncertainties)
 decision (exact string from allowed labels)
Below is the paper you will classify:
=======================
\end{Verbatim}

\subsection{Level of Meaning-Making Prompt}
\label{meaning-prompt}
\begin{Verbatim}[breaklines,breakanywhere,fontsize=\small]

You are an expert qualitative researcher trained in multiple qualitative methodologies (e.g., Straussian grounded theory, constructivist grounded theory, thematic analysis, interpretative phenomenological analysis, discourse analysis, narrative analysis, content analysis). Your task is to evaluate a given qualitative research paper and judge the primary level of meaning-making at which the paper operates.
Critical clarifications (read carefully and apply strictly):
Methodology is not Level 4. The use of qualitative methods and coding procedures (e.g., open/axial/selective coding, constant comparison, memoing, IPA steps, thematic analysis workflows, discourse-analytic techniques, codebooks, inter-coder reliability) does NOT by itself constitute Level 4 meaning-making. Level 4 is about the substance of the interpretive claims and what they depend on, not which qualitative methodology is used.
Theory citation is not Level 4 unless it is applied. Merely citing theories/frameworks as background, literature comparison, positioning, or discussion (e.g., introduction, related work, limitations, implications, future directions) does NOT qualify as Level 4. A paper qualifies as Level 4 only when external theory/frameworks/domain knowledge outside the corpus are actually used as interpretive machinery during coding, analysis, or theory development to generate or justify core findings.
Level 4 requires external interpretive resources beyond the corpus. To qualify as Level 4, the paper must rely on concepts, constructs, models, typologies, or explanatory systems that are not entailed by the corpus’s semantic frame alone and that typically require domain expertise or specialized theoretical knowledge. These external resources must actively shape how codes/themes are constructed, linked, or elevated into the paper’s central claims.
Level 3 is not constrained to a fixed checklist. When judging Level 3, do not treat any list of sub-dimensions (e.g., teleology, emotion, causation, social meaning, norms, implicit assumptions, temporal structures,  pragmatics) as exhaustive. Level 3 can surface any kind of latent content that is implicitly entailed by the semantic frame of the data. The rule of thumb is: Level 3 uncovers non-trivial & interesting implicit/hidden meaning relevant to the research topic; Level 2 primarily condenses and organizes what is already explicit into categories.
Your goal is to infer what kind of meaning the paper primarily produces from its data: surface description, categorization, interpretive surfacing of latent meaning, or theory-based reframing using applied external frameworks.
Definitions of the four levels of meaning-making:
Level 1: Descriptive level of meaning-making
The paper primarily restates, summarizes, or reports what is explicitly present in the data. It stays close to participants’ words or observable events. Outputs are concrete descriptions of actions, experiences, events, and statements without abstraction beyond paraphrase or recounting. There is no systematic grouping into higher-level categories and no inference about implicit meaning. The central analytic move answers: what is said or observed?
Level 2: Categorical level of meaning-making
The paper primarily organizes descriptive observations into categories, themes, domains, or types. It condenses multiple surface details into higher-level labels that summarize recurring patterns. The categories remain largely faithful to the frame of the data and do not systematically surface latent motives, hidden meanings, implicit norms, emotions, assumptions, or causal structures. The central analytic move answers: what kind of phenomenon or recurring pattern is present?
Level 3: Interpretive level of meaning-making
The paper primarily brings forward implicit or latent meanings that are not explicitly stated in the data but are entailed by the semantic frame of the situations described. It uses reasoning and inference grounded in the text to articulate what is implicitly going on (e.g., goals/purposes, emotions, causal relations, social meaning, normative expectations, functional roles, assumptions, relational/coordination structure, pragmatic intent), but it is not limited to these; any latent content that is warranted by the data’s frame may qualify. Level 3 does not depend on specialized external theories or domain-specific conceptual systems. The central analytic move answers: what is implicitly committed or implied by the data, beyond categorization?
Level 4: Theoretical level of meaning-making
The paper’s core analysis relies on external theories, frameworks, or domain-specific conceptual systems that are outside the corpus itself and not entailed by its semantic frame alone. These external interpretive resources are actively applied during coding, analysis, or theory development to generate, structure, or justify the main findings (e.g., coding through a theoretical lens, mapping data onto a formal model/typology, interpreting instances as manifestations of a named theoretical construct, using specialized constructs as the organizing backbone of themes). Crucially, merely mentioning or comparing to theory in the introduction/discussion does not qualify; the external framework must function as an analytic engine that shapes the results. The central analytic move answers: what broader conceptual system explains or re-situates the data as an instance of a larger theory?
Decision instructions:
A) Judge the primary level across the paper, not isolated passages. If multiple levels appear, choose the level that best characterizes the dominant analytic contribution (the main “value add”).
B) Do not classify as Level 4 simply because the paper uses a named qualitative methodology, uses the term “theory,” or includes theoretical citations.
C) Classify as Level 4 only if external theory/framework/domain knowledge outside the corpus is materially used to construct or justify codes, themes, relationships among themes, or the paper’s central explanatory claims.
D) If the paper mainly reports themes/categories without systematically surfacing latent meaning, prefer Level 2. If it systematically surfaces implicit meaning beyond categorization using reasoning grounded in the data, prefer Level 3.
E) In your rationale, cite concrete indicators from the paper’s analytic moves: how codes/themes are formed, what kinds of claims are made, and whether external theory is applied as an analytic mechanism versus merely cited for context.
Output format requirements:
Return valid JSON with exactly two fields and no additional text:
{
"judgement": "Level 1" or "Level 2" or "Level 3" or "Level 4",
"rationale": "Concise but specific justification grounded in the definitions above, explicitly distinguishing methodology-only references from applied external theoretical frameworks, and explaining why the paper’s primary meaning-making level is the chosen one."
}
\end{Verbatim}

\subsection{Level of Modeling Prompt}
\label{modeling-prompt}
\begin{Verbatim}[breaklines,breakanywhere,fontsize=\small]
You are an expert qualitative researcher and methodologist familiar with how qualitative papers represent models (e.g., taxonomies/typologies, configuration/concept maps, stage/process models, causal graphs and DAG-style pathway models, workflows/finite state machines, and complex systems / systems-dynamics style models). Your task is to evaluate a given qualitative research paper and judge the primary level of modeling at which the paper operates.
Critical clarifications (read carefully and apply strictly):


Methodology is not a modeling level. The use of grounded theory, thematic analysis, IPA, discourse analysis, codebooks, inter-coder reliability, memoing, etc. does NOT determine the level of modeling. Judge the level only from the structure of the model the paper produces (figures, tables, or core result text).

A diagram is not required, but modeling commitments must be explicit. A paper may have no figure and still be any level if it clearly presents a taxonomy/configuration, a stage model, a directed dependence model (DAG/workflow), or a complex-systems interdependency model.

Terminology is not decisive; semantics are. Authors may use “dynamic,” “system,” “model,” “framework,” or “causal” loosely. Classify based on what nodes/edges mean and what the model is designed to explain, not the label the authors attach.
Loops alone are not sufficient for Level 4, but “stateful stock–flow” is not required either. Level 4 is not restricted to numeric stocks/flows or continuous quantities. Level 4 is awarded when the paper’s model is best understood as a complex system of interacting components with rich interdependencies and transaction mechanisms, typically featuring multiple loops and no single end-to-end pathway structure.

Workflows and end-to-end pipelines are not Level 4 by default. A model can use “system components/modules” and still be Level 3 if it is essentially an input→output process with largely one-directional dependencies, clear start/end points, and limited feedback structure. Use the presence of clear sources/sinks and a dominant forward path as a strong Level 3 signal.

Source/sink test is a key discriminator. If the model has clear source nodes (starts) and sink nodes (ends)—i.e., it can be read as a pipeline with inputs and outputs—that strongly indicates Level 3 (even if complex). Level 4 system interdependency models typically lack a single clear start or end and are organized around mutual coupling and loops.

Level 3 causal graphs are mostly DAG-like pathway models. Their primary purpose is usually to identify which factors/events contribute to which outcomes and to clarify pathways/mediators. Even if cycles appear, the model’s focus is not primarily feedback/interdependency structure of the system as a whole.

Level 4 models privilege system components and their interaction structure. Level 4 does not require explicit “state-update” semantics at the finest level; instead, it requires (i) persistent system components as the modeling units, (ii) concrete depictions of transactions/interactions among components, and (iii) a loopy/interdependent structure where feedback is central to how the system is represented (not merely an incidental cycle).
Your goal is to infer what kind of model the paper primarily produces: static structure, staged progression, directed pathway/transition logic (mostly DAG/workflow), or complex-systems interdependency modeling.

=============
Definitions of the four levels of modeling:
Level 1: Taxonomy and Configuration Models
The paper’s primary model is a static classificatory structure: a taxonomy/typology, descriptive conceptual map, component/configuration diagram, or topological organization. It organizes entities by hierarchy, grouping, part–whole structure, or simple association.
Level 1 models do not support analytic inference about relationships among categories beyond classification or co-presence. They do not systematically compare, align, contrast, or integrate categories across dimensions, cases, or levels in order to identify patterns, tensions, or higher-order relational structure.
The model does not represent temporal progression, staged movement, directed pathway logic, or system feedback/interdependency as its core. Connections indicate descriptive organization rather than analytic relations (e.g., dependency, opposition, alignment, or interaction).
The central modeling move answers: what exists, and how is it descriptively organized, rather than how elements relate analytically, change over time, or shape one another.
Level 2: Stage / Transition Models
The paper’s primary model is a discrete, ordered process described as movement through qualitatively distinct stages or phases (Step 1, Step 2, Step 3, …). Time is represented via explicit stage boundaries or milestones, and progression is defined in terms of entering, occupying, or exiting stages.
Stages are the representational primitive: they are treated as temporally ordered segments rather than analytically distinct social states or interacting components. Transitions primarily describe what typically happens next, not why movement occurs under specific conditions.
The model may be linear or may include relapse, looping, or return to earlier stages, but such recurrence is descriptive rather than explanatory. Movement between stages is not modeled as conditional on structured dependencies, mechanisms, or interacting sub-processes.
The model does not specify directed causal pathways among multiple factors, nor does it articulate feedback mechanisms or mutual interdependence among system components.
The central modeling move answers: what are the stages, and how does an entity progress through them over time?
Level 3: Directed Pathway Models (Mostly DAG Causal Graphs and Workflows/FSMs)
The paper’s primary model commits to directed dependence in a way that reads as a pathway, pipeline, or transition-logic structure. Directionality is analytically meaningful: edges specify that certain factors, events, or states enable, constrain, mediate, or lead to others.
• Causal graphs (mostly DAG-like): nodes represent factors, events, or variables, and directed edges represent which elements contribute to which outcomes (A → B). Models commonly include branching pathways, mediators, and identifiable end outcomes. Even if limited cycles appear, the overall structure retains recognizable sources and sinks and is used primarily to clarify contribution pathways, not to depict the system as mutually coupled.
• Workflows / FSMs: nodes represent steps, states, or modules, and edges represent conditional or sequential dependencies (e.g., if/then logic, decision points). The structure typically has clear inputs/starts and outputs/ends (or a dominant forward flow). Loops usually indicate rework, retry, or iteration within a pathway, rather than feedback that reorganizes the system as a whole.
Directed dependence—not temporal ordering alone—is the representational primitive. While time or sequence may be involved, the model’s analytic force comes from specifying which elements depend on which others, and under what conditions.
The model does not treat persistent system components as mutually shaping one another through dense interdependence, nor does it center feedback as a primary explanatory device.
The central modeling move answers: what directed pathways, dependencies, or transition logic link inputs or contributing factors to outcomes?
Level 4: Complex Systems Interdependency Models (Loopy component coupling + transactions)
The paper’s primary model treats the phenomenon as a system of interacting components whose behavior arises from dense interdependencies and concrete interaction/transaction mechanisms, rather than from a single pathway linking sources to sinks. The model is organized around multiple feedback loops, not a dominant forward progression.
• Units are persistent system components (e.g., institutions, policies, infrastructures, capacities, collective behaviors, constraints, resources, modules) that endure and interact over time, rather than isolated events, episodic factors, or stages in a pipeline.
• Edges represent concrete interactions or transactions among components (e.g., mutual influence, constraint or enablement, coupling, co-determination, reinforcement or balancing effects, cross-impacts). These interactions are treated as explanatory, not merely illustrative.
• The structure is characteristically loopy and interdependent: there is no single privileged “start” or “end,” and feedback is structurally central to how the phenomenon is represented and explained, not an incidental add-on.
• System behavior is explained via internal coupling, such that changes in one component alter the conditions under which other components operate. The model aims to account for emergent properties (e.g., nonlinearity, amplification, cascading effects, stability, resilience, or breakdown), even if numeric stocks/flows or formal dynamics are not specified.
The model cannot be meaningfully reduced to a staged progression or directed pathway without losing its explanatory intent.
The central modeling move answers: how do interacting system components mutually shape one another through interdependencies and feedback structure?
Decision instructions:
A) Judge the dominant modeling contribution across the paper.
 Assess the paper’s primary substantive model of the phenomenon, not isolated figures, illustrative examples, or secondary discussions. If multiple model forms appear, classify according to the model that carries the main explanatory burden of the results.
B) Do not assign Level 4 based on language or surface cues.
 Terms such as dynamic, system, iterative, complex, or citations to systems thinking are not sufficient. A paper qualifies for Level 4 only if its core model explicitly represents:
persistent system components,
concrete interaction or transaction mechanisms among those components, and
multiple feedback loops that are central to explanation.
 A single loop, recursive step, or metaphorical system reference is insufficient.
C) Apply the source–sink test as a strong discriminator.
 If the model has identifiable starting conditions and end outcomes—and can be read as a start-to-end pathway, pipeline, or transition logic—strongly prefer Level 3, even if the pathway is complex or includes limited cycles.
D) Prefer Level 4 only when mutual coupling is the primary representational logic.
 Assign Level 4 only when the model is best understood as a mutually coupled system:
no single dominant trajectory or privileged start/end point,
dense interdependencies rather than mostly one-directional influence, and
explicit depiction of interaction mechanisms that explain system behavior through feedback.
E) Ground the rationale in concrete representational features.
 In the justification, explicitly reference:
what nodes represent (categories vs stages vs events/variables vs persistent system components),
what edges encode (association vs temporal progression vs directed dependence vs interaction/transaction),
whether the structure is pathway-like with sources/sinks or system-like with pervasive coupling, and
whether feedback/interdependency is structurally central or merely incidental.
IMPORTANT (algorithm vs domain): Judge the level of modeling based on the paper’s substantive model of the target phenomenon, not the computational pipeline. Multi-step LLM workflows (generation → clustering → evaluation), iterative loops, or module diagrams describe the method, not the domain model. If the main output is themes/categories/hierarchies of labels, classify as Level 2 (or Level 1 if it’s mostly descriptive with no real categorization). Only assign Level 3/4 when the paper’s results explicitly model directed dependence (L3) or loopy component interdependencies (L4) in the real-world system being studied.
Output format requirements:
Return valid JSON with exactly two fields and no additional text:
{
"judgement": "Level 1" or "Level 2" or "Level 3" or "Level 4",
"rationale": "Concise but specific justification grounded in the definitions above, explicitly referencing the paper’s primary model form, what nodes/edges mean, whether the structure is source→sink pathway-like (Level 3) versus loopy component-coupled system (Level 4), and why the chosen level best matches the dominant modeling contribution."
}

\end{Verbatim}

\section{Operationalizability Experiment Details}
\label{app:op-exp-details}

\begin{figure} [H]
    \centering
    \includegraphics[width=1.05\linewidth]{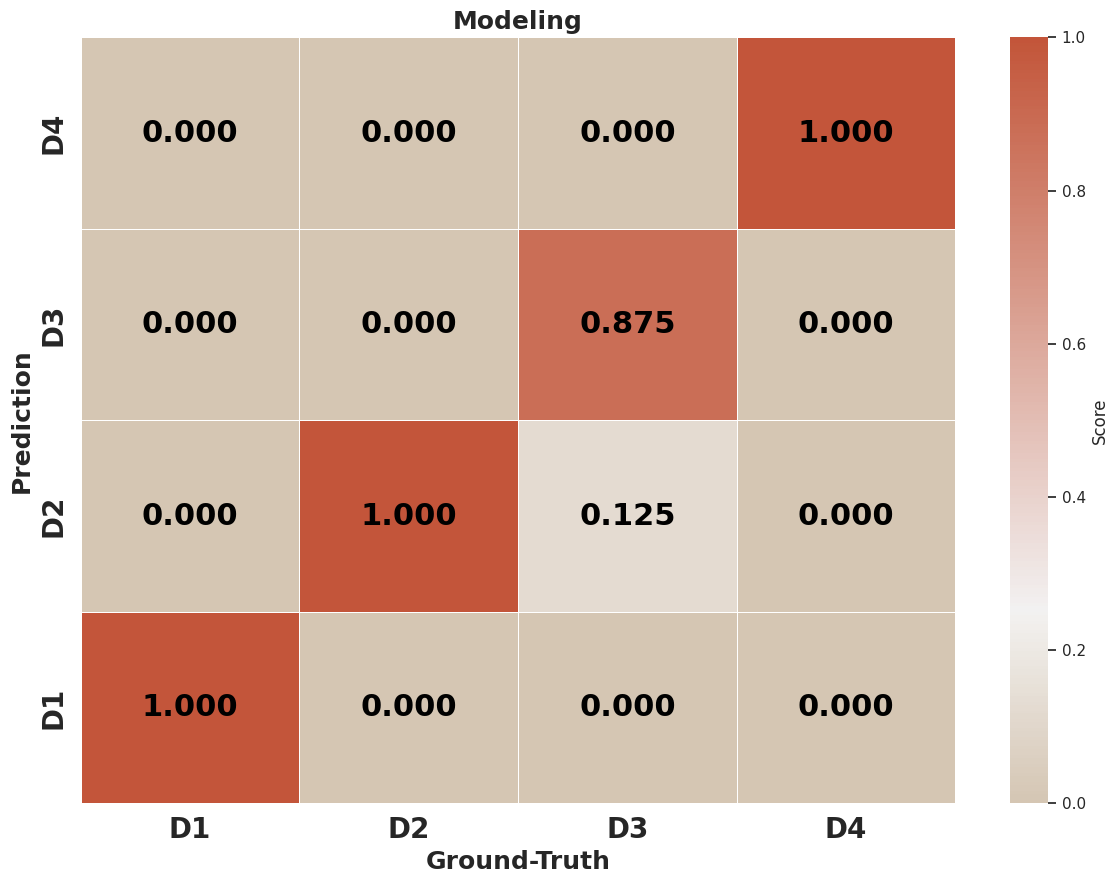}
    \label{fig:heatmap-modeling}
    \centering
    \includegraphics[width=1.05\linewidth]{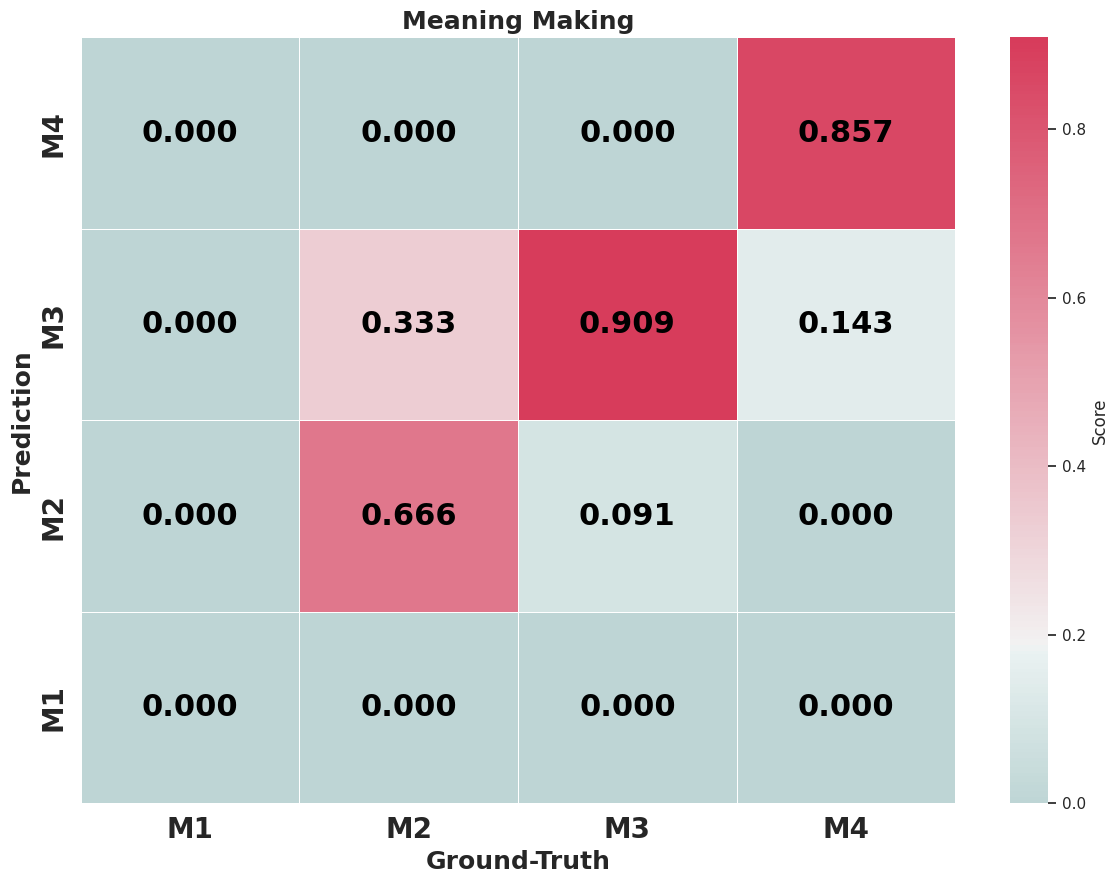}
    \label{fig:heatmap-meaning}
    \vspace{-6mm}
\end{figure}

\begin{table}[htbp]
\centering
\caption{Ground-truth distribution of sampled papers across \textcolor{orange}{Level of Meaning-Making} and \textcolor[HTML]{2567E9}{Level of Modeling}}.
\label{tab:gt_meaning}
\begin{tabular}{c c c c}
\toprule
\textbf{Levels} & \textbf{\# Papers} & \textbf{Levels} & \textbf{\# Papers}\\
\midrule
\mean{1} & 0 & \model{1} & 3\\ 
\mean{1} & 3 & \model{1} & 11\\
\mean{1} & 11 & \model{1} & 16\\
\mean{1} & 21 & \model{1} & 3\\
\bottomrule
\end{tabular}
\end{table}

\newblock
\textbf{Cohen's Kappa Scores:}
\begin{itemize}
    \item \textcolor{orange}{Meaning-Making} : \textbf{0.744}
    \item \textcolor[HTML]{2567E9}{Modeling}: \textbf{0.906}
\end{itemize}

We computed two confusion matrices comparing GPT-5.2 predictions (rows) with human ground-truth annotations (columns) on 35 sampled papers for both \textcolor{orange}{Level of Meaning-Making} and \textcolor[HTML]{2567E9}{Level of Modeling}. To calculate Cohen’s Kappa scores, the values in the confusion matrices were first normalized per ground-truth level to account for the uneven distribution of levels for both \textcolor{orange}{Level of Meaning-Making} and \textcolor[HTML]{2567E9}{Level of Modeling}, and the resulting matrices were used for the computation.

\section{Operationalizing the Agenda: Protocols, Fit Metrics, Privacy, and Induction}
\label{app:agenda_operationalization}

This appendix provides concrete “first implementation paths” for several agenda items raised in Sec.\ref{sec:agenda} — especially (i) moving from \model{1} coding outputs to \model{2}–\model{4} structured models, and (ii) operationalizing “fitness-to-corpus” for \model{2}/\model{3}/\model{4} structures (e.g., evidence coverage, contradiction rate, semantic consistency) in a way that is auditable and automatable.  These operationalizations are designed to be compatible with our modeling-level definitions (\model{2} time-without-causality; \model{3} directed causal pathways; \model{4} state+update+iteration/feedback) and with the paper’s emphasis on traceability and data-governance constraints for qualitative corpora.

\subsection{Representational protocols: a typed, evidence-anchored graph standard for \model{2}/\model{3}/\model{4}}
\label{app:repr_protocols}

\paragraph{Goal.}
We want one interoperable representation that (i) cleanly encodes the additional commitments as we move from \model{2}$\rightarrow$\model{3}$\rightarrow$\model{4}, (ii) keeps a strict audit trail from every node/edge to supporting excerpts (mitigating loss-of-traceability risks), and (iii) can be serialized, validated, and scored by “fitness-to-corpus” metrics (next subsection).

\paragraph{Core object: \texttt{QualGraph} (typed property multigraph).}
Let a qualitative model be a directed, typed property multigraph:
\[
G = (V, E), \quad v \in V \text{ nodes}, \quad e \in E \text{ edges}.
\]
Each node/edge carries (a) a \emph{semantic type}, (b) a \emph{human-readable definition}, and (c) \emph{evidence anchors} into the corpus.

\paragraph{Evidence anchors (required for every node and edge).}
We standardize an \texttt{EvidenceItem} that references an excerpt and records why it supports the claim.
\begin{Verbatim}[breaklines,breakanywhere,fontsize=\small]
EvidenceItem := {
  excerpt_id: str,            # stable ID
  doc_id: str,                # stable ID (may be hashed)
  span: [int, int],           # char offsets in excerpt text (or token offsets)
  time: Optional[TimeRef],    # if available; see \model{2}/\model{4}
  support_label: {SUPPORTS, CONTRADICTS, MENTIONS},
  rationale: str,             # short justification (human or model-generated)
  source: {HUMAN, MODEL, HYBRID},
  confidence: float           # [0,1], optional but recommended
}
\end{Verbatim}
This directly supports the “evidence-linked” requirement emphasized in the paper’s discussion of rigor and trustworthiness.

\paragraph{Node schema (shared across \model{2}/\model{3}/\model{4}).}
\begin{Verbatim}[breaklines,breakanywhere,fontsize=\small]
Node := {
  id: str,
  node_type: str,             # e.g., STAGE, EVENT, CONSTRUCT, STATE_VAR, ACTOR, POLICY
  label: str,                 # short name
  definition: str,            # what counts as an instance of this node
  aliases: List[str],         # optional, for retrieval/normalization
  attributes: Dict,           # optional typed fields (below)
  evidence: List[EvidenceItem]
}
\end{Verbatim}

\paragraph{Edge schema (shared across \model{2}/\model{3}/\model{4}).}
\begin{Verbatim}[breaklines,breakanywhere,fontsize=\small]
Edge := {
  id: str,
  src: str, dst: str,         # node IDs
  edge_type: str,             # e.g., NEXT, ENABLES, INCREASES, DECREASES
  polarity: Optional[int],    # +1/-1 for signed influence; None otherwise
  qualifiers: Dict,           # conditions, moderators, delay, scope, etc.
  evidence: List[EvidenceItem]
}
\end{Verbatim}

\paragraph{\model{2} protocol: stage/phase/timeline graphs (time without causality).}
\model{2} models make time first-class, but edges mean “next/then” rather than “produces/changes”. We encode \model{2} with:
\begin{itemize}
  \item \textbf{Node types:}
  \begin{itemize}
    \item \texttt{STAGE}: a phase/episode with an operational definition and boundary cues.
    \item \texttt{EVENT}: an atomic happening (optionally used; \model{2} may be stage-only).
    \item \texttt{MARKER} (optional): boundary markers (e.g., “handoff”, “policy change”).
  \end{itemize}
  \item \textbf{Edge types (temporal semantics only):}
  \begin{itemize}
    \item \texttt{NEXT(STAGE\_i, STAGE\_j)}: $i$ precedes $j$.
    \item \texttt{CONTAINS(STAGE, EVENT)}: event occurs within a stage.
    \item \texttt{OVERLAPS(STAGE\_i, STAGE\_j)} (optional): permitted if the analysis allows co-occurring phases.
  \end{itemize}
  \item \textbf{Required attributes:}
  \begin{itemize}
    \item For each \texttt{STAGE}: boundary rules or cues (\texttt{attributes.boundary\_cues}), e.g., lexical markers, role transitions, or criteria.
    \item For each \texttt{NEXT} edge: evidence excerpts that (i) instantiate both stages and (ii) warrant ordering (timestamps, narrative order, “before/after” language).
  \end{itemize}
\end{itemize}

\paragraph{\model{3} protocol: causal pathway graphs (directed influence).}
\model{3} edges encode causal/mechanism-like dependencies (enable/constrain/produce/change/mediate). We encode \model{3} with:
\begin{itemize}
  \item \textbf{Node types:} \texttt{CONSTRUCT} (factor/state/condition/strategy), \texttt{ACTOR} (optional), \texttt{OUTCOME} (optional).
  \item \textbf{Edge types (directional semantics):}
  \begin{itemize}
    \item \texttt{CAUSES}, \texttt{ENABLES}, \texttt{INHIBITS}, \texttt{CONSTRAINS}, \texttt{MEDIATES}, \texttt{MODERATES}.
  \end{itemize}
  \item \textbf{Required qualifiers on edges:}
  \begin{itemize}
    \item \texttt{qualifiers.mechanism\_sketch}: one-sentence “how” claim (even if partial).
    \item \texttt{qualifiers.conditions}: when/for whom the edge holds (if present in the corpus).
    \item \texttt{polarity}: optional sign for monotone effects (e.g., “more A tends to increase B”).
  \end{itemize}
  \item \textbf{Evidence requirement:} each edge must include (a) \emph{support excerpts} and (b) at least one \emph{searched-for counterexample} excerpt (even if labeled \texttt{IRRELEVANT}), to make contradiction-rate measurable (next subsection).
\end{itemize}

\paragraph{\model{4} protocol: qualitative system dynamics graphs (state + update + iteration).}
\model{4} models treat the phenomenon as an evolving system whose behavior depends on iterative state change (state at $t$ shapes state at $t{+}1$), often with feedback and delays. We encode \model{4} with:
\begin{itemize}
  \item \textbf{Node types:}
  \begin{itemize}
    \item \texttt{STATE\_VAR}: a system component with an interpretable state (capacity, resource, sentiment, constraint, norm, etc.).
    \item \texttt{FLOW} (optional): if using stock--flow style; otherwise omit.
    \item \texttt{REGIME} (optional): named qualitative regimes (e.g., “tight control”, “relaxed control”).
  \end{itemize}
  \item \textbf{Edge types (influence + dynamics qualifiers):}
  \begin{itemize}
    \item \texttt{INCREASES(src, dst)} with \texttt{polarity=+1}; \texttt{DECREASES} with \texttt{polarity=-1}.
    \item \texttt{DELAYED\_INFLUENCE}: same as above but with \texttt{qualifiers.delay\_type} (e.g., “short/long”, or “administrative delay”).
    \item \texttt{TRANSACTION}: concrete interaction updating states (used when the analysis specifies a mechanism-like transaction).
  \end{itemize}
  \item \textbf{Loop objects (first-class, derived or explicit):}
  \begin{Verbatim}[breaklines,breakanywhere,fontsize=\small]
Loop := {loop_id, node_ids: List[str], loop_sign: {REINFORCING,BALANCING,UNKNOWN},
         evidence: List[EvidenceItem]}
  \end{Verbatim}
  Loop sign can be computed from edge polarities when available; otherwise left \texttt{UNKNOWN} and evaluated via evidence coverage/closure (next subsection).
  \item \textbf{Minimal update semantics:}
  Each \texttt{STATE\_VAR} has a discrete qualitative state $\in \{\text{LOW},\text{MID},\text{HIGH}\}$ (or ordinal bins). Each timestep applies signed influences:
  \[
  x_j(t{+}1) = \text{clip}\Big(x_j(t) + \sum_{i} s_{ij}\cdot \Delta(x_i(t))\Big),
  \]
  where $s_{ij}\in\{-1,+1\}$ is edge polarity and $\Delta(\cdot)$ maps ordinal state to a signed “pressure” (e.g., LOW=-1, MID=0, HIGH=+1). This is intentionally a toy formalization, but it is sufficient to produce directional predictions that can be compared to corpus-extracted trend statements.
\end{itemize}

\paragraph{Serialization and validation.}
A first implementation can store \texttt{QualGraph} as JSON (+ JSONSchema / Pydantic validation), and load into NetworkX for graph operations. Recommended minimum validators:
(i) all nodes/edges have $\ge 1$ evidence item;
(ii) \model{2} \texttt{NEXT} edges form an acyclic order unless explicitly flagged as recurrent;
(iii) \model{4} edges have polarity if they are used in loop sign or simulation.

\subsection{Goodness-of-fit for qualitative models: evidence-anchored model checking with LLM judges}
\label{app:fit_metrics}

\paragraph{Motivation.}
The agenda calls for evaluation methodologies that operationalize “fitness-to-corpus” for \model{2}/\model{3}/\model{4} (coverage, contradiction rate, pattern matching, semantic consistency), but such metrics are nonstandard partly because QR outputs are narrative and because closed corpora complicate benchmarking. Here we define a scoring pipeline that produces (a) element-level support/contradiction labels, and (b) a model-level scalar score with transparent components.

\paragraph{Data model (corpus as testable datapoints).}
Let the corpus be segmented into excerpts:
\[
\begin{aligned}
\mathcal{X} &= \{x_1,\dots,x_n\},\\
x &:= (\text{text},\ \text{doc\_id},\ \text{time},\ \text{metadata}).
\end{aligned}
\]
Time may be an absolute timestamp (e.g., log time), a relative narrative index, or a document-local order. For interviews without timestamps, \texttt{time} can be set to the turn index.

Let a qualitative model (\model{2}/\model{3}/\model{4}) be a \texttt{QualGraph} $G$ with a set of \emph{atomic claims}:
\[
\mathcal{K}(G) = \mathcal{K}_{\text{node}} \cup \mathcal{K}_{\text{edge}} \cup \mathcal{K}_{\text{struct}},
\]
where:
\begin{itemize}
  \item $\mathcal{K}_{\text{node}}$: “Excerpt $x$ instantiates node $v$” (stage membership, construct presence, state-var mention).
  \item $\mathcal{K}_{\text{edge}}$: “Excerpt $x$ supports relation $e$” (NEXT, ENABLES, INCREASES, etc.).
  \item $\mathcal{K}_{\text{struct}}$: structure-dependent claims (e.g., \model{2} order constraints; \model{4} loop closure; \model{4} trend predictions from qualitative simulation).
\end{itemize}

\paragraph{LLM judge as a claim-labeling function (with auditability).}
Define a judge function that takes a claim $k$ and an excerpt $x$ and returns:
\[
J(k, x) \rightarrow (y, c, r),
\]
where $$y\in\{\textsc{Supports},\textsc{Contradicts},\textsc{Irrelevant}\}$$, $c\in[0,1]$ is confidence, and $r$ is a short rationale. We store $(y,c,r)$ back into \texttt{EvidenceItem}. This explicitly addresses the “loss of traceability” risk if multi-step synthesis is not evidence-linked.

\paragraph{Retrieval: choosing which excerpts to judge.}
For each node/edge, retrieve candidate excerpts via a lexical+embedding query using \texttt{label}, \texttt{aliases}, and \texttt{definition}. A first implementation can use:
\begin{itemize}
  \item BM25 over excerpts for lexical recall;
  \item dense embeddings for semantic recall;
  \item optional filters: same \texttt{doc\_id}, time window, speaker role.
\end{itemize}

\paragraph{Toy formalization: per-claim support and global fitness.}
For any claim $k$, let $S_k$ be the set of retrieved excerpts judged \textsc{Supports}, and $C_k$ those judged \textsc{Contradicts}. Define:
\[
\text{score}(k) = \frac{|S_k| - \lambda |C_k|}{|S_k| + |C_k| + \epsilon},
\]
with $\lambda>1$ penalizing contradictions and $\epsilon$ preventing division by zero.
Define \emph{coverage}:
\[
\text{cov}(k) = \mathbb{I}[|S_k| \ge m],
\]
i.e., the claim is “covered” if at least $m$ supporting excerpts exist.

Then:
\[
\begin{aligned}
\text{Fit}(G,\mathcal{X}) \;=\;
&\underbrace{\frac{1}{|\mathcal{K}|}\sum_{k\in\mathcal{K}}\text{score}(k)}_{\text{support--contradiction}}
\\[4pt]
&+\;\beta\underbrace{\frac{1}{|\mathcal{K}|}\sum_{k\in\mathcal{K}}\text{cov}(k)}_{\text{evidence coverage}}
\\[4pt]
&-\;\gamma\underbrace{\text{Complexity}(G)}_{\text{penalty}} .
\end{aligned}
\]

where a simple complexity term is $\text{Complexity}(G)=|V|+|E|$ (or $|V|+|E|+|\text{Loops}|$ for \model{4}). This yields an explicit lever for \emph{adaptive level selection}: prefer the most committal model (\model{4}>\model{3}>\model{2}) that wins under a complexity-penalized fit.

\paragraph{\model{2}-specific fit components (time without causality).}
\model{2}’s signature is ordering and segmentation. We add:
\begin{itemize}
  \item \textbf{Stage assignment coverage:} fraction of excerpts assignable to at least one stage with high-confidence \textsc{Supports}.
  \item \textbf{Order consistency:} extract a per-document stage sequence $\pi_d$ by labeling excerpts with stages and collapsing consecutive repeats; compute violations of \texttt{NEXT} constraints.
  \item \textbf{Boundary clarity:} for each adjacent pair of stages, measure how often the judge can identify boundary cues (explicit markers, turning points) versus “ambiguous transition” labels.
\end{itemize}

\paragraph{\model{3}-specific fit components (directed influence).}
\model{3} edges require directional warrant and mechanism evidence. We add:
\begin{itemize}
  \item \textbf{Directional support:} judge whether excerpts warrant $A\rightarrow B$ rather than $B\rightarrow A$.
  \item \textbf{Mechanism specificity:} judge whether the excerpt provides (even partial) “how” content matching \texttt{qualifiers.mechanism\_sketch}.
  \item \textbf{Counterevidence search:} retrieve excerpts mentioning $A$ without $B$ (or vice versa) and allow the judge to label them as \textsc{Contradicts} / \textsc{Irrelevant}—this operationalizes contradiction rate.
\end{itemize}

\paragraph{\model{4}-specific fit components (state + update + iteration).}
\model{4} requires evidence for feedback/iteration, not just rhetorical “systems” language. We add:
\begin{itemize}
  \item \textbf{Loop closure:} for each loop, require that each constituent edge has $\ge m$ supports \emph{and} at least one excerpt supports recurrence/cycling across time (e.g., “it comes back,” “repeats,” “oscillates,” “tighten then relax”).
  \item \textbf{Trend consistency via qualitative simulation:} run the toy ordinal update semantics to generate predicted directions (up/down/no-change) for key variables; extract trend statements from the corpus (LLM-based) and compute agreement.
\end{itemize}

\paragraph{Pseudocode}

\begin{Verbatim}[breaklines,breakanywhere,fontsize=\tiny]
Algorithm 1: Evidence-Anchored Goodness-of-Fit (\model{2}/\model{3}/\model{4})

Inputs:
  X: list of excerpts x = (text, doc_id, time, metadata)
  G: QualGraph with Nodes V and Edges E (and optional Loops)
  k: number of retrieved excerpts per claim
  m: minimum supports for coverage
  Judge(): LLM judge returning (label, confidence, rationale)

Preprocess:
  Build retrieval index over X (BM25 + embeddings).
  Optionally normalize time fields (per-doc ordering if needed).

1) NODE CHECKING (coverage + support)
  For each node v in V:
    Qv := build_query(v.label, v.aliases, v.definition)
    Cands := RetrieveTopK(X, Qv, k)
    For each x in Cands:
      y,c,r := Judge(claim="x instantiates v", excerpt=x, node=v)
      Store EvidenceItem(x, y,c,r) on v
    node_score[v] := score(claim=v, from stored EvidenceItems)

2) EDGE CHECKING (support + contradiction)
  For each edge e=(u->v) in E:
    Qe := build_query(u, v, e.edge_type, e.qualifiers)
    Cands_sup := RetrieveTopK(X, Qe, k)
    Cands_ctr := RetrieveCounterevidence(X, u, v, e)   # e.g., u w/out v, reverse order
    For each x in Cands_sup U Cands_ctr:
      y,c,r := Judge(claim="edge e holds", excerpt=x, edge=e)
      Store EvidenceItem(x, y,c,r) on e
    edge_score[e] := score(claim=e, from stored EvidenceItems)

3) STRUCTURE CHECKING (level-specific)
  If G is \model{2}:
     sequences := ExtractStageSequences(X, V, Judge)  # label excerpts -> stages -> collapse
     \model{2}_struct_score := 1 - OrderViolationRate(sequences, NEXT edges)
  If G is \model{4}:
     loops := DetectOrReadLoops(G)
     \model{4}_loop_score := LoopClosureScore(loops, edge_score)
     sim_pred := QualitativeSimulate(G, T=some horizon)
     trend_obs := ExtractTrendsFromCorpus(X, STATE_VAR nodes, Judge)
     \model{4}_trend_score := TrendAgreement(sim_pred, trend_obs)

4) AGGREGATE
  Fit := mean(node_score) + mean(edge_score) + level_struct_score - gamma*Complexity(G)
  Return Fit and a diagnostic report (worst edges, contradictions, low-coverage nodes).
\end{Verbatim}

\paragraph{Minimal judge templates (implementation hint).}
A workable first prompt pattern is:
\begin{itemize}
  \item \textbf{Node claim:} “Given the node definition, does the excerpt instantiate it? Output \{SUPPORTS/IRRELEVANT\} and one sentence why.”
  \item \textbf{Edge claim:} “Given edge type and direction (and polarity/conditions), does the excerpt support it, contradict it, or is it irrelevant? Output label + short rationale; do not infer beyond the excerpt.”
\end{itemize}
Multiple independent judge samples (e.g., $n{=}3$) and majority vote can reduce variance; low-agreement cases become candidates for human adjudication, consistent with the paper’s caution about hallucination and over-interpretation.

\subsection{Privacy-preserving “gold artifacts”: de-identification pipelines and tooling}
\label{app:privacy}

\paragraph{Why this is not optional.}
The paper explicitly notes that primary qualitative materials and codebooks are often closed due to privacy/confidentiality constraints, limiting reuse and evaluation, and that using external model services raises data governance and confidentiality risks unless de-identification and secure deployment are in place. Therefore, any recommendation to release shareable gold artifacts must come with a concrete privacy path.

\paragraph{What to share (pragmatic tiering).}
A realistic “shareable gold” bundle can be tiered:
\begin{itemize}
  \item \textbf{Tier A (lowest risk):} codebook (definitions + boundary/counterexamples), model graphs (\model{2}/\model{3}/\model{4}) with evidence pointers to excerpt IDs, and aggregate statistics (co-occurrence, precedence matrices). No raw text.
  \item \textbf{Tier B (moderate risk):} de-identified excerpts (redacted/pseudonymized) plus trace links to nodes/edges.
  \item \textbf{Tier C (high risk, controlled):} original raw transcripts under DUA / secure enclave; evaluation code runs “in place.”
\end{itemize}

\paragraph{De-identification operations (what to remove/transform).}
At minimum, target direct and quasi-identifiers:
\begin{itemize}
  \item \textbf{Direct identifiers:} names, emails, phone numbers, IDs, usernames.
  \item \textbf{Quasi-identifiers:} specific dates, fine-grained locations, rare job titles, unique events.
\end{itemize}
Typical transformations:
\begin{itemize}
  \item \textbf{Redaction:} remove span entirely (e.g., emails).
  \item \textbf{Pseudonymization:} replace with consistent placeholders (e.g., \texttt{[PERSON\_7]}), preserving within-document coreference.
  \item \textbf{Generalization:} “March 12, 2023” $\rightarrow$ “March 2023” or “2023”; “Cambridge, MA” $\rightarrow$ “Northeast US”.
  \item \textbf{Date shifting (optional):} add a random per-document offset while preserving intervals.
\end{itemize}

\paragraph{Python tooling suggestions (drop-in building blocks).}
A first implementation can combine:
\begin{itemize}
  \item \texttt{presidio-analyzer} + \texttt{presidio-anonymizer} (PII detection + replacement rules)
  \item \texttt{spacy} (NER; custom entity ruler for domain-specific identifiers)
  \item \texttt{stanza} (alternative NER)
  \item \texttt{scrubadub} (simple PII scrubbing for common patterns)
  \item \texttt{phonenumbers} (robust phone parsing), regex for IDs, \texttt{dateparser} for date normalization
  \item \texttt{rapidfuzz} (string similarity to cluster mentions for consistent pseudonyms)
\end{itemize}
De-identification is not a guarantee; high-risk releases should still use Tier C controls.

\paragraph{Pseudocode}
\begin{Verbatim}[breaklines,breakanywhere,fontsize=\tiny]
Algorithm 2: De-identify a qualitative corpus with consistent pseudonyms

Inputs:
  X: excerpts (text, doc_id, time, ...)
  pii_detectors: [Presidio, spaCy NER, regex detectors]
  policy: which entity types to redact vs pseudonymize vs generalize
Outputs:
  X': de-identified excerpts
  map_store: secure mapping (entity -> pseudonym) stored separately

1) Detect PII spans:
   spans := []
   for x in X:
     spans_x := UnionDetect(x.text, pii_detectors)  # merge overlapping spans
     spans.append((x.excerpt_id, spans_x))

2) Canonicalize + cluster entities for consistency:
   ents := ExtractEntityStrings(spans)
   clusters := ClusterByStringSimilarity(ents, method=rapidfuzz, threshold=t)

3) Assign replacements:
   for cluster in clusters:
     pseud := MakePlaceholder(cluster.type)  # e.g., [PERSON_3], [ORG_2]
     StoreSecure(map_store, cluster -> pseud)  # access-controlled

4) Apply replacements:
   for x in X:
     x'.text := ReplaceSpans(x.text, spans_x, map_store, policy)
     x'.doc_id := HashOrRemap(doc_id)  # optional
   return X'
\end{Verbatim}

\subsection{A simple code-relation induction algorithm: traceable pairwise induction + graph sparsification}
\label{app:code_relation_induction}

\paragraph{Goal.}
To move beyond treating codes as independent labels (\model{1}) and instead induce explicit inter-code relations (axial coding), we need algorithms that operationalize the transition from coding datapoints to modeling relations. Below is a lightweight baseline that (i) proposes candidate relations via simple statistics/ordering cues, (ii) types and orients them using an LLM judge \emph{with evidence}, and (iii) produces a sparse, auditable graph ready for \model{2}/\model{3}/\model{4} representations.

\paragraph{Inputs.}
Assume a coded corpus where each excerpt $x$ has a set of codes $C(x)\subseteq \mathcal{C}$ (from a codebook), and (optionally) a time index.

\paragraph{Step A: candidate generation (fast, deterministic).}
Compute for each code pair $(a,b)$:
\begin{itemize}
  \item \textbf{Co-occurrence strength:} PMI or log-odds based on counts over excerpts:
  \[
  \text{PMI}(a,b)=\log \frac{p(a,b)}{p(a)p(b)}.
  \]
  \item \textbf{Temporal precedence (if time exists):} within each document, count how often $a$ appears before $b$ within a window $w$:
    {\small
    \[
    \begin{aligned}
    \text{Prec}(a\!\rightarrow\! b)
    &=\frac{\#(a\ \text{before}\ b)}{
    \#(a\ \text{before}\ b)+\#(b\ \text{before}\ a)+\epsilon }.
    \end{aligned}
    \]
    }
\end{itemize}
Generate candidate edges by thresholds (e.g., PMI $>$ $\tau_{\text{pmi}}$ for association; precedence $>$ $\tau_{\text{prec}}$ for \texttt{NEXT}-like direction).

\paragraph{Step B: relation typing and direction via evidence-anchored LLM judgments.}
For each candidate pair, retrieve a small evidence set:
\begin{itemize}
  \item \textbf{Support candidates:} excerpts where both codes occur (or occur in close temporal proximity).
  \item \textbf{Counterevidence candidates:} excerpts where $a$ occurs without $b$, or where order reverses.
\end{itemize}
Ask the judge to classify relation type from a small ontology:
\[
\begin{aligned}
\{&
\texttt{NEXT},\ \texttt{CO\_OCCURS},\ \texttt{ENABLES},\ \texttt{INHIBITS},\\
&\texttt{CAUSES},\ \texttt{PART\_OF},\ \texttt{MODERATES},\ \texttt{NONE}
\}.
\end{aligned}
\]

Store its decisions as \texttt{EvidenceItem}s on the induced edge.

\paragraph{Step C: graph sparsification (keep it interpretable).}
To avoid “hairball” graphs:
\begin{itemize}
  \item Keep top-$K$ edges per node by support score.
  \item For \model{2} candidates, apply transitive reduction on \texttt{NEXT} edges (when acyclic) to keep only necessary order constraints.
  \item For \model{4} candidates, detect cycles and aggregate them into \texttt{Loop} objects; compute loop sign if edges are signed.
\end{itemize}

\paragraph{What this buys us immediately.}
This baseline directly instantiates the paper’s call for “code relation induction algorithms” that explicitly move from open coding to inter-code relations (axial coding), and it produces artifacts that are (i) graph-native (usable for \model{2}/\model{3}/\model{4}) and (ii) evidence-linked (scorable by the goodness-of-fit pipeline above), while acknowledging that sensitive corpora require de-identification and careful governance.

\end{document}